\documentclass[10pt,journal,compsoc]{IEEEtran}

\ifCLASSOPTIONcompsoc
  \usepackage[nocompress]{cite}
\else
  \usepackage{cite}
\fi

\ifCLASSINFOpdf
\else
\fi

\usepackage{amsmath}
\usepackage{amsfonts}
\usepackage{algorithmic}
\usepackage{graphicx}
\usepackage{mathtools}
\usepackage{multirow}
\usepackage[marginal]{footmisc}
\usepackage{ntheorem}
\usepackage{url}

\newtheorem{myThm}{Theorem}
\newtheorem*{prf}{Proof}

\newcommand*{\QEDB}{\hfill\ensuremath{\square}}

\usepackage{color}

\newcommand{\bm}[1]{\mbox{\boldmath{$#1$}}}

\begin{document}

\title{Symbiotic Graph Neural Networks \\for 3D Skeleton-based Human Action Recognition and Motion Prediction}
\author{Maosen~Li,~\IEEEmembership{Student Member,~IEEE,}
        Siheng~Chen,~\IEEEmembership{Member,~IEEE,}
        Xu~Chen,
        Ya~Zhang,~\IEEEmembership{Member,~IEEE,}
        Yanfeng~Wang, 
        and~Qi~Tian~\IEEEmembership{Fellow,~IEEE}
\IEEEcompsocitemizethanks{\IEEEcompsocthanksitem M. Li is with the Cooperative Medianet Innovation Center and the Shanghai Key Laboratory of Multimedia Processing and Transmissions, Shanghai
Jiao Tong University, Shanghai 200240, China.\protect\\
E-mail: maosen\_li@sjtu.edu.cn
\IEEEcompsocthanksitem S. Chen is with Mitsubishi Electric Research Laboratories, Cambridge, MA 02139, USA.\protect\\
E-mail: schen@merl.com.
\IEEEcompsocthanksitem X. Chen, Y. Zhang and Y. Wang are with the Cooperative Medianet Innovation Center and the Shanghai Key Laboratory of Multimedia Processing and Transmissions, Shanghai Jiao Tong University, Shanghai 200240, China.\protect
\IEEEcompsocthanksitem Q. Tian is with the Huawei Noah's Ark Lab, Shenzhen, Guangdong 518129, China.\protect\\
E-mail: tianqi1@huawei.com}
\thanks{Parts of this paper appear in~\cite{Li_cvpr_2019}.}}

% \markboth{}

\IEEEtitleabstractindextext{
\begin{abstract}

3D skeleton-based action recognition and motion prediction are two essential problems of human activity understanding. In many previous works: 1) they studied two tasks separately, neglecting internal correlations; 2) they did not capture sufficient relations inside the body. To address these issues, we propose a symbiotic model to handle two tasks jointly; and we propose two scales of graphs to explicitly capture relations among body-joints and body-parts. Together, we propose symbiotic graph neural networks, which contains a backbone, an action-recognition head, and a motion-prediction head. Two heads are trained jointly and enhance each other. For the backbone, we propose multi-branch multi-scale graph convolution networks to extract spatial and temporal features. The multi-scale graph convolution networks are based on joint-scale and part-scale graphs. The joint-scale graphs contain actional graphs, capturing action-based relations, and structural graphs, capturing physical constraints. The part-scale graphs integrate body-joints to form specific parts, representing high-level relations. Moreover, dual bone-based graphs and networks are proposed to learn complementary features. We conduct extensive experiments for skeleton-based action recognition and motion prediction with four datasets, NTU-RGB+D, Kinetics, Human3.6M, and CMU Mocap. Experiments show that our symbiotic graph neural networks achieve better performances on both tasks compared to the state-of-the-art methods. The code is relased at {\url{github.com/limaosen0/Sym-GNN}}
\end{abstract}

\begin{IEEEkeywords}
3D skeleton-based action recognition, motion prediction, multi-scale graph convolution networks, graph inference.
\end{IEEEkeywords}}

\maketitle
\IEEEdisplaynontitleabstractindextext
\IEEEpeerreviewmaketitle

\IEEEraisesectionheading{\section{Introduction}\label{sec:introduction}}

\IEEEPARstart{H}{uman} action recognition and motion prediction are crucial problems in computer vision, being widely applicable to surveillance~\cite{Gaur_iccv_2011}, pedestrian tracking~\cite{Huang_eccv_2014}, and human-machine interaction~\cite{gui-2018-110272}. {Respectively, action recognition aims to accurately classify the categories of query actions~\cite{6942210}; and motion prediction forecasts the future movements based on observations~\cite{Shi_2018_ECCV}.} 

The data of actions can be represented with various formats, including RGB videos~\cite{Wang_2018_ECCV} and 3D skeleton data~\cite{8327922}. Notably, 3D skeleton data, locating 3D body-joints, is shown to be effective in action representation, efficient in computation, as well as robust against environmental noise~\cite{ijcai_ChaoLi}. In this work, we focus on action recognition and motion prediction based on the 3D skeleton data.

In most previous studies, 3D skeleton-based action recognition and motion prediction are treated separately as the former needs to discriminate the classes; while the latter generates the future poses. For action recognition, methods employed full action sequences for pattern learning~\cite{Vemulapalli_2014_CVPR, Fernando_2015_CVPR, Du_2015_CVPR, vis_cnn, AAAI1817135}; however, with the long-term inputs, these methods failed in some real-time applications due to the hysteretic discrimination, while the model should response as early as possible. As for motion prediction, previous works built generative models~\cite{Lehrmann_2014_CVPR, Jain_2016_CVPR, Martinez_2017_CVPR, Gui_2018_ECCV}; they learned motion dynamics, but often {ignored} semantics. Actually, there are mutual promotions between the tasks of action recognition and motion prediction, while previous works rarely explored them. For example, the classifier provides the action categories as the auxiliary information to guide prediction, as well as the predictor preserves more detailed information for accurate recognition via self-supervision. Considering to exploit mutual promotions, we aim to develop a symbiotic method to enable action recognition and motion prediction simultaneously.

\begin{figure*}
\centering
    \includegraphics[width=15.6cm]{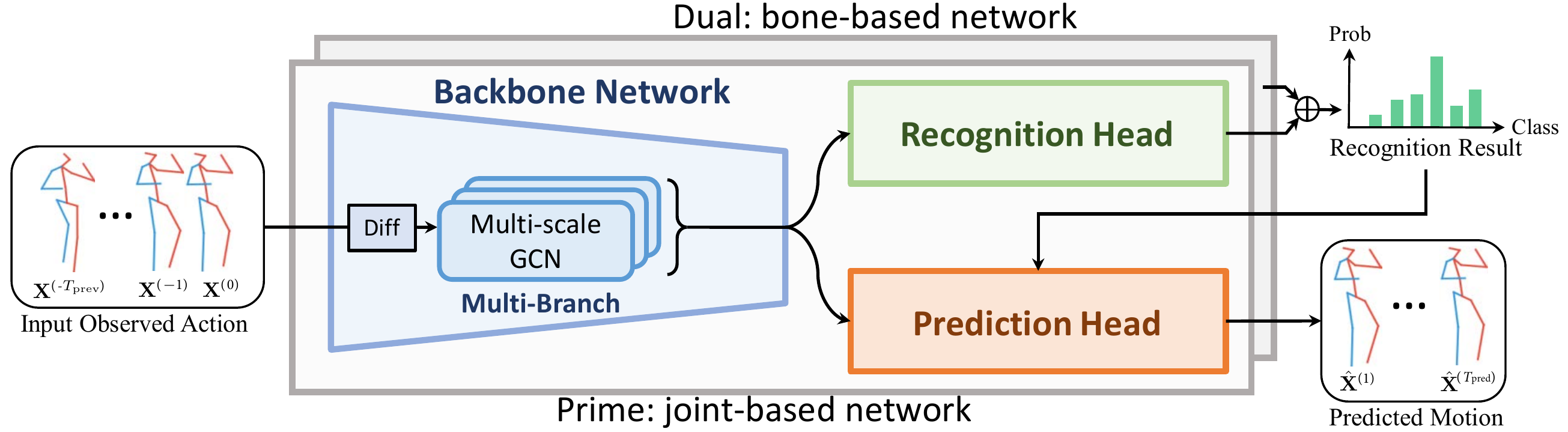}
    \caption{Symbiotic Graph Neural Networks (Sym-GNN) contains a prime joint-based network to learn body-joint-based features, and a dual bone-based network to learn body-bone-based features. Each network has three main modules: a backbone, an action-recognition head, and a motion-prediction head. The backbone is essentially multi-branch multi-scale graph convolution networks (multi-branch multi-scale GCN). The action-recognition head and the motion-prediction head predict the action category and future poses, respectively. The predicted action category is further used in the motion-prediction head. This symbiotic design allows the two heads to enhance each other.
    }
    \label{fig:pipeline}
    \vspace{-10pt}
\end{figure*}

{For both 3D skeleton-based action recognition and motion prediction, the key is to effectively capture the motion patterns of various actions.} A lot of efforts have been made to push towards this direction. Concretely, some traditional attempts often vectorized all the joints to a pose vector and built hand-crafted models for feature learning~\cite{cvpr_wang_2012,ijcai_Hussein_2013,Vemulapalli_2014_CVPR,Lehrmann_2014_CVPR, NIPS2005_2783, Taylor_2007_NIPS, NIPS2008_3567}. Recently, some deep models based on either convolutional neural networks (CNN) or recurrent neural networks (RNN) learned high-level features from data~\cite{Fernando_2015_CVPR, a8014941, Du_2015_CVPR, vis_cnn, Jain_2016_CVPR, Martinez_2017_CVPR,quater, Gui_2018_ECCV,Gui_2018_ECCV2,AAAI_Kundu}; however, these methods rarely {investigated} the joint relations, missing crucial activity dynamics. To capture richer features, several works {exploited} joint relations from various aspects. ~\cite{AAAI1817135} proposed skeleton-graphs with nodes as joints and edges as bones. ~\cite{Si_2018_ECCV,Si_2019_CVPR,Jain_2016_CVPR} built the relations between body-parts, such as limbs. ~\cite{AAAI_Guo} merged individual part features. ~\cite{Li_2018_CVPR} leveraged spatial convolutions on pose vectors, but it varied on joint permutation. These works aggregated information from local or coarse neighborhoods. Notably, some relations may exist among action-related joints, such as hands and feet moving collaboratively during walking. Moreover, some methods of motion prediction fed ground-truth action categories to enhance performance in both training and testing phases, but the true labels are hard to obtain during the real-world scenarios. To solve those issues, we construct graphs to model both local and long-range body relations and use graph convolutions to capture informative spatial features.

In this paper, we propose a novel model called \emph{symbiotic graph neural network} (Sym-GNN),
which handles 3D skeleton-based action recognition and motion prediction simultaneously and uses graph-based operations to capture spatial features. As basic operators of Sym-GNN, we propose the \emph{joint-scale graph convolution} (JGC) {  and \emph{part-scale graph convolution} (PGC) operators to extract multi-scale spatial information.} JGC is based on two types of graphs: actional graphs and structural graphs. The actional graphs are learned from 3D skeleton data by an \emph{actional graph inference module} (AGIM), capturing action-based relations; the structural graphs are built by extending the skeleton graphs, capturing physical constraints. {  PGC is based on a part-scale graph, whose nodes are integrated body-part features and edges are based on body-part connections.} We also propose a difference operator to extract multiple orders of motion differences, reflecting positions, velocities, and accelerations of body-joints.

The proposed Sym-GNN consists of a backbone, called \emph{multi-branch multi-scale graph convolutional network} (multi-branch multi-scale GCN), an action-recognition head and a motion-prediction head; see Fig.~\ref{fig:pipeline}. The backbone uses joint-scale and part-scale graphs for spatial relations presentation and high-level feature extraction; two heads work on two separated tasks. Moreover, there are task promotions, i.e. the action-recognition head determines action categories, which is used to enhance prediction performance; meanwhile, the motion-prediction head predicts poses and improves recognition by promoting self-supervision and preserving detailed features. The model is trained through a multitasking paradigm. Additionally, we build a dual bone-based network which treats bones as graph nodes and learns bone features to obtain complementary semantics for more effective classification and prediction.

To validate the Sym-GNN, we conduct extensive experiments on four large-scale datasets: NTU-RGB+D~\cite{Shahroudy_2016_CVPR}, Kinetics~\cite{AAAI1817135}, Human 3.6M~\cite{h36m}, and CMU Mocap\footnote{http://mocap.cs.cmu.edu/}. The results show that 1) Sym-GNN outperforms the state-of-the-art methods in both action recognition and motion prediction; 2) Using the symbiotic model to train the two tasks simultaneously produces better performance than using individual models; and 3) the multi-scale graphs model complicated relations between body-joints and body-parts, and the proposed JGC extract informative spatial features. 

Overall, the main contributions in this paper are summarized as follows:
{
\begin{itemize}
    \item {\bf Multitasking framework.} We propose novel \emph{symbiotic graph neural networks} (Sym-GNN) to achieve 3D skeleton-based action recognition and motion prediction in a multitasking framework. Sym-GNN contains a backbone, an action-recognition head, and a motion-prediction head. We exploit the mutual promotion between two heads, leading to  improvements in both tasks; see Section~\ref{sec:sym_gnn}

    \item {\bf Basic operators.} We propose novel operators to extract information from 3D skeleton data: 
    1) a \emph{joint-scale graph convolution} operator is proposed to extract joint-level spatial features based on both actional and structral graphs; see Section~\ref{sec:as-gtc}; 
    2) a \emph{part-scale graph convolution} operator is proposed to extract part-level spatial features based on part-scale graphs; see Section~\ref{sec:p-gtc}; 
    3) a pair of \emph{bidirectional fusion} operators is proposed to fuse information across two scales; see Section~\ref{sec:multi_scale_gcn} 
    and 4) a \emph{difference} operator is proposed to extract temporal features; see Section~\ref{sec:difference_operator}; and
    
    \item {\bf Experimental findings.} We conduct extensive experiments for both tasks of 3D skeleton-based action recognition and motion prediction. The results show that Sym-GNN outperforms  the state-of-the-art methods in both tasks; see Section~\ref{sec:exp}
\end{itemize}
}

\section{Related Works}
{\bf 3D skeleton-based action recognition. } 
Numerous methods are proposed for 3D skeleton-based action recognition.
Conventionally, some models learned semantics based on hand-crafted features and physical intuitions~\cite{cvpr_wang_2012,ijcai_Hussein_2013,Vemulapalli_2014_CVPR}. In the deep learning era, models automatically learn features from data. Some recurrent-neural-network-based (RNN-based) models captured the temporal dependencies between consecutive frames~\cite{Du_2015_CVPR,Liu_2016_eccv}. Moreover, convolutional neural networks (CNN) also achieve remarkable results~\cite{a8014941,vis_cnn}. Recently, the graph-based approaches drew many attentions~\cite{AAAI1817135, Si_2018_ECCV,Si_2019_CVPR,2sAGCN,DGNN,Li_cvpr_2019,STGR-GCN,motif-GCN}. In this work, we adopt the graph-based approach. We construct multi-scale graphs adaptively from data, capturing useful and comprehensive information about actions.

{\bf 3D skeleton-based motion prediction. } 
In earlier studies, state models were considered to predict future motions~\cite{Lehrmann_2014_CVPR,NIPS2005_2783,nips_2000}. Recently, deep learning technique plays increasingly important roles. Some RNN-based methods learned the dynamics from sequences~\cite{Jain_2016_CVPR, Martinez_2017_CVPR, Walker_2017_ICCV, NIPS2016_6552,Fragkiadaki_2015_ICCV,AAAI_Kundu}. Moreover, adversarial mechanics and geodesic loss could further improve predictions~\cite{Gui_2018_ECCV}. As for our method, we use graph structures to explicitly model the relations between body-joints and body-parts, guiding the networks to learn local and non-local motion patterns.

{\bf Graph deep learning. } 
Graphs, focused on by many recent studies, are effective to express data associated with non-grid structures~\cite{AAAI1817135, Verma_2018_CVPR, valsesia2018learning,Li2018learning, NIPS2016_6081, kipf_iclr2017, NIPS2017_6703, Si_2018_ECCV}. Given the fixed topologies, previous works explored to propagate node features based on the spectral domain~\cite{NIPS2016_6081, kipf_iclr2017} or the vertex domain~\cite{NIPS2017_6703}. \cite{AAAI1817135, Li_cvpr_2019,Si_2019_CVPR,DGNN,2sAGCN} leveraged graph convolution for 3D skeleton-based action recognition. \cite{Jain_2016_CVPR} also considered the skeleton-based relations for motion prediction. In this paper, we propose multi-scale graphs to represent multiple relations: joint-scale and part-scale relations. Then, we propose novel graph convolution operators to extract deep features for action recognition and motion prediction. Different from~\cite{Li_cvpr_2019} obtaining multiple actional graphs with complicated inference processes, our method employs more efficient graph learning operations.

\section{Problem Formulation}
\label{sec:problem}
In this paper, we study 3D skeleton-based action recognition and motion prediction jointly. We here use the 3D joint positions along the time to represent the action sequences. Mathematically, let the action pose at time stamp $t$ be $\mathbf{X}^{(t)}\in \mathbb{R}^{M\times D_{\bf x}}$, where $t>0$ indicates the future frames, otherwise the observed frames; notably, $t=0$ denotes the current frame. $M$ is the number of joints and $D_{\bf x}=3$ reflects the 3D joint positions.  The action pose is essentially associated with a skeleton graph, which represents the pairwise bone connectivity. We can represent a skeleton graph by a binary adjacent matrix; that is, $\mathbf{A} \in\{0,1\}^{M\times M}$, where the $(i,j)$th elements $\left( \mathbf{A} \right)_{ij}=1$ when the $i$th and the $j$th body-joints are connected with bones, and $\left(\mathbf{A} \right)_{ij}=0$, otherwise. Note that $\mathbf{A}$ includes self-loops.  

For an action sequence belonging to one class, we have $\{\mathcal{X}_{\rm prev}, \mathcal{X}_{\rm pred}, \mathbf{y}\}$, where $\mathcal{X}_{\rm prev}=[\mathbf{X}^{(-T_{\rm prev})},\dots,\mathbf{X}^{(0)}]\in\mathbb{R}^{T_{\rm prev}\times M\times D_{\bf x}}$ denotes the previous motion tensor; $\mathcal{X}_{\rm prev}=[\mathbf{X}^{(1)},\dots,\mathbf{X}^{(-T_{\rm pred})}]\in\mathbb{R}^{T_{\rm pred}\times M\times D_{\bf x}}$ denotes the future motion tensor; $T_{\rm prev}$ and $T_{\rm pred}$ are the frame numbers of previous and future motions, respectively; and one-hot vector $\mathbf{y}\in\{0,1\}^{C}$ denotes the class-label in $C$ possible classes. Let $\mathcal{F}(\cdot)$ be the overall model. The discriminated class category $\hat{\mathbf{y}}$ and the predicted motion $\hat{\mathcal{X}}_{\rm pred}$ are formulated as
$$\hat{\mathbf{y}}, \hat{\mathcal{X}}_{\rm pred} = \mathcal{F}(\mathcal{X}_{\rm prev};{\bm \theta}_{\rm bk}, {\bm \theta}_{\rm recg}, {\bm \theta}_{\rm pred}),$$
where ${\bm \theta}_{\rm bk}$, ${\bm \theta}_{\rm recg}$ and ${\bm \theta}_{\rm pred}$ denote trainable parameters of the backbone, the action-recognition head and the motion-prediction head, respectively.

% Existing works usually considered action recognition and motion prediction as independent tasks and studied them sepatately; however, they neglected protential mutual promotions between each other. Concretely, the motion predictor can preserve more detailed information to improve recognition; and the recognizer can provide inferred labels to guide more precise prediction performance. We attempt to model the action recognition and motion prediction jointly with a multitasking framework, exploiting the mutual promotion between two tasks.

% Moreover, previous works of both 3D skeleton-based action recognition or motion prediction though tried to consider the joint relations during the actions~\cite{AAAI1817135, Jain_2016_CVPR}; they usually built graphs only according to body structures, which failed to exploit more general movement relations for various actions. In other words, these graphs only represented joint spatial dependencies in very local regions, thus some structurally distant relations are missed. Many distant joints will move collaboratively and carry some key patterns of actions. We should construct more general graphs from data, which model the actional relations between the interacting joints and represent the specific movements.

\section{Basic Components}
In this section, we propose some novel components in our model. We first propose some joint-scale graph operators, extracting features among body-joints; we next propose the part-scale graph operators, extracting features among body-parts; finally, we propose a difference operator to provide richer motion priors.

\subsection{Joint-Scale Graph Operators}
\begin{figure}[tb]
    \centering
    \includegraphics[width=7cm]{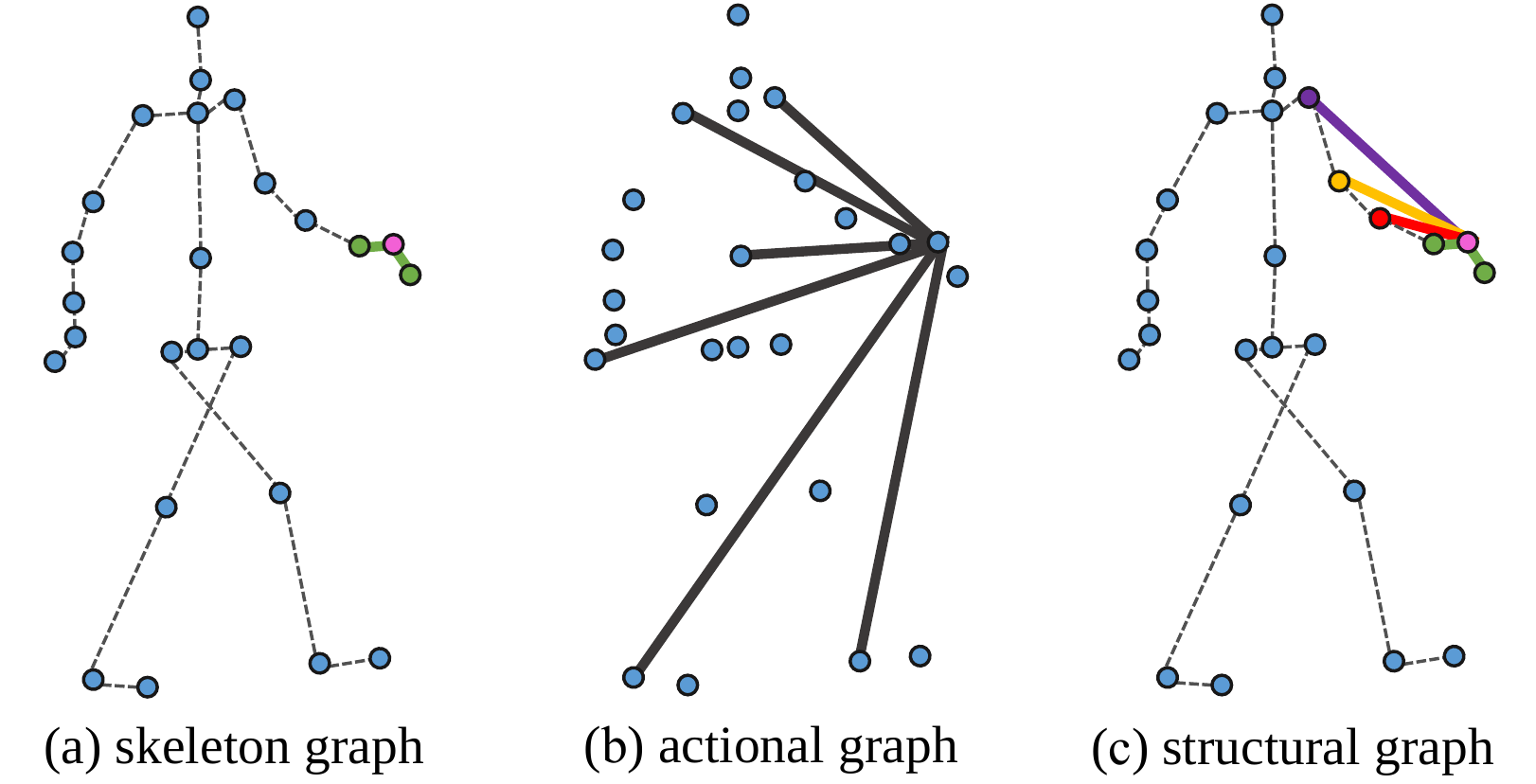}
    \vspace{-10pt}
    \caption{Examples of joint-scale graphs for walking. In the joint scale, we consider an actional graph (Plot (b)) and a structural graph (Plot (c)), which is an extension of a skeleton graph (Plot (a)).  In each graph, the edges from "Left Hand" to its neighbors are shown in solid lines and other links in the skeleton are shown in dashed lines.}
    \label{fig:link_type}
    \vspace{-10pt}
\end{figure}
To model the joint relations, we build joint-scale graphs including actional graphs, capturing moving interactions between joints even without bone-connection, and structural graphs, extending the skeleton structures to represent physical constraints. Fig.~\ref{fig:link_type} sketches some examples. Plot (a) shows a skeleton graph with local neighborhood; plot (b) shows an actional graph, which captures action-based dependencies, e.g. `Left Hand' is linked with `Right Hand' and feet during walking; plot (c) shows a structural graph, which allows `Left Hand' to link with entire arm. 

As follows, we propose the construction of joint-scale actional and structural graphs. And, we present the joint-scale graph and temporal convolution (J-GTC) block to learn spatial and temporal features of sequential actions.

\subsubsection{Actional Graph Convolution}
\label{sec:joint-scale-graph}
For different movements, some structurally distant joints may interact, leading to action-based relations. For example, a walking person moves hands and feet collaboratively. To represent actional relations, we employ an actional graph: $\mathcal{G}_{\rm act}({V},  \mathbf{A}_{\rm act})$, where ${V}=\{v_1, \dots, v_M\}$ is the joint set and  $\mathbf{A}_{\rm act} \in \mathbb{R}^{M \times M}$ is the adjacency matrix that reveals the pairwise joint-scale actional relations. To obtain this topology, we propose a data-adaptive module, called \emph{actional graphs inference module} (AGIM), to learn  $\mathbf{A}_{\rm act}$ purely from observations without knowing action categories. 

To utilize the body dynamics, we let the vector representation of the $i$th joint positions across all observed frames be 
$\mathbf{x}_i  = {\rm vec} \left( \mathcal{X}_{\mathrm{prev}}[:, i, :] \right) \in \mathbb{R}^{D_{\bf x} T_{\mathrm{prev}}}$, 
which includes the previous positions. To learn all relations, we propagate pose information between body-joints and possible edges.  We first initialize $\mathbf{p}_i^{\left\langle 0 \right\rangle}=f_{\rm v}^{\left\langle 0 \right\rangle }(\mathbf{x}_i) \in\mathbb{R}^{D_{\rm v}}$, where $f_{\rm v}^{\left\langle 0 \right\rangle }(\cdot)$  is a multilayer perceptron (MLP) that maps the raw joint moving data $\mathbf{x}_i$ to joint features $\mathbf{p}_i^{\left\langle 0 \right\rangle}$.  In the $k$th iteration,  the features are propagated as follows:
\begin{subequations}
\begin{align}
\label{eq:n2e}
    \mathbf{q}_{i,j}^{\left\langle k \right\rangle} &= f_{\rm e}^{\left\langle k \right\rangle}  \left(  \left[ \mathbf{p}_i^{\left\langle k-1 \right\rangle},\mathbf{p}_j^{\left\langle k-1 \right\rangle} \right]  \right) \in  \mathbb{R}^{D_{\rm e}},
    \\
\label{eq:e2n}
    \mathbf{p}_i^{\left\langle k \right\rangle} &= f_{\rm v}^{\left\langle k \right\rangle} \left(\frac{1}{M-1}\sum_{v_j \in {\cal V}, j\neq i}\mathbf{q}_{i,j}^{\left\langle k \right\rangle} \right)  \in \mathbb{R}^{D_{\rm v}},
\end{align}
\end{subequations}
where $\mathbf{p}_{i}^{\left\langle k \right\rangle}$, $\mathbf{q}_{i,j}^{\left\langle k \right\rangle}$  are  the  feature vectors of the $i$th joints and  the edge connecting the $i$th and $j$th joints  at the $k$th iteration; $f_{\rm e}^{\left\langle k \right\rangle}(\cdot)$ and $f_{\rm v}^{\left\langle k \right\rangle}(\cdot)$ are two MLP-formed feature extractors; $[\cdot,\cdot]$ is the concatenation; $D_{\rm e}$ and $D_{\rm v}$ denote the dimensions of edge and joint features, respectively.~\eqref{eq:n2e} maps a pair of joint features to the in-between edge features; ~\eqref{eq:e2n} aggregates all edge features associated with the same joint and maps to the corresponding joint features. After $K$ iterations, information are fully propagated between joints and edges; in other words, each joint feature obtained has aggregated the integrated information in a long-range.

Given any joint feature after $K$ iterations, $\mathbf{p}^{\left\langle K \right\rangle}$, we compute the relation strength between each pair of joints, leading to an actional graph. We build two individual embedding networks, $f_{\rm emb}(\cdot)$ and $g_{\rm emb}(\cdot)$, to further learn the high-level representations of joints. The $(i,j)$th element of the adjacent matrix of actional graph is formulated as
\begin{equation}
\label{eq:action_graph}
  \left( \mathbf{A}_{\rm act} \right)_{i,j} \ = \ \frac{\exp{(\mathbf{f}_{i}^{\rm T}\mathbf{g}_{j})}}{\sum_{k=1}^M\exp{(\mathbf{f}_{i}^{\rm T}\mathbf{g}_{k})}} \ \in \ [0,1] 
\end{equation}
where  $\mathbf{f}_{i} = f_{\rm emb}(\mathbf{p}_i^{\left\langle K \right\rangle})$  and $\mathbf{g}_{i} = g_{\rm emb}(\mathbf{p}_i^{\left\langle K \right\rangle}) \in \mathbb{R}^{D_{\rm emb}}$ are the two different embeddings of joint $v_i$. Notably, $(\mathbf{A}_{\rm act})_{i,j}\neq(\mathbf{A}_{\rm act})_{j,i}$, indicating incoming and outgoing relations between joints.~\eqref{eq:action_graph} uses the softmax to normalize the edge weights and promote a few large ones.
\begin{figure}[tb]
    \centering
    \includegraphics[width=8.8cm]{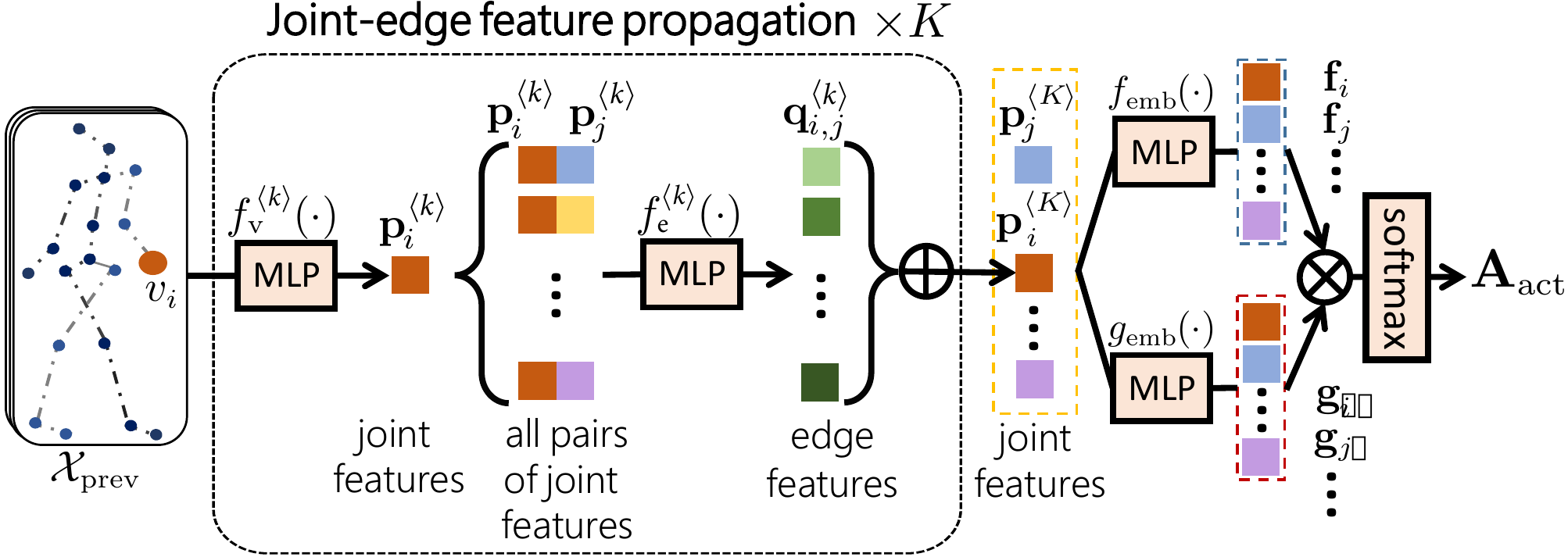}
    \vspace{-10pt}
    \caption{Actional graphs inference module (AGIM) propagates features between joints and edges for $K$ iterations and uses correlations between joint features to obtain actional graphs.}
    \label{fig:AGIM}
    \vspace{-10pt}
\end{figure}

The structure of AGIM is illustrated in Fig.~\ref{fig:AGIM}, where the information is propagated for $K$ times between joints and any possible edges and two embedded joint features are used to calculate actional graphs in the end.

Given the joint-scale actional graphs with adjacent matrix $\mathbf{A}_{\rm act}$, we aggregate the joint information along the action-based relations. We design an actional graph convolution (AGC) to capture the actional features. Mathematically, let the input features at frame $t$ be $\mathbf{X}^{(t)}\in\mathbb{R}^{M\times D_{\bf x}}$, the output features be $\mathbf{Y}^{(t)}_{\rm AGC}\in \mathbb{R}^{M\times D_{\bf x}}$, the AGC works as
\begin{equation}
    \mathbf{Y}^{(t)}_{\rm AGC}={\rm AGC}(\mathbf{X}^{(t)})=\mathbf{A}_{\rm act}\mathbf{X}^{(t)}\mathbf{W}_{\rm act}^{\top}
    \label{eq:agc}
\end{equation}
where $\mathbf{W}_{\rm act}\in\mathbb{R}^{D_{\bf y}\times D_{\bf x}}$ is a trainable weight. Therefore, the model aggregates the action-based information from collaboratively moving joints even in the distance.

\subsubsection{Structural Graph Convolution}
Intuitively, the joint dynamics is limited  due to physical constraints, namely bone connections. To capture these relations, we develop a structural graph, $\mathcal{G}_{\rm str}({V},\mathbf{A}_{\rm str})$. Let ${\bf A}$ is the adjacency matrix of the skeleton graph (see Section~\ref{sec:problem}), the normalized adjacency matrix be  
\begin{equation*}
    {\tilde{\bf A}} = {\bf D}^{-1} {\bf A},
    \label{eq:graph_norm}
\end{equation*}
where  $\mathbf{D}\in\mathbb{N}^{M\times M}$ is  a diagonal degree matrix with $(\mathbf{D})_{ii}=\sum_j(\mathbf{A})_{ij}$. ${\tilde{\bf A}}$ provides nice initialization to learn the edge weights and avoids multiplication explosion~\cite{ilprints361, DBLP:journals/tsp/ChenTFVK18}. 

We note that $\tilde{\bf A}$ only describes the 1-hop neighborhood on body; that is, the bone-connected joints. To represent long-range relations, we use the high-order polynomial of $\tilde{\bf A}$. Let the $\gamma$-order polynomial of $\tilde{\bf A}$ be $\tilde{\bf A}^{\gamma}$, which could be directly computed from the skeleton structure; $\tilde{\bf A}^{\gamma}$ indicates the relations between each joint and its $\gamma$-hop neighbors on skeleton. Given the high-order topologies, we introduce several individual edge-weight matrices $\mathbf{M}^{(\gamma)}\in\mathbb{R}^{M\times M}$ corresponding to $\tilde{\bf A}^{\gamma}$, where each element is trainable to reflect the relation strength. We finally obtain the $\gamma$-order structural graph, whose weighted adjacency matrices is
\begin{equation*}
    \vspace{-2pt}
    \mathbf{A}_{\mathrm{str}}^{(\gamma)} = {\tilde{\bf A}}^{\gamma} \odot \mathbf{M}^{(\gamma)} \in\mathbb{R}^{M\times M},
    \vspace{-2pt}
\end{equation*}
where $\odot$ denotes the element-wise multiplication. In this way, we are able to model the structure-based relations between one joint and others in relatively longer range. Practically, we consider the order $\gamma=1,\dots,\Gamma$, thus we have multiple structural graphs for one body. See plot (c) in Fig.~\ref{fig:link_type}, the hand is correlated with the entire arm.

Given the structural graphs $\mathbf{A}_{\mathrm{str}}^{(\gamma)}$, we propose the structural graph convolution (SGC) operator. Let the input feature at frame $t$ be $\mathbf{X}^{(t)}\in\mathbb{R}^{M\times D_{\bf x}}$, the output feature be $\mathbf{Y}^{(t)}_{\rm SGC}\in\mathbb{R}^{M\times D_{\bf y}}$, the SGC operator is formulated as
\begin{equation}
    \vspace{-2pt}
    \label{eq:sgc}
    \mathbf{Y}^{(t)}_{\rm SGC} = {\rm SGC}(\mathbf{X}^{(t)})=\sum_{\gamma=1}^{\Gamma}\mathbf{A}_{\rm str}^{(\gamma)}\mathbf{X}^{(t)}\mathbf{W}_{\rm str}^{(\gamma)\top}
    \vspace{-2pt}
\end{equation}
where $\mathbf{W}_{\rm str}^{(\gamma)}\in\mathbb{R}^{D_{\bf y}\times D_{\bf x}}$ is the trainable model parameters. Notably, the multiple structural graphs have different corresponding weights, which help to extract richer features.

\subsubsection{Joint-Scale Graph Convolution}

To extract both spatial and temporal features of actions, we now propose the joint-scale graph and temporal convolution block. Based on AGC~\eqref{eq:agc} and SGC~\eqref{eq:sgc}, we present the joint-scale graph convolution (JGC) to capture comprehensive joint-scale spatial features. Mathematically, let the input joint features at frame $t$ be $\mathbf{X}^{(t)}\in\mathbb{R}^{M\times D_{\bf x}}$, the output features be $\mathbf{Y}_{\rm JGC}^{(t)}\in\mathbb{R}^{M\times D_{\bf y}}$, the JGC is formulated as
\begin{equation}
    % \vspace{-2pt}
    \label{eq:ASGC}
    \mathbf{Y}_{\rm JGC}^{(t)} = {\rm JGC}(\mathbf{X}^{(t)}) = \lambda_{\rm act}{\rm AGC}(\mathbf{X}^{(t)})+{\rm SGC}(\mathbf{X}^{(t)}),
    % \vspace{-2pt}
\end{equation}
where $\lambda_{\rm act}$ is a hyper-parameter to trade off the contribution between actional and structural features. Some non-linear activation functions can be applied to it. In this way, the joint features are effectively aggregated to update each center joint according to the joint-scale graphs. 

We further show the stability of the proposed activated joint-scale graph convolution layer; that is, when input 3D skeleton data is disturbed, the distortion of the output features is upper bounded.
\begin{myThm} (Stability)
\label{thm:stability}
{Let two joint-scale feature matrices be $\mathbf{X}$ and $\mathbf{X}^*\in\mathbb{R}^{M\times D_{\bf x}}$ associated with a skeleton graph $\mathbf{A}\in\{0,1\}^{M\times M}$, where $D_{\bf x}=3$ and $\left\| \mathbf{X}^{*} - \mathbf{X} \right\|_F \leq \epsilon$ ($\epsilon\geq0$). } Let $\mathbf{Y}  = \rho \left( {\rm JGC} \left(\mathbf{X} \right) \right)$ and $\mathbf{Y}^{*}  =  \rho \left( {\rm JGC} \left(\mathbf{X}^{*} \right) \right)\in\mathbb{R}^{M\times D_{\bf y}}$.  Let $\mathbf{A}^{*}_{\rm act}$ and $\mathbf{A}_{\rm act}\in[0,1]^{M\times M}$ be the joint-scale actional graph inferred from $\mathbf{X}^{*}$ and $\mathbf{X}$, respectively, where $$\left\| \mathbf{A}^{*}_{\rm act} \mathbf{X}^{*} - \mathbf{A}_{\rm act} \mathbf{X} \right\|_F \leq C \epsilon^q,$$ with $q$ the amplify factor and $C$ some constant. Let  $\mu_{\rm act} = \left\| \mathbf{W}_{\rm act}\right\|_{\max}$,
$\eta^{(\gamma)} = \| \mathbf{M}^{(\gamma)} \|_{\max}$ and  $\mu_{\rm str}^{(\gamma)} = \| \mathbf{W}_{\rm str}^{(\gamma)} \|_{\max}$, where $\mathbf{W}_{\rm act}$, $\mathbf{W}_{\rm str}^{(\gamma)}\in\mathbb{R}^{D_{\bf y}\times D_{\bf x}}$ and $\mathbf{M}^{(\gamma)}\in\mathbb{R}^{M\times M}$.
Then, 
\begin{eqnarray*}
\left\|\mathbf{Y}^{*} - \mathbf{Y}\right\|_F 
& \leq & \sqrt{3 D_{\bf y}} \bigg( \epsilon^q  \lambda_{\rm act} \mu_{\rm act} C  + 
\\
&& \epsilon  \sum_{\gamma=1}^{\Gamma} \sqrt{\left\| {\bf A}^{\gamma} \right\|_0} \eta^{(\gamma)}  \mu_{\rm str}^{(\gamma)} \bigg)
\\
& = & O\left( \max \left( \epsilon^q, \epsilon \right) \right).
\end{eqnarray*}
Note that $\|\cdot\|_F$ denotes Frobenius norm and $\|\cdot\|_0$ is zero norm. $\rho(\cdot)$ denotes ReLU-activation on each element of the data. $O(\cdot)$ denotes the effects that rely on `$\max \left( \epsilon^q, \epsilon \right)$'.
\end{myThm}
See the proof in Appendix. Theorem~\ref{thm:stability} only shows the joint-scale graph convolution at the first layer, but the bound can be extended to the subsequent layers. We make this distinction because the actional graph only depends on the input data. Theorem~\ref{thm:stability} shows that 1) the outputs of JGC followed by the activation function can be upper bounded, reflecting its robustness against perturbation of inputs; 
and 2) given a fixed model, the bound is mainly related to the amplify factor $q$, reflecting how much the actional graph would amplify the perturbation. In the experiments, we show that $q$ is around 1. We also test JGC's robustness against the input perturbation, ensuring that Sym-GNN has stable performance given small noises.

\subsubsection{Joint-Scale Graph and Temporal Convolution Block}
\label{sec:as-gtc}
\begin{figure}[tb]
    \centering
    \includegraphics[width=8.8cm]{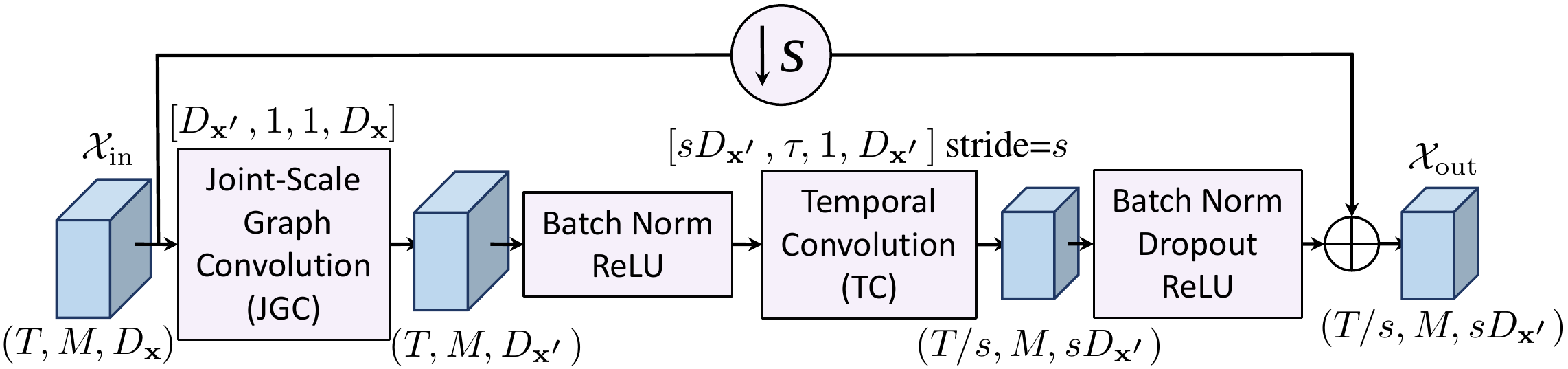}
    \vspace{-10pt}
    \caption{J-GTC block consists of JGC~\eqref{eq:ASGC} and temporal convolution (TC). {The triples below the blocks denote the tensor shapes. The quaternions are the shapes of parameters in JGC and TC operators.}}
    \label{fig:AS-GTC}
    \vspace{-10pt}
\end{figure}

While the JGC operator leverages the joint spatial relations and extracts rich features, we should consider modeling the temporal dependencies among consecutive frames. We develop the temporal convolution operator (TC); that is, a convolution along time to learn the movements. Stacking JGC and TC, we build the \emph{joint-scale graph and temporal convolution block} (J-GTC block), which learn the spatial and temporal features in tandem. 
Mathematically, let $\mathcal{X}_{\rm in} \in \mathbb{R}^{T \times  M \times D_{{\bf x}}} $ be an input tensor, each J-GTC block works as
\begin{subequations}
\begin{align}
\label{eq:encoder_asgc}
 (\mathcal{X}')_t &=  \rho \left( {\rm JGC} ((\mathcal{X}_{\rm in})_t ) \right) \in  \mathbb{R}^{M \times D_{{\bf x}'}}, 
\\
\label{eq:encoder_tc}
 \mathcal{X}_{\rm out} &=   \rho \left(  {\rm TC}(\mathcal{X}' )  \right) \in  \mathbb{R}^{T/s \times  M \times (sD_{{\bf x}'})},
\end{align}
\end{subequations}
where $\rho(\cdot)$ represents a nonlinear ReLU function, ${\rm TC}(\cdot)$ is a standard 1D convolution along the time axis, whose temporal kernel size is $\tau$; $s$ is the convolution stride along time to shrink the temporal dimension; $t$ is the time stamp. In each J-GTC block,~\eqref{eq:encoder_asgc}  extracts spatial features using the multiple spatial relations between joints; and~\eqref{eq:encoder_tc}  extracts temporal features by aggregating the information in several consecutive frames. Our J-GTC also includes batch normalizations and dropout operations. Moreover, there is a residual connection preserving the input features. The architecture of one J-GTC block is illustrated in Fig.~\ref{fig:AS-GTC}.
By stacking several J-GTC blocks in a hierarchy, we could gradually convert the motion dynamics from the sample space to the feature space; that is we capture the high-level semantic information of input sequence for action recognition and motion prediction downstream.

\subsection{Part-Scale Graph Operators}
The joint-scale graphs treat body-joints as nodes and model their relations, but some action patterns depend on more abstract movements of body-parts. For example, `hand waving' shows a rising arm, but the finger and wrist are less important. To model the part dynamics, we propose a part-scale graph and the part-scale graph and temporal convolution (P-GTC) block to extract part-scale features.

\subsubsection{Part-Scale Graph Convolution}
\label{sec:p-gtc}
\begin{figure}[tb]
    \centering
    \includegraphics[width=8.6cm]{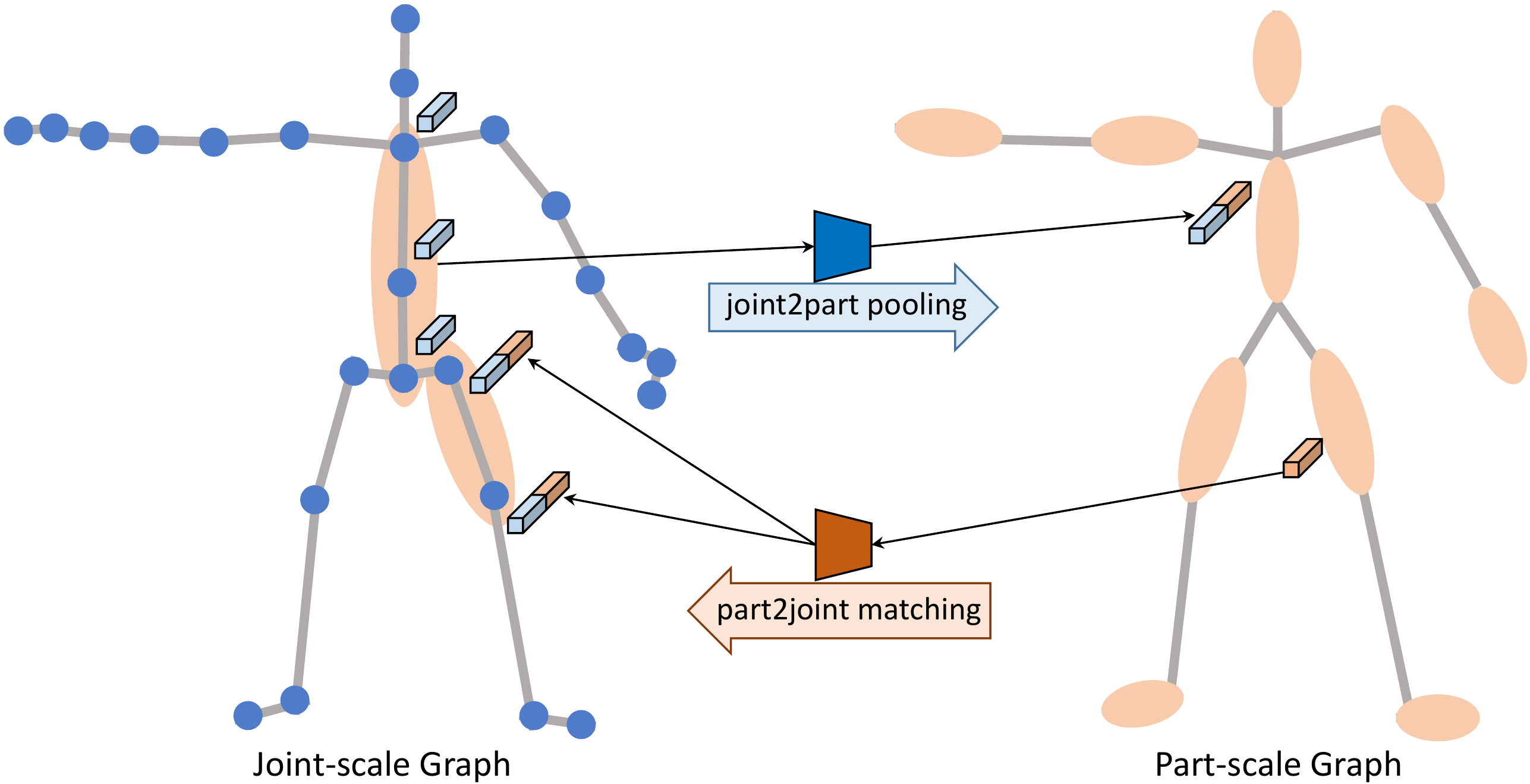}
    \vspace{-10pt}
    \caption{A joint-scale graph consists of body-joints represented as blue nodes and
    a part-scale graph consists of body-parts represented as orange nodes. The bidirectional fusion converts features across two scales through the operations of joint2part pooling and part2joint matching. We only plot the 1-hop structural graph for the joint-scale graph.}
    \label{fig:multi-scale-graph}
    \vspace{-10pt}
\end{figure}

For a part-scale graph, we define $M_{\rm p}=10$ body-parts as graph nodes: `head', `torso', pairs of `upper arms', `forearms', `thighs' and `crura', which integrates the covered joints on joint-scale body. And we build the edges according to body nature. The right plot of Fig.~\ref{fig:multi-scale-graph} shows an example of part-scale graph, whose vertices are $10$ parts and edges are based on nature. The self-looped binary adjacent matrix of part-scale graph is $\mathbf{A}_{\rm p}\in\{0,1\}^{M_{\rm p}\times M_{\rm p}}$, where $(\mathbf{A}_{{\rm p}})_{ij}=1$ if the $i$th and $j$th parts are connected. We normalize $\mathbf{A}_{\rm p}$ by
\begin{equation*}
    \mathbf{A}_{\rm part}=(\mathbf{D}_{\rm p}^{-1}\mathbf{A}_{\rm p})\odot\mathbf{M}_{\rm p}\in\mathbb{R}^{M_{\rm p}\times M_{\rm p}}
\end{equation*}
where ${\bf D}_{\rm p}\in\mathbb{N}^{M_{\rm p}\times M_{\rm p}}$ is the diagonal degree matrix of ${\bf A}_{\rm p}$; ${\bf M}_{\rm p}\in\mathbb{R}^{M_{\rm p}\times M_{\rm p}}$ is a trainable weight matrix and $\odot$ is the element-wise multiplication. 

Similarly to the JGC operator~\eqref{eq:ASGC}, we propose the part-scale graph convolution (PGC) for spatial feature learning. Let the part features at time $t$ be $\mathbf{X}_{\rm p}^{(t)}\in\mathbb{R}^{M_{\rm p}\times D_{\bf x}}$, the  output features be $\mathbf{Y}_{\rm PGC}^{(t)}\in\mathbb{R}^{M\times D_{\bf y}}$, the PGC works as
\begin{equation}
\label{eq:PGC}
\mathbf{Y}_{\rm p}^{(t)} = {\rm PGC}(\mathbf{X}_{\rm p}^{(t)})
                 = \mathbf{A}_{\rm part}\mathbf{X}_{\rm p}^{(t)}\mathbf{W}_{\rm part}^{\top},
\end{equation}
where $\mathbf{W}_{\rm part}^{\top}$ is the trainable parameters. With~\eqref{eq:PGC}, we propagate information between body-parts on the part-scale graph, leading to abstract spatial patterns. Notably, We do not need a part-scale actional graph, because the part-scale graph includes some integrated relations internally, as well as it has a shorter distance to build long-range links.

\subsubsection{Part-Scale Graph and Temporal Convolution Block}
Considering the temporal evolution, we use the same temporal convolution as in J-GTC block to form the \emph{part-scale graph and temporal convolution block} (P-GTC block).
Let the input part feature tensor be $\mathcal{X}_{{\rm p},{\rm in}}\in\mathbb{R}^{T\times M_{\rm p}\times D_{\bf x}}$, we have
\begin{subequations}
\begin{align}
\label{eq:encoder_pgc}
 (\mathcal{X}_{\rm p}')_t &=  \rho \left( {\rm PGC} ((\mathcal{X}_{{\rm p},{\rm in}})_t ) \right) \in  \mathbb{R}^{M_{\rm p} \times D_{{\bf x}'}}, 
\\
\label{eq:encoder_ptc}
 \mathcal{X}_{{\rm p},{\rm out}} &=   \rho \left( {\rm TC}(\mathcal{X}_{\rm p}' )  \right) \in  \mathbb{R}^{T/s \times  M_{\rm p} \times (sD_{{\bf x}'})},
\end{align}
\end{subequations}
where $t$ denotes the time stamp and $s$ is the temporal convolution stride. Comparing to the J-GTC block, the P-GTC block extracts the spatial and temporal  features of actions in a higher scale.

\begin{figure*}[tb]
    \centering
    \includegraphics[width=17.2cm]{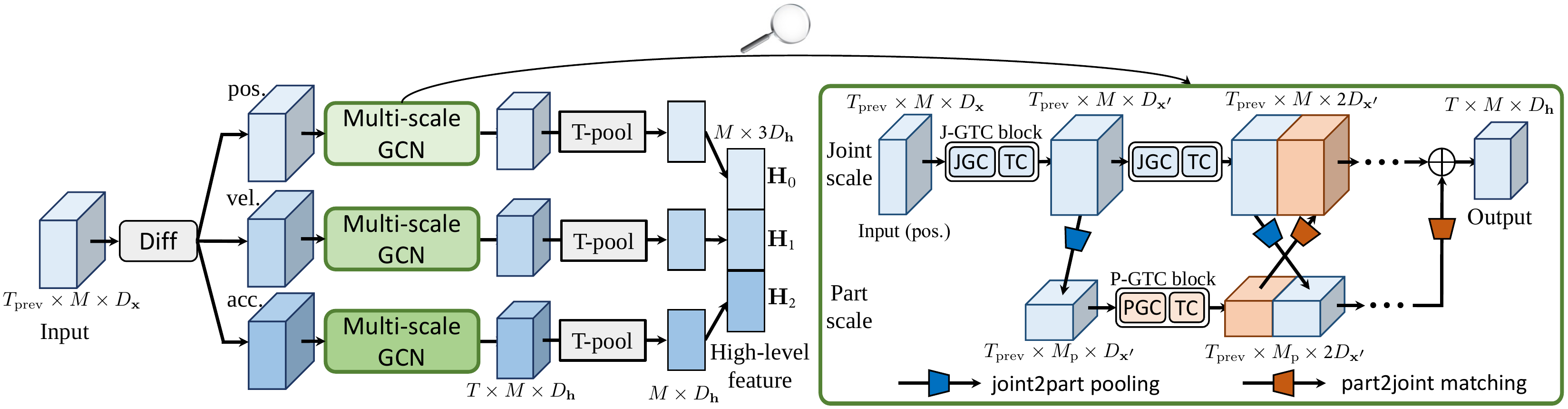}
    \vspace{-10pt}
    \caption{Backbone is essentially multi-branch multi-scale graph convolution networks. It uses three individual multi-scale GCNs to extract spatial and temporal features. A difference operator (`Diff') calculate three orders of differences, which represent joint positions (`pos.'), velocities (`vel.') and accelerations (`acc.'). Each multi-scale GCN takes one order as input and uses multiple J-GTC, P-GTC blocks and bidirectional fusion to learn spatial and temporal features from two scales.}
    \label{fig:backbone}
    \vspace{-10pt}
\end{figure*}

\subsection{Difference Operator}
\label{sec:difference_operator}
Intuitively, the states of motion, such as velocity and acceleration, carry important  dynamics information and make it easier to extract spatial-temporal features. To achieve this, we employ a difference operator to preprocess the input sequences. The idea is to compute high-order differences of the pose sequences, guiding the model to learn motion information more easily. The zero-order difference is $\Delta^{0}\mathbf{X}^{(t)}  = \mathbf{X}^{(t)}  \in \mathbb{R}^{M \times D_{\bf x}}$, where $\mathbf{X}^{(t)}$ is the pose at the time $t$, and the $\beta$-order difference ($\beta>0$) of the pose is
\begin{equation}
    \Delta^{\beta+1}\mathbf{X}^{(t)} = \Delta^{\beta}\mathbf{X}^{(t)}-\Delta^{\beta}\mathbf{X}^{(t-1)}  \in \mathbb{R}^{M \times D_{\bf x}},
    \label{eq:difference}
\end{equation}
where $\Delta^{\beta}$ denotes the $\beta$th-order difference operator. We use zero paddings to handle boundary conditions. We take the first three orders ($\beta=0,1,2$) to our model, reflecting positions, velocities, and accelerations. In the model, the three differences can be efficiently computed in parallel.

\section{Symbiotic Graph Neural Networks}
\label{sec:sym_gnn}
To construct the multitasking model, we need a deep backbone for high-level action pattern extraction as well as two task-specific modules.
In this section, we present the architecture of our Symbiotic Graph Neural Networks (Sym-GNN). First, we present the deep backbone network, which uses multi-scale graphs for feature learning; We then present the action-recognition head and the motion-prediction head with an effective multitasking scheme. Finally, we present a dual network, which learns features from body bones, instead of body-joints, and provides complementary information for downstream tasks.

\subsection{Backbone: Multi-Branch Multi-Scale Graph Convolution Networks} %\Note{SC: I feel you should build the connections between the previous section and this section. Refer the equations or Section labels.}

To learn the high-level action pattern, the proposed Sym-GNN consists of a deep backbone called \emph{multi-branch multi-scale graph convolution network} (multi-branch multi-scale GCN). It employs parallel multi-scale GCN branches to treat high-order action differences for rich dynamics learning and also considers multi-scale graphs for spatial feature extraction. Fig.~\ref{fig:backbone} shows the backbone, where the left plot is the backbone framework including three branches of multi-scale GCNs; the right plot is the structure of each branch of multi-scale GCN. As follows, we propose the backbone architecture in detail.

\subsubsection{Multiple Branches}
The backbone has three branches of multi-scale GCN. Each branch uses a distinct order of action differences as input, treating the motion states for dynamics learning (see Fig.~\ref{fig:backbone}). Calculating the differences by difference operators, we first obtain the proxies of `positions', `velocities' and `accelerations' as the input of the network; see~\eqref{eq:difference}. The three branches have identical network architectures. We obtain semantics of high-order differences and concatenate them together for action recognition and motion prediction.

\subsubsection{Multi-Scale GCN}
\label{sec:multi_scale_gcn}
To learn the detailed and general action features comprehensively,  each branch of the backbone is a multi-scale GCN based on two scales of graphs: \textbf{joint-scale} and \textbf{part-scale} graphs. {For each scale, we use the corresponding operators, i.e. J-GTC blocks (see section~\ref{sec:as-gtc}) and P-GTC block (see section~\ref{sec:p-gtc}) to extract spatial and temporal features.}

Concretely, in the joint scale, the body is modeled by joint-scale graphs, where both the actional graph and structural graphs are used to capture body-joint correlations. This joint scale uses a cascade of {J-GTC blocks} based on the learned actional-structural graphs. In the part scale, we use part-scale graphs whose nodes are body-parts to represent high-level body instances, and we stack multiple P-GTC blocks for feature capturing. To be aware of the multi-scale immediate representations and learn rich and consistent patterns, we introduce a fusion mechanism between the hidden layers of two scales; called~\emph{bidirectional fusion}.

\textbf{Bidirectional Fusion.}
The bidirectional fusion exchanges features from both the joint scale and the part scale; see illustrations in Fig.~\ref{fig:multi-scale-graph} and Fig.~\ref{fig:backbone}. It contains two operations:
\begin{itemize}
    \vspace{-2pt}
    \item {\bf Joint2part pooling.} For the joint scale, we use pooling to average the joint features on the same part to represent a super node. Then, we concatenate the pooling result to the corresponding part feature in the part scale. As shown in Fig.~\ref{fig:multi-scale-graph}, we average torso joints to obtain a node in the part-scale graph and concatenate it to the original part-scale features.
    \item {\bf Part2joint matching.} For the part scale, the part features are copied for several times to  match the number of corresponding joints, as well as we concatenate the copied parts to the joints. As shown in Fig.~\ref{fig:multi-scale-graph}, we copy the thigh twice and concatenate them to the hip and knee in the joint scale.
    \vspace{-2pt}
\end{itemize}
Fig.~\ref{fig:backbone} (right plot) shows the internal operations and one bidirectional fusion in multi-scale GCN. Given the joint-scale input attributes, we first use a J-GTC block to extract the initial joint-scale features, and a joint2part pooling is applied on the joint-scale features to compute the initial part-scale features. We next feed them into two parallel J-GTC and P-GTC blocks. Then we concatenate the responses mapped by joint2part pooling and part2joint matching to the features in opposite scales. Therefore, both scales have good adaptability to multi-scale information. After multiple interactive J-GTC and P-GTC blocks in the multi-scale GCN, we fuse the outputs of two scales through summation, followed by the average pooling to remove the temporal dimension, and obtain the high-level features. 

Finally, we concatenate the outputs from three branches together and use them as the comprehensive semantics for action recognition and motion prediction.

\subsection{Multitasking I: Action Recognition}
For action recognition, Sym-GNN can be represented as $\hat{\mathbf{y}}=\mathcal{F}_{\rm recg}(\mathcal{X}_{\rm prev};{\bm \theta}_{\rm bk}, {\bm \theta}_{\rm recg})$, where $\mathcal{F}_{\rm recg}(\cdot)$ is the recognition sub-model of entire Sym-GNN, $\mathcal{F}(\cdot)$; that is, the recognition performance depends on the backbone network and a recognition module constructed following the backbone. Given the high-level features extracted by three branches of backbone, $\mathbf{H}_0$, $\mathbf{H}_1$ and $\mathbf{H}_2\in\mathbb{R}^{M\times D_{\bf h}}$, we concatenate them and employ an MLP to produce the fused feature:
\begin{equation*}
    \mathbf{H}_{\rm recg}={\rm MLP}—_{\rm recg}\left(\left[\mathbf{H}_{0},\mathbf{H}_{1},\mathbf{H}_{2}\right]\right)\in\mathbb{R}^{M\times D_{\bf h}},
\end{equation*}
where ${\rm MLP}_{\rm recg}(\cdot)$ denotes the fusing network of recognition task and $[\cdot,\cdot,\cdot]$ is the concatenation operator of three matrices along feature dimension.
To integrate the joint dynamics, we apply the global averaging pooling on the $M$ joints of $\mathbf{H}_{\rm recg}$ and obtain a feature vector $\mathbf{h}_{\rm recg}\in\mathbb{R}^{D_{\bf h}}$ that represents the whole body. To generate the recognition results, we finally feed the vector into a 1-layer network with a softmax classifier, obtaining $\hat{\mathbf{y}}\in[0,1]^{C}$.

\subsection{Multitasking II: Motion Prediction}
\begin{figure}[tb]
    \centering
    \includegraphics[width=8.6cm]{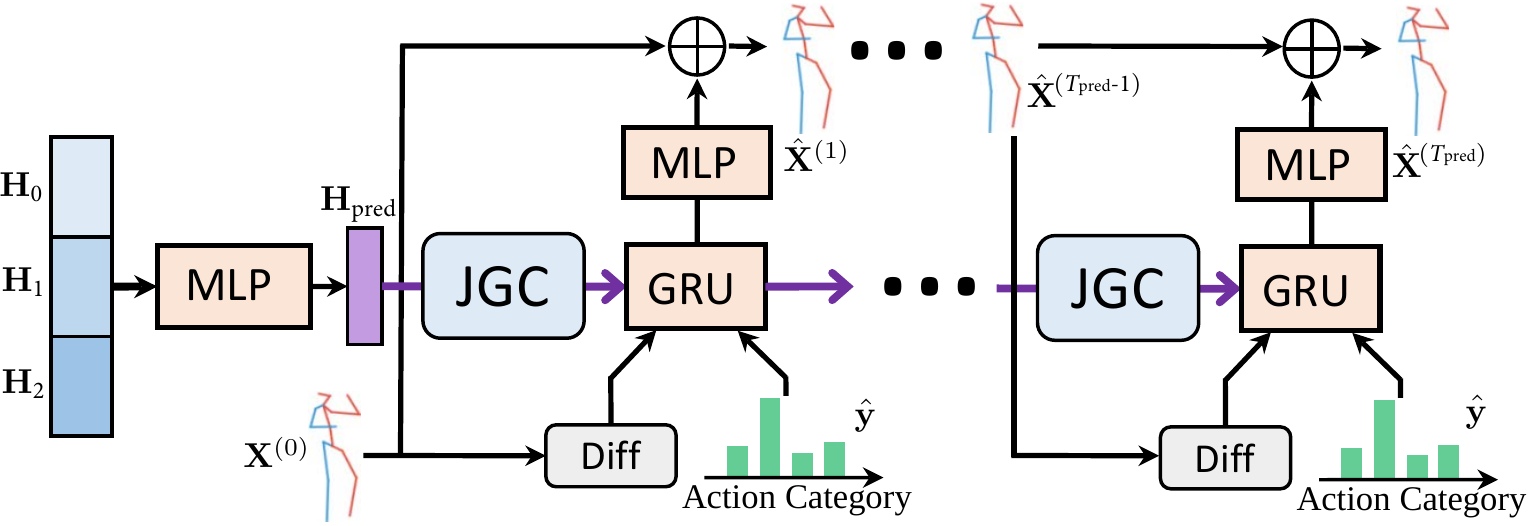}
    \vspace{-10pt}
    \caption{The motion-prediction head of Sym-GNN uses JGC~\eqref{eq:ASGC}, the difference operator~\eqref{eq:difference} and GRU to predict the future poses sequentially.}
    \label{fig:prediction_head}
    \vspace{-10pt}
\end{figure}

For motion preidction, our Sym-GNN works as $\hat{\mathbf{X}}_{\rm pred}=\mathcal{F}_{\rm pred}(\mathcal{X}_{\rm prev};{\bm \theta}_{\rm bk}, {\bm \theta}_{\rm pred})$, using the backbone and a prediction module, where $\mathcal{F}_{\rm pred}(\cdot)$ is the prediction sub-model of Sym-GNN. Therefore, we additionally build a motion-prediction head, whose functionality is to sequentially predict the future poses. Fig.~\ref{fig:prediction_head} shows the overall structure. We adopt the self-regressive mechanics and identical connection in the motion-prediction head, which utilizes gated recurrent unit (GRU) to model the temporal evolution. 

Concretely, an MLP is first employed to embed the features of three action differences:
\begin{equation*}
    \mathbf{H}_{\rm pred}={\rm MLP}_{\rm pred}(\left[\mathbf{H}_{0},\mathbf{H}_{1},\mathbf{H}_{2}\right])\in\mathbb{R}^{M \times D_{\bf h}},
\end{equation*}
where ${\rm MLP}_{\rm pred}(\cdot)$ denotes the fusing network of prediction task.
Let $\mathbf{H}_{\rm pred}^{(0)} = \mathbf{H}_{\rm pred}$ be the initial states of GRU-based predictor and $\hat{\mathbf{X}}^{(0)}=\mathbf{X}^{(0)}$ be the pose in the current time stamp. To produce the $(t+1)$th pose $(t\geq0)$, the motion-prediction head works as
\begin{subequations}
\begin{align}
\label{eq:decoder_asgc}
   \widetilde{\mathbf{H}}_{\rm pred}^{(t)} &= {\rm JGC}(\mathbf{H}^{(t)}_{\rm pred}),
    \\
\label{eq:decoder_gru}
    \mathbf{H}^{(t+1)}_{\rm pred} &= {\rm GRU}( 
[ \hat{\mathbf{X}}^{(t)}, \Delta^1 \hat{\mathbf{X}}^{(t)}, \Delta^2  \hat{\mathbf{X}}^{(t)}, \hat{\mathbf{y}}], \widetilde{\mathbf{H}}^{(t)}_{\rm pred}),
    \\
\label{eq:decoder_residual}
    \hat{\mathbf{X}}^{(t+1)} &= \hat{\mathbf{X}}^{(t)} + f_{\rm pred}(\mathbf{H}^{(t+1)}_{\rm pred} ),
        \vspace{-4mm}
\end{align}
\end{subequations}
where ${\rm JGC}(\cdot)$, ${\rm GRU}(\cdot)$ and $f_{\rm pred}(\cdot)$ represent JGC operator~\eqref{eq:ASGC}, GRU cell and output MLP, respectively. The following $\hat{\mathbf{X}}^{(t)}$s are the predictions obtained sequentially and used recylingly. In Step~\eqref{eq:decoder_asgc}, we apply the JGC to update the hidden states; in Step~\eqref{eq:decoder_gru},  we feed the updated hidden states, current pose and classified labels into the GRU cell to produce the features that reflect future displacement;  In Step~\eqref{eq:decoder_residual}, we  add the predicted displacement to the previous pose to predict the next frame.

The motion-prediction head has three advantages: (i) we use JGC to update hidden features, capturing more complicated motion patterns; (ii) we input multiple orders of poses differences and classified labels to the GRU, providing explicit motion priors; and (iii) Connected by the residual, the GRU and MLP predict the displacement for each frame; this makes predictions precise and robust.

\subsection{Multi-Objective Optimization}
To train action recognition and motion prediction simultaneously, we consider a multi-objective scheme.

To recognize actions, we minimize the cross entropy between the ground-truth categorical labels and the inferred ones. Let the true label of the $n$th sample be $(\mathbf{y})_n\in\{0,1\}^C$ and the corresponding classification results be $(\hat{\mathbf{y}})_n\in\{0,1\}^C$. For $N$ training samples in one mini-batch, the action recognition loss is formulated as
\begin{equation}
\mathcal{L}_{\rm recg} = -\frac{1}{N}\sum_{n=1}^{N}(\mathbf{y})^{\top}_{n}\log(\hat{\bf y})_n,
\label{eq:loss_recg}
\end{equation}
where $\top$ denotes the transpose operation.

For motion prediction, we minimize the $\ell_1$ distance between the target motions and the predicted clips. Let the $n$th target and predictions be $(\mathcal{X}_{\rm pred})_n$ and $(\hat{\mathcal{X}}_{\rm pred})_n$, for $N$ samples in one mini-batch, the prediction loss is
\begin{equation}
\mathcal{L}_{\rm pred} = \frac{1}{N}\sum_{n=1}^{N}\|(\mathcal{X}_{\rm pred})_n-(\hat{\mathcal{X}}_{\rm pred})_n\|_1,
\label{eq:loss_pred}
\end{equation}
where $\|\cdot\|_1$ denotes the $\ell_1$ norm. According to our experiments, the $\ell_1$ norm leads to more precise predictions compared to the common $\ell_2$ norm. 

To integrate two losses for training, we propose a convex combination that weighted sums~\eqref{eq:loss_recg} and~~\eqref{eq:loss_pred}; that is 
\begin{equation*}
    \mathcal{L}=\lambda\mathcal{L}_{\rm recg}+(1-\lambda)\mathcal{L}_{\rm pred},
\end{equation*}
where $\lambda$ trade-offs the importances of two tasks. To balance action recognition and motion prediction, instead of using a fixed $\lambda$ selected by hand, we employ the `multiple-gradient descent algorithm' (MGDA) to obtain the proper coefficient $\lambda$ for multitasking loss terms. Following~\cite{MGDA}, the parameter $\lambda$ is adaptively adjusted  during training by matching one of Karush-Kuhn-Tucker (KKT) conditions of the optimization problem. Therefore, the optimized $\lambda$ allocates weights for the two tasks adaptively, leading to better performances for both tasks.
In our training scheme, all the model parameters are trained end-to-end with the stochastic gradient descent algorithm~\cite{SGD}; see more details in Appendix.

% \Note{SC: replace Bone-scale by Bone-based. You can say very compactly: we input bone features; then everything else follows the skeleton-based graph. No need to dive into details}
\subsection{Bone-based Dual Graph Neural Networks}
% \Note{SC: The terminology is a big issue now. You have joint-scale graph, part-scale graph, joint-based graph, bone-based graph, which could be confusing. I suggest instead of using bone-based graph, you call it dual graph: you would have joint-scale graph, part-scale graph, bone-scale dual graph, part-scale dual graph.}
While the joints contain some information of action representation from the joint aspect, the attributes of bones, such as lengths and orientations, are crucial to provide some complementary information. In this section, we construct a bone-based dual graph against original joint-scale graph, whose vertices are bones and edges link bones.

To represent the feature of each bone, we compute the subtraction of two endpoint joints coordinates, which includes information of bone lengths and orientations. The subtraction order is from the centrifugal joint $v_j$ to the centripetal $v_i$. Let the joint locations along time be $\mathbf{x}_i, \mathbf{x}_j\in\mathbb{R}^{D_{\bf x}T_{\rm prev}}$, the bone attribute is $\mathbf{b}_{i,j}=\mathbf{x}_j-\mathbf{x}_i \in\mathbb{R}^{D_{\bf x}T_{\rm prev}}$. Then, we construct the bone-based dual actional and structural graphs to model the bone relations; and we also build the part-scale dual graph. The dual actional graph is learned from bone features by AGIM (see section~\ref{sec:joint-scale-graph}); for dual structural graph, the 1-hop edges are linked when two bones with articulated joints and the high-hop edges are extended from the 1-hop edges; the part-scale attributes are obtained by integrating bone attributes and the part-scale graph is built according to body nature; The bone-based graphs are dual of joint-scale graphs, which are employed to extract complementary bone features. 

Given the bone-based graphs, we train a \emph{bone-based graph neural network}. We input bone attributes; then everything else follows the joint-based network. Finally, we fuse the joint-based and bone-based recognition outputs before softmax functions by a weighted summation to calculate the classification results, which tend to be more accurate and improve motion prediction effectively.

\section{Experiments and Analysis}
\label{sec:exp}
In this section, we evaluate the proposed Sym-GNN. First, we introduce the datasets and model settings in detail; then, the performance comparisons between Sym-GNN and other state-of-the-art methods are presented; and we finally show the ablation studies of our model.

\subsection{Datasets and Model Setting}

\subsubsection{Dataset}
We conduct extensive experiments on four large-scale  datasets: NTU-RGB+D~\cite{Shahroudy_2016_CVPR}, Kinetics~\cite{AAAI1817135}, Human 3.6M~\cite{h36m} and CMU Mocap. The details is shown as follow.

{\bf NTU-RGB+D:} NTU-RGB+D, containing $56,880$ skeleton action sequences completed by one or two performers and categorized into $60$ classes, is one of the largest  datasets for 3D skeleton-based action recognition. It provides the 3D spatial coordinates of $25$ joints for each subject in an action. For method evaluation, two protocols are recommended: `Cross-Subject' (CS) and `Cross-View' (CV). In CS, $40,320$ samples performed by $20$ subjects are separated into the training set, and the rest belong to the test set. CV assigns data according to camera views, where training and test set have $37,920$ and $18,960$ samples, respectively.

{\bf Kinetics:} Kinetics is a large  dataset for human action analysis, containing over $240,000$ video clips. There are $400$ classes of actions. Due to only RGB videos, we obtain skeleton data by estimating joint locations on pixels with OpenPose toolbox~\cite{Cao_2017_CVPR}. The toolbox generates 2D pixel coordinates $(x, y)$ and confidence score $c$ for totally $18$ joints. We represent each joint as a three-element feature vector: $[x, y, c]^{\top}$. For the multiple-person cases, we select the body with the highest average joint confidence in each sequence. Therefore, one clip with $T$ frames is transformed into a skeleton sequence with the dimension of $18 \times 3 \times T$.

{\bf Human 3.6M:} Human 3.6M (H3.6M) is a large motion capture dataset and also receives increasing popularity. Seven subjects are performing $15$ classes of actions, where each subject has $32$ joints. We downsample all sequences by two. The models are trained on six subjects and tested on the specific clips of the $5$th subject. Notably, the dataset provides the joint locations in angle space, and we transform them into exponential maps and only use the joints with non-zero values (actually $21$ joints).

{\bf CMU Mocap:} CMU Mocap includes five major action categories, and each subject in CMU Mocap has $38$ joints, which are presented by angle positions. We use the same strategy presented in~\cite{Li_2018_CVPR} to select the actions. Thus we choose eight actions: `Basketball', `Basketball Signal', `Directing Traffic', `Jumping', `Running', `Soccer', `Walking' and `Washing Window'. We preprocess the data and compute the corresponding exponential maps with the same approach as we do for Human 3.6M dataset.

\subsubsection{Model Setting and Implementation Details}
The models are implemented with PyTorch 0.4.1.
Since different datasets have distinctive patterns and complexities, we employ specific configurations of Sym-GNN networks on corresponding datasets for features learning.

For NTU-RGB+D and Kinetics, the backbone network of Sym-GNN contains $9$ J-GTC blocks and $8$ P-GTC blocks. In each three J-GTC and P-GTC blocks, the feature dimensions are respectively $64$, $128$ and $256$. The kernel size of TC is $9$ and it shrinks the temporal dimension with stride $2$ after the $3$rd and $6$th blocks, where we use bidirectional fusion mechanisms. $\lambda_{\rm act}=0.5$. The action-recognition head is a $2$-layer MLP, whose hidden dimension is $256$. For the motion-prediction head, the hidden dimensions of GRU and output MLP are $256$. For the actional graph inference module (AGIM), we use $2$-layer $128$-D MLPs with ReLU, batch normalization and dropout in each iteration. We use SGD algorithm to train Sym-GNN, where the learning rate is initially $0.1$ and decays by $10$ every $30$ epochs. The model is trained with batch size $48$ for $100$ epochs on $8$ GTX-1080Ti GPUs. For both NTU-RGB+D and Kinetics, the last 10 frames are used for motion prediction and other previous frames are fed into Sym-GNN for action recognition.

As for Human 3.6M and CMU Mocap, due to the simpler dynamics and fewer categories, we propose a light version of Sym-GNN, which extracts meaningful features with more shallow networks, improving efficiency for motion prediction. In the backbone, we use $4$ J-GTC blocks and $3$ P-GTC blocks, whose feature dimensions are $32$, $64$, $128$ and $256$; the temporal convolution strides in $4$ blocks are: $1$, $2$, $2$, $2$, respectively. We apply bidirectional fusions at the last 3 layers. $\lambda_{\rm act}=1.0$. The recognition and motion-prediction heads, as well as AGIM, leverage the same architecture as we set for NTU-RGB+D. We train the model using Adam optimizer with the learning rate $1\times 10^{−4}$ and batch size $64$ for $10^5$ iterations on one GTX-1080Ti GPU. All the hyper-parameters are selected using a validation set.

\subsection{Comparison with State-of-the-Arts}
On the three large-scale skeleton-formed datasets, we compare the proposed Sym-GNN with state-of-the-art methods for human action recognition and motion prediction.

\subsubsection{3D Skeleton-based Action Recognition}
For action recognition, we first show the classification accuracies of Sym-GNN and baselines on two recommended benchmarks of NTU-RGB+D, i.e. Cross-Subject and Cross-View~\cite{Shahroudy_2016_CVPR}. The state-of-the-art models are based on manifold analysis~\cite{Vemulapalli_2014_CVPR}, recurrent neural networks~\cite{Du_2015_CVPR,Shahroudy_2016_CVPR,Si_2018_ECCV}, convolution networks~\cite{a8014941,vis_cnn,ijcai_ChaoLi}, and graph networks~\cite{AAAI1817135,Tang_2018_CVPR,Si_2018_ECCV,2sAGCN,Si_2019_CVPR,DGNN,STGR-GCN,Li_cvpr_2019,motif-GCN}. Moreover, to investigate different components of Sym-GNN, such as multiple graphs and multitasking, we test several model variants, including Sym-GNN using only joint-scale structural graphs (Only J-S), only joint-scale actional graphs (Only J-A), only-part scale graph (Only P), no bone-based dual graphs (No bone), no prediction for multitasking (No pred) and complete model. Table~\ref{tab:recg_ntu} presents recognition accuracies of methods.
\begin{table}[tb]
    \centering
    \caption{Comparison of action recognition on NTU-RGB+D. The accuracies on both Cross-Subject (CS) and Cross-View (CV) benchmarks.}
    \vspace{-10pt}
    \setlength{\tabcolsep}{5mm}{
    \begin{tabular}{c|cc}
        \hline
        Methods & CS & CV \\
        \hline
        Lie Group~\cite{Vemulapalli_2014_CVPR} & 50.1\% & 52.8\% \\
        H-RNN~\cite{Du_2015_CVPR} & 59.1\% & 64.0\% \\
        Deep LSTM~\cite{Shahroudy_2016_CVPR} & 60.7\% & 67.3\% \\
        PA-LSTM~\cite{Shahroudy_2016_CVPR} & 62.9\% & 70.3\% \\
        ST-LSTM+TS~\cite{Liu_2016_eccv} & 69.2\% & 77.7\% \\
        Temporal Conv~\cite{a8014941} & 74.3\% & 83.1\% \\
        Visualize CNN~\cite{vis_cnn} & 76.0\% & 82.6\% \\
        C-CNN+MTLN & 79.6\% & 84.8\% \\
        ST-GCN~\cite{AAAI1817135} & 81.5\% & 88.3\% \\
        DPRL~\cite{Tang_2018_CVPR} & 83.5\% & 89.8\% \\
        SR-TSL~\cite{Si_2018_ECCV} & 84.8\% & 92.4\% \\
        HCN~\cite{ijcai_ChaoLi} & 86.5\% & 91.1\% \\
        STGR-GCN~\cite{STGR-GCN} & 86.9\% & 92.3\% \\
        motif-GCN~\cite{motif-GCN} & 84.2\% & 90.2\% \\
        AS-GCN~\cite{Li_cvpr_2019} & 86.8\% & 94.2\% \\
        2s-AGCN~\cite{2sAGCN} & 88.5\% & 95.1\% \\
        AGC-LSTM~\cite{Si_2019_CVPR} & 89.2\% & 95.0\% \\
        DGNN~\cite{DGNN} & 89.9\% & 96.1\% \\
        \hline
        Sym-GNN (Only J-S) & 88.3\% & 94.5\% \\
        Sym-GNN (Only J-A) & 85.7\% & 93.7\% \\
        Sym-GNN (Only P) & 86.5\% & 87.3\% \\
        Sym-GNN (No bone) & 87.1\% & 93.8\% \\
        Sym-GNN (No pred) & 89.0\% & 95.7\% \\
        Sym-GNN & {\bf90.1\%} & {\bf96.4\%} \\
        \hline
    \end{tabular}}
    \label{tab:recg_ntu}
    \vspace{-10pt}
\end{table}
\begin{table}[tb]
    \centering
    \caption{Comparison of action recognition on Kinetics. The top-1 and top-5 classification accuracies are listed.}
    \vspace{-10pt}
    \setlength{\tabcolsep}{5.5mm}{
    \begin{tabular}{c|cc}
        \hline
        Methods & Top-1 & Top-5\\
        \hline
        Feature Encoding~\cite{Fernando_2015_CVPR} & 14.9\% & 25.8\%\\
        Deep LSTM~\cite{Shahroudy_2016_CVPR} & 16.4\% & 35.3\%\\
        Temporal Conv~\cite{a8014941} & 20.3\% & 40.0\%\\
        ST-GCN~\cite{AAAI1817135} & 30.7\% & 52.8\%\\
        STGR-GCN~\cite{STGR-GCN} & 33.6\% & 56.1\% \\
        AS-GCN~\cite{Li_cvpr_2019} & 34.8\% & 56.3\% \\
        2s-AGCN~\cite{2sAGCN} & 36.1\% & {\bf 58.7\%}\\
        DGNN~\cite{DGNN} & 36.9\% & 56.9\%\\ 
        \hline
        Sym-GNN (No pred) & 36.4\% & 57.4\%\\
        Sym-GNN & {\bf37.2\%} & {58.1\%}\\
        \hline
    \end{tabular}}
    % \vspace{-10pt}
    \label{tab:recg_kinetics}
\end{table}
\begin{table}[tb]
    \centering
    \vspace{-10pt}
    \caption{Comparison of action recognition on Human 3.6M and CMU Mocap dataset. The top-1 and top-5 classification accuracies are listed.}
    \vspace{-10pt}
    \begin{tabular}{c|cc|cc}
        \hline
        ~ & \multicolumn{2}{|c|}{Human 3.6M} & \multicolumn{2}{|c}{CMU Mocap} \\
        \hline
        Methods & Top-1 & Top-5 & Top-1 & Top-5\\
        \hline
        ST-GCN~\cite{AAAI1817135} & 40.2\% & 78.4\% & 87.5\% & 96.9\%\\
        HCN~\cite{ijcai_ChaoLi} & 47.6\% & 88.8\% & 95.4\% & 99.2\% \\
        2s-AGCN~\cite{2sAGCN} & 55.4\% & 94.1\% & 97.1\% & 99.8\% \\
        \hline
        Sym-GNN (Only J) & 55.6\% & 93.9\% & 96.5\% & 99.4\%\\
        Sym-GNN (Only P) & 54.3\% & 93.1\% & 94.9\% & 98.0\%\\
        Sym-GNN (No bone) & 53.5\% & 93.2\% & 93.5\% & 95.8\%\\
        Sym-GNN (No pred) & 55.2\% & 94.1\% & 96.6\% & 99.4\%\\
        Sym-GNN & {\bf56.5\%} & {\bf95.3\%} & {\bf98.8\%} & {\bf100\%}\\
        \hline
    \end{tabular}
    \vspace{-10pt}
    \label{tab:recg_h36m_cmu}
\end{table}
We see that the complete Sym-GNN, which utilizes joint-scale and part-scale graphs, motion motion-prediction head and dual bone-based network, outperforms the baselines on both benchmarks. The results reveal that richer joint relations promote to capture more useful patterns, and additional motion prediction and complementary bone-based features improve the discrimination.
\begin{table*}[htb]
    \centering
    \caption{Comparison of PCK@0.05 (\%) between Sym-GNN and state-of-the-art methods for short-term motion prediction on NTU-RGB+D. The variant of Sym-GNN (No recg) denotes our model without using the recognition task to enhance motion prediction.}
    \vspace{-10pt}
    \small
    \setlength{\tabcolsep}{1.7mm}{

        \begin{tabular}{c|c|cccccccccc|c}
        \hline
        Benchmarks & Future frames & 1 & 2 & 3 & 4 & 5 & 6 & 7 & 8 & 9 & 10 & Average \\
        \hline
        \multirow{6}{*}{Cross-Subject} & ZeroV~\cite{Martinez_2017_CVPR} & 70.32 & 52.77 & 32.64 & 23.19 & 18.05 & 14.11 & 11.92 & 9.84 & 8.35 & 6.04 & 24.72\\
        ~ & Res-sup~\cite{Martinez_2017_CVPR} & 76.25 & 55.96 & 40.31 & 29.47 & 22.81 & 16.96 & 13.65 & 11.57 & 10.13 & 8.87 & 28.60\\
        ~ & CSM~\cite{Li_2018_CVPR} & 82.38 & 68.11 & 56.84 & 45.26 & 38.65 & 30.41 & 26.15 & 22.74 & 17.52 & 15.96 & 40.40\\
        ~ & Skel-TNet~\cite{AAAI_Guo} & 93.62 & 86.44 & 81.03 & 75.85 & 70.81 & 66.57 & 59.60 & 54.45 & 46.92 & 40.18 & 67.55\\
        \cline{2-13}
        ~ & Sym-GNN (No recg) & 98.74 & 97.07 & 94.95 & 93.94 & {\bf93.48} & 91.79 & 90.69 & 89.27 & 87.87 & 86.01 & 92.34\\
        ~ & Sym-GNN & {\bf99.00} & {\bf97.65} & {\bf95.89} & {\bf95.10} & {93.44} & {\bf92.70} & {\bf91.75} & {\bf90.65} & {\bf89.54} & {\bf89.10} & {\bf93.48}\\
        \hline
        \hline
        \multirow{6}{*}{Cross-View} & ZeroV~\cite{Martinez_2017_CVPR} & 75.69 & 55.72 & 39.88 & 29.60 & 21.91 & 15.23 & 12.06 & 10.18 & 8.70 & 7.33 & 27.63\\
        ~ & Res-sup~\cite{Martinez_2017_CVPR} & 78.85 & 59.91 & 43.82 & 32.37 & 24.32 & 18.51 & 14.86 & 12.29 & 10.38 & 8.85 & 30.42\\
        ~ & CSM~\cite{Li_2018_CVPR} & 85.41 & 71.75 & 58.20 & 46.69 & 39.07 & 31.85 & 28.43 & 24.17 & 19.66 & 18.93 & 42.62\\
        ~ & Skel-TNet~\cite{AAAI_Guo} & 94.81 & 89.12 & 83.85 & 79.00 & 72.74 & 69.11 & 62.39 & 66.97 & 48.88 & 42.70 & 70.66\\
        \cline{2-13}
        ~ & Sym-GNN (No recg) & 98.99 & 96.79 & 95.55 & 94.68 & 93.03 & 92.13 & 90.88 & 89.69 & 88.70 & 87.57 & 92.79\\
        ~ & Sym-GNN & {\bf99.25} & {\bf97.87} & {\bf96.38} & {\bf95.21} & {\bf94.06} & {\bf92.95} & {\bf91.94} & {\bf91.01} & {\bf90.18} & {\bf89.27} & {\bf93.81}\\
        \hline
        \end{tabular}}

    \label{tab:pred_ntu}
    \vspace{-10pt}
\end{table*}
\begin{table*}[htb]
    \centering
    \caption{Comparisons of MAEs between Sym-GNN and state-of-the-art methods for short-term motion prediction on the 4 representative actions of H3.6M. Sym-GNN (J-A) and Sym-GNN (J-S) are Sym-GNN with joint-scale actional graphs only and with joint-scale structural graph only, respectively.  Sym-GNN (No recg) represents the model trained without action classification.}
    \vspace{-10pt}
    \footnotesize
    \setlength{\tabcolsep}{1.9mm}{

        \begin{tabular}{c|cccc|cccc|cccc|cccc}
        \hline
        Motion & \multicolumn{4}{|c|}{Walking} & \multicolumn{4}{|c|}{Eating} & \multicolumn{4}{|c|}{Smoking} & \multicolumn{4}{|c}{Discussion}\\
        \hline
        milliseconds & 80&160&320&400 & 80&160&320&400 & 80&160&320&400 & 80&160&320&400 \\
        \hline

        ZeroV~\cite{Martinez_2017_CVPR} & 0.39 & 0.68 & 0.99 & 1.15 & 0.27 & 0.48 & 0.73 & 0.86 & 0.26 & 0.48 & 0.97 & 0.95 & 0.31 & 0.67 & 0.94 & 1.04 \\
        ERD~\cite{Fragkiadaki_2015_ICCV} & 0.93 & 1.18 & 1.59 & 1.78 & 1.27 & 1.45 & 1.66 & 1.80 & 1.66 & 1.95 & 2.35 & 2.42 & 2.27 & 2.47 & 2.68 & 2.76 \\
        Lstm3LR~\cite{Fragkiadaki_2015_ICCV} & 0.77 & 1.00 & 1.29 & 1.47 & 0.89 & 1.09 & 1.35 & 1.46 & 1.34 & 1.65 & 2.04 & 2.16 & 1.88 & 2.12 & 2.25 & 2.23 \\
        SRNN~\cite{Jain_2016_CVPR} & 0.81 & 0.94 & 1.16 & 1.30 & 0.97 & 1.14 & 1.35 & 1.46 & 1.45 & 1.68 & 1.94 & 2.08 & 1.22 & 1.49 & 1.83 & 1.93 \\
        DropAE~\cite{GhoshSAH17} & 1.00 & 1.11 & 1.39 & / & 1.31 & 1.49 & 1.86 & / & 0.92 & 1.03 & 1.15 & / & 1.11 & 1.20 & 1.38 & / \\
        Samp-loss~\cite{Martinez_2017_CVPR} & 0.92 & 0.98 & 1.02 & 1.20 & 0.98 & 0.99 & 1.18 & 1.31 & 1.38 & 1.39 & 1.56 & 1.65 & 1.78 & 1.80 & 1.83 & 1.90 \\
        Res-sup~\cite{Martinez_2017_CVPR} & 0.27 & 0.46 & 0.67 & 0.75 & 0.23 & 0.37 & 0.59 & 0.73 & 0.32 & 0.59 & 1.01 & 1.10 & 0.30 & 0.67 & 0.98 & 1.06 \\
        CSM~\cite{Li_2018_CVPR} & 0.33 & 0.54 & 0.68 & 0.73 & 0.22 & 0.36 & 0.58 & 0.71 & 0.26 & 0.49 & 0.96 & 0.92 & 0.32 & 0.67 & 0.94 & 1.01 \\
        TP-RNN~\cite{abs-1810-09676} & 0.25 & 0.41 & 0.58 & 0.65 & 0.20 & 0.33 & 0.53 & 0.67 & 0.26 & 0.47 & 0.88 & 0.90 & 0.30 & 0.66 & 0.96 & 1.04 \\
        QuaterNet~\cite{quater} & 0.21 & 0.34 & 0.56 & 0.62 & 0.20 & 0.35 & 0.58 & 0.70 & 0.25 & 0.47 & 0.93 & 0.90 & 0.26 & 0.60 & 0.85 & 0.93\\
        AGED~\cite{Gui_2018_ECCV} & 0.21 & 0.35 & 0.55 & 0.64 & 0.18 & {\bf 0.28} & 0.50 & 0.63 & 0.27 & 0.43 & 0.81 & 0.83 & 0.26 & {0.56} & {\bf0.77} & {\bf0.84} \\
        BiHMP-GAN~\cite{AAAI_Kundu} & 0.33 & 0.52 & 0.63 & 0.67 & 0.20 & 0.33 & 0.54 & 0.70 & 0.26 & 0.50 & 0.91 & 0.86 & 0.33 & 0.65 & 0.91 & 0.95 \\
        Skel-TNet~\cite{AAAI_Guo} & 0.31 & 0.50 & 0.69 & 0.76 & 0.20 & 0.31 & 0.53 & 0.69 & 0.25 & 0.50 & 0.93 & 0.89 & 0.30 & 0.64 & 0.89 & 0.98 \\
        VGRU-r1~\cite{Gopalakrishnan_2019_CVPR} & 0.34 & 0.47 & 0.64 & 0.72 & 0.27 & 0.40 & 0.64 & 0.79 & 0.36 & 0.61 & 0.85 & 0.92 & 0.46 & 0.82 & 0.95 & 1.21 \\
        \hline
        Sym-GNN (Only J-A) & 0.19 & 0.35 & 0.54 & 0.63 & 0.18 & 0.34 & 0.54 & 0.66 & 0.23 & 0.43 & 0.84 & 0.82 & 0.26 & 0.62 & 0.81 & 0.87 \\
        Sym-GNN (Only J-S) & 0.19 & 0.33 & 0.54 & 0.69 & 0.17 & 0.32 & 0.52 & 0.66 & 0.21 & 0.41 & 0.83 & 0.82 & 0.24 & 0.64 & 0.93 & 1.01 \\
        Sym-GNN (No recg) & 0.18 & {\bf0.31} & {\bf0.50} & {\bf0.59} & {\bf0.16} & {0.29} & 0.49 & 0.61 & {\bf0.21} & {\bf0.40} & 0.80 & {\bf0.80} & {0.22} & {0.57} & {0.85} & {0.93} \\
        Sym-GNN & {\bf0.17} & {\bf0.31} & {\bf0.50} & {0.60} & {\bf0.16} & {0.29} & {\bf0.48} & {\bf0.60} & {\bf0.21} & {\bf0.40} & {\bf0.76} & {\bf0.80} & {\bf0.21} & {\bf0.55} & {\bf0.77} & 0.85 \\
        \hline
        \end{tabular}}
        \vspace{-10pt}
    \label{tab:pred_h36m_4}
\end{table*}

Then, we evaluate our model for action recognition on Kinetics and compare it with six previous models, including a hand-crafted based method, Feature Encoding~\cite{Fernando_2015_CVPR}, two deep models, Deep LTSM~\cite{Shahroudy_2016_CVPR} and Temporal ConvNet~\cite{a8014941}, and three graph-based methods, ST-GCN~\cite{AAAI1817135}, 2s-AGCN~\cite{2sAGCN}, and DGNN~\cite{DGNN}. Table~\ref{tab:recg_kinetics} shows the top-1 and top-5 classification results, and `no pred' denotes the Sym-GNN variant without motion-prediction head and multitasking framework. We see that Sym-GNN outperforms other methods on top-1 recognition accuracy and achieves competitive results on top-5 recognition accuracy.

Additionally, we evaluate our model for action recognition on Human 3.6M and CMU Mocap. Table~\ref{tab:recg_h36m_cmu} presents the top-1 and top-5 classification accuracies for both two datasets. Here we compare Sym-GNN with a few recently proposed methods: ST-GCN~\cite{AAAI1817135}, HCN~\cite{ijcai_ChaoLi}, and 2s-AGCN~\cite{2sAGCN}. We also show the effectiveness of our model.
\begin{table*}[htb]
    \centering
    \caption{Comparisons of MAEs between Sym-GNN and previous methods for short-term motion prediction on other $11$ actions of H3.6M dataset.}
    \vspace{-10pt}
    \footnotesize
    \setlength{\tabcolsep}{0.75mm}{

        \begin{tabular}{c|cccc|cccc|cccc|cccc|cccc|cccc}
        \hline
        Motion & \multicolumn{4}{|c|}{Directions} & \multicolumn{4}{|c|}{Greeting} & \multicolumn{4}{|c|}{Phoning} 
               & \multicolumn{4}{|c|}{Posing} & \multicolumn{4}{|c|}{Purchases} & \multicolumn{4}{|c}{Sitting}\\
        \hline
        millisecond & 80&160&320&400 & 80&160&320&400 & 80&160&320&400 & 80&160&320&400 & 80&160&320&400 & 80&160&320&400 \\
        \hline
        ZeroV~\cite{Martinez_2017_CVPR} & 0.39 & 0.59 & 0.79 & 0.89 & 0.54 & 0.89 & 1.30 & 1.49 & 0.64 & 1.21 & 1.65 & 1.83 & 0.28 & 0.57 & 1.13 & 1.37 & 0.62 & 0.88 & 1.19 & 1.27 & 0.40 & 1.63 & 1.02 & 1.18 \\
        Res-sup~\cite{Martinez_2017_CVPR} & 0.41 & 0.64 & 0.80 & 0.92 & 0.57 & 0.83 & 1.45 & 1.60 & 0.59 & 1.06 & 1.45 & 1.60 & 0.45 & 0.85 & 1.34 & 1.56 & 0.58 & 0.79 & 1.08 & 1.15 & 0.41 & 0.68 & 1.12 & 1.33 \\
        CSM~\cite{Li_2018_CVPR} & 0.39 & 0.60 & 0.80 & 0.91 & 0.51 & 0.82 & 1.21 & 1.38 & 0.59 & 1.13 & 1.51 & 1.65 & 0.29 & 0.60 & 1.12 & 1.37 & 0.63 & 0.91 & 1.19 & 1.29 & 0.39 & 0.61 & 1.02 & 1.18 \\
        TP-RNN~\cite{abs-1810-09676} & 0.38 & 0.59 & 0.75 & 0.83 & 0.51 & 0.86 & 1.27 & 1.44 & 0.57 & 1.08 & 1.44 & 1.59 & 0.42 & 0.76 & 1.29 & 1.54 & 0.59 & 0.82 & 1.12 & 1.18 & 0.41 & 0.66 & 1.07 & 1.22 \\
        AGED~\cite{Gui_2018_ECCV} & {\bf0.23} & {\bf0.39} & 0.62 & 0.69 & 0.54 & 0.80 & 1.29 & 1.45 & 0.52 & 0.96 & {\bf1.22} & 1.43 & 0.30 & 0.58 & 1.12 & 1.33 & 0.46 & 0.78 & 1.00 & 1.07 & 0.41 & 0.75 & 1.04 & 1.19 \\
        Skel-TNet~\cite{AAAI_Guo} & {0.36} & {0.58} & 0.77 & 0.86 & 0.50 & 0.84 & 1.28 & 1.45 & 0.58 & 1.12 & {1.52} & 1.64 & 0.29 & 0.62 & 1.19 & 1.44 & 0.58 & 0.84 & 1.17 & 1.24 & 0.40 & 0.61 & 1.01 & 1.15 \\
        \hline
        Sym-GNN (No recg)& {0.24} & 0.45 & {0.61} & {0.67} & {0.36} & {0.61} & {0.98} & {1.17} & {0.50} & {0.86} & {1.29} & {1.43} & {\bf0.18} & {\bf0.44} & {0.99} & {1.22} & {\bf0.40} & {0.62} & {1.00} & {1.08} & {\bf0.23} & {\bf0.41} & {0.80} & {0.97} \\
        Sym-GNN & {\bf0.23} & 0.42 & {\bf0.57} & {\bf0.65} & {\bf0.35} & {\bf0.60} & {\bf0.95} & {\bf1.15} & {\bf0.48} & {\bf0.80} & {1.28} & {\bf1.41} & {\bf0.18} & {0.45} & {\bf0.97} & {\bf1.20} & {\bf0.40} & {\bf0.60} & {\bf0.97} & {\bf1.04} & {0.24} & {\bf0.41} & {\bf0.77} & {\bf0.95} \\
        \hline
        \hline
        Motion & \multicolumn{4}{|c|}{Sitting Down} & \multicolumn{4}{|c|}{Taking Photo} & \multicolumn{4}{|c|}{Waiting} 
               & \multicolumn{4}{|c|}{Walking Dog} & \multicolumn{4}{|c|}{Walking Together} & \multicolumn{4}{|c}{Average}\\
        \hline
        millisecond & 80&160&320&400 & 80&160&320&400 & 80&160&320&400 & 80&160&320&400 & 80&160&320&400 & 80&160&320&400 \\
        \hline
        ZeroV~\cite{Martinez_2017_CVPR} & 0.39 & 0.74 & 1.07 & 1.19 & 0.25 & 0.51 & 0.79 & 0.92 & 0.34 & 0.67 & 1.22 & 1.47 & 0.60 & 0.98 & 1.36 & 1.50 & 0.33 & 0.66 & 0.94 & 0.99 & 0.39 & 0.77 & 1.05 & 1.21\\
        Res-sup.~\cite{Martinez_2017_CVPR} & 0.47 & 0.88 & 1.37 & 1.54 & 0.28 & 0.57 & 0.90 & 1.02 & 0.32 & 0.63 & 1.07 & 1.26 & 0.52 & 0.89 & 1.25 & 1.40 & 0.27 & 0.53 & 0.74 & 0.79 & 0.40 & 0.69 & 1.04 & 1.18\\
        CSM~\cite{Li_2018_CVPR} & 0.41 & 0.78 & 1.16 & 1.31 & 0.23 & 0.49 & 0.88 & 1.06 & 0.30 & 0.62 & 1.09 & 1.30 & 0.59 & 1.00 & 1.32 & 1.44 & 0.27 & 0.52 & 0.71 & 0.74 & 0.38 & 0.68 & 1.01 & 1.13 \\
        TP-RNN~\cite{abs-1810-09676} & 0.41 & 0.79 & 1.13 & 1.27 & 0.26 & 0.51 & 0.80 & 0.95 & 0.30 & 0.60 & 1.09 & 1.28 & 0.53 & 0.93 & 1.24 & 1.38 & 0.23 & 0.47 & 0.67 & 0.71 & 0.37 & 0.66 & 0.99 & 1.11 \\
        AGED~\cite{Gui_2018_ECCV} & 0.33 & {0.61} & 0.97 & 1.08 & 0.23 & 0.48 & 0.81 & 0.95 & 0.25 & 0.50 & 1.02 & 1.12 & 0.50 & 0.82 & 1.15 & 1.27 & 0.23 & 0.42 & 0.56 & 0.63 & 0.33 & 0.58 & 0.94 & 1.01 \\
        Skel-TNet~\cite{AAAI_Guo} & 0.37 & {0.72} & 1.05 & 1.17 & 0.24 & 0.47 & 0.78 & 0.93 & 0.30 & 0.63 & 1.17 & 1.40 & 0.54 & 0.88 & 1.20 & 1.35 & 0.27 & 0.53 & 0.68 & 0.74 & 0.36 & 0.64 & 0.99 & 1.02 \\
        \hline
        Sym-GNN (No recg)& {0.30} & 0.62 & {0.91} & {1.03} & {0.16} & {0.34} & {0.55} & {0.66} & {\bf0.22} & {0.49} & {0.89} & {1.09} & {\bf0.42} & {0.74} & {1.09} & {1.25} & {0.17} & {0.34} & {0.52} & {0.57} & {\bf0.26} & {0.50} & {0.82} & {0.94}\\ 
        Sym-GNN & {\bf0.28} & {\bf0.60} & {\bf0.89} & {\bf0.99} & {\bf0.14} & {\bf0.32} & {\bf0.53} & {\bf0.64} & {\bf0.22} & {\bf0.48} & {\bf0.87} & {\bf1.06} & {\bf0.42} & {\bf0.73} & {\bf1.08} & {\bf1.22} & {\bf0.16} & {\bf0.33} & {\bf0.50} & {\bf0.56} & {\bf0.26} & {\bf0.49} & {\bf0.79} & {\bf0.92}\\ 
        \hline
        \end{tabular}}
        \vspace{-10pt}
    \label{tab:pred_h36m_11}
\end{table*}
\begin{table*}[htbp]
    \centering
    \caption{Comparisons of MAEs between our model and the state-of-the-art methods on the 8 actions of CMU Mocap dataset. We evaluate the model for long-term prediction and present the MAEs at both short and long-term prediction time stamps. }
    \vspace{-10pt}
    \footnotesize
    \setlength{\tabcolsep}{1.30mm}{

        \begin{tabular}{c|ccccc|ccccc|ccccc|ccccc}
        \hline
        Motion & \multicolumn{5}{|c|}{Basketball} & \multicolumn{5}{|c|}{Basketball Signal} & \multicolumn{5}{|c|}{Directing Traffic} &  \multicolumn{5}{|c}{Jumping} \\ \hline
        milliseconds & 80 & 160 & 320 & 400 & 1000 & 80 & 160 & 320 & 400 & 1000 & 80 & 160 & 320 & 400 & 1000 & 80 & 160 & 320 & 400 & 1000\\ \hline
        Res-sup~\cite{Martinez_2017_CVPR} & 0.49 & 0.77 & 1.26 & 1.45 & 1.77 & 0.42 & 0.76 & 1.33 & 1.54 & 2.17 & 0.31 & 0.58 & 0.94 & 1.10 & 2.06 & 0.57 & 0.86 & 1.76 & 2.03 & 2.42\\
        Res-uns~\cite{Martinez_2017_CVPR} & 0.53 & 0.82 & 1.30 & 1.47 & 1.81 & 0.44 & 0.80 & 1.35 & 1.55 & 2.17 & 0.35 & 0.62 & 0.95 & 1.14 & 2.08 & 0.59 & 0.90 & 1.82 & 2.05 & 2.46\\
        CSM~\cite{Li_2018_CVPR} & 0.37 & 0.62 & 1.07 & 1.18 & 1.95 & 0.32 & 0.59 & 1.04 & 1.24 & 1.96 & 0.25 & 0.56 & 0.89 & 1.00 & 2.04 & 0.39 & 0.60 & {1.36} & {1.56} & 2.01 \\
        BiHMP-GAN~\cite{AAAI_Kundu} & 0.37 & 0.62 & 1.02 & 1.11 & 1.83 & 0.32 & 0.56 & 1.01 & 1.18 & 1.89 & 0.25 & 0.51 & 0.85 & 0.96 & 1.95 & 0.39 & 0.57 & 1.32 & 1.51 & 1.94 \\
        Skel-TNet~\cite{AAAI_Guo} & 0.35 & 0.63 & 1.04 & 1.14 & 1.78 & 0.24 & 0.40 & 0.69  & 0.80 & 1.07 & 0.22 & 0.44 & 0.78 & 0.90 & 1.88 & 0.35 & {\bf0.53} & {\bf1.28} & {\bf1.49} & 1.85 \\
        \hline
        Sym-GNN (No recg)& {0.33} & {\bf0.48} & {0.95} & {1.09} & {\bf1.47} & {0.15} & {0.26} & {0.47} & {0.56} & {1.04} & {\bf0.20} & {\bf0.41} & {0.77} & {0.89} & {1.95} & {\bf0.32} & {0.55} & 1.40 & 1.60 & {1.87} \\ 
        Sym-GNN & {\bf0.32} & {\bf0.48} & {\bf0.91} & {\bf1.06} & {\bf1.47} & {\bf0.12} & {\bf0.21} & {\bf0.38} & {\bf0.49} & {\bf0.94} & {\bf0.20} & {\bf0.41} & {\bf0.75} & {\bf0.87} & {\bf1.84} & {\bf0.32} & {0.55} & 1.40 & 1.60 & {\bf1.82} \\ 
        \hline
        \hline
        Motion & \multicolumn{5}{|c|}{Running} & \multicolumn{5}{|c|}{Soccer} & \multicolumn{5}{|c|}{Walking} & \multicolumn{5}{|c}{Washing Window} \\ \hline
        milliseconds & 80 & 160 & 320 & 400 & 1000 & 80 & 160 & 320 & 400 & 1000 & 80 & 160 & 320 & 400 & 1000 & 80 & 160 & 320 & 400 & 1000\\ \hline
        Res-sup~\cite{Martinez_2017_CVPR} & 0.32 & 0.48 & 0.65 & 0.74 & 1.00 & 0.29 & 0.50 & 0.87 & 0.98 & 1.73 & 0.35 & 0.45 & 0.59 & 0.64 & 0.88 & 0.32 & 0.47 & 0.74 & 0.93 & 1.37 \\
        Res-uns~\cite{Martinez_2017_CVPR} & 0.35 & 0.50 & 0.69 & 0.76 & 1.04 & 0.31 & 0.51 & 0.90 & 1.00 & 1.77 & 0.36 & 0.47 & 0.62 & 0.65 & 0.93 & 0.33 & 0.47 & 0.75 & 0.95 & 1.40\\
        CSM~\cite{Li_2018_CVPR} & 0.28 & 0.41 & {\bf0.52} & 0.57 & {0.67} & 0.26 & 0.44 & 0.75 & 0.87 & 1.56 & 0.35 & 0.44 & 0.45 & 0.50 & 0.78 & 0.30 & 0.47 & 0.80 & 1.01 & 1.39 \\
        BiHMP-GAN~\cite{AAAI_Kundu} & 0.28 & 0.40 & 0.50 & 0.53 & 0.62 & 0.26 & 0.44 & 0.72 & 0.82 & 1.51 & 0.35 & 0.45 & 0.44 & 0.46 & 0.72 & 0.31 & 0.46 & 0.77 & 0.92 & 1.31 \\
        Skel-TNet~\cite{AAAI_Guo} & 0.38 & 0.48 & 0.57 & 0.62 & 0.71 & 0.24 & 0.41 & 0.69 & 0.79 & 1.44 & 0.33 & 0.41 & 0.45 & 0.48 & 0.73 & 0.31 & 0.46 & 0.79 & 0.96 & 1.37 \\
        \hline
        Sym-GNN (No recg)& {\bf0.21} & {\bf0.33} & 0.53 & {\bf0.56} & {0.66} & {0.22} & {0.38} & {0.72} & {0.83} & {1.38} & {\bf0.26} & {\bf0.32} & {0.38} & {0.41} & {0.54} & {\bf0.22} & {\bf0.33} & {0.62} & {0.83} & {1.07}\\ 
        Sym-GNN & {\bf0.21} & {\bf0.33} & 0.53 & {\bf0.56} & {\bf0.65} & {\bf0.19} & {\bf0.32} & {\bf0.66} & {\bf0.78} & {\bf1.32} & {\bf0.26} & {\bf0.32} & {\bf0.35} & {\bf0.39} & {\bf0.52} & {\bf0.22} & {\bf0.33} & {\bf0.55} & {\bf0.73} & {\bf1.05}\\ 
        \hline
    \end{tabular}}
    \label{tab:pred_cmu}
\end{table*}
Notably, for Human 3.6M, there is a relatively large gap between the top-1 and top-5 accuracies, because the input motions are some fragmentary clips of long sequences with incomplete semantics and activities have subtle differences (e.g. `Eating' and `Smoking' are similar). In other words, Sym-GNN learns the common features and provides reasonable discrimination, resulting in high top-5 accuracy; but it confuses in non-semantic variances, causing not high top-1 accuracy. However, CMU Mocap has more distinctive actions, where we obtain high classification accuracies.

\subsubsection{3D Skeleton-based Motion Prediction}
To validate the model for predicting future motions, we train the Sym-GNN on NTU-RGB+D, Human 3.6M, and CMU Mocap. There are two specific tasks: short-term and long-term motion prediction. Concretely, the target of short-term prediction is commonly to predict poses within 400 milliseconds, while the long-term prediction aims to predict poses in 1000 ms or longer. To reveal the effectiveness of Sym-GNN, we introduce many state-of-the-art methods, which learned dynamics from pose vectors~\cite{Fragkiadaki_2015_ICCV,Martinez_2017_CVPR,abs-1810-09676,Gui_2018_ECCV,AAAI_Kundu} or separate body-parts~\cite{Jain_2016_CVPR,Li_2018_CVPR,AAAI_Guo}. We also introduce a naive baseline, named ZeroV~\cite{Martinez_2017_CVPR}, which sets all predictions to be the last observed frame.

{\bf Short-term motion prediction:} We validate Sym-GNN on two datasets: NTU-RGB+D and Human 3.6M. For NTU-RGB+D, we train Sym-GNN to generate future 10 frames for each input sequence. We compare our Sym-GNN with several previous methods and the Sym-GNN variant which abandons auxiliary action-recognition head (No recg). As the metric, We use the percentage of correct points within a normalized region 0.05 (PCK@0.05), that is, a joint is counted as correctly predicted if the normalized distance between the predicted location and ground-truth is less than 0.05. The PCK@0.05 of different models are presented in Table~\ref{tab:pred_ntu}.
We see: 1) our model extremely outperforms the baselines with a large margin especially for the longer term; 2) Using action recognition and motion prediction together obtains the highest PCK@0.05 along time, demonstrating the enhancements from recognition task for dynamics learning.

Then, we compare Sym-GNN to baselines for short-term prediction on Human 3.6M, where the models generate poses up to the future 400 ms. We analyze several variants of Sym-GNN with different components, including using only joint-scale actional graphs (Only J-A) or joint-scale structural graphs (Only J-S), as well as no recognition task (No recg). As another metric, the mean angle errors (MAE) between the predictions and the ground truths are computed, representing the errors from predicted poses to targets in angle space. We first test $4$ representative actions: `Walking', `Eating', `Smoking' and `Discussion'. Table~\ref{tab:pred_h36m_4} shows MAEs of different methods that predict motions up to 400 ms.
As we see, when Sym-GNN simultaneously employs multiple graphs and multitasking, our method outperforms all the baselines and its own ablations.

We also test Sym-GNN on the remaining $11$ actions in Human 3.6M, where the MAEs of some recent methods are shown in Table~\ref{tab:pred_h36m_11}. 
Sym-GNN also achieves the best performance on most actions and the lowest average MAE on $15$ motions. Although the mentioned top-1 classification accuracy on this dataset is not very high (see Table~\ref{tab:recg_h36m_cmu}), we note that the estimated soft labels cover the common motion factors, resulting in high top-5 recognition accuracy. For example, people walk in `Walking', `Walking Dog' and `Walking Together', and we need the walking factors instead of the specific labels for motion generation. Given the soft labels, the model tends to obtain precise predictions.

{\bf Long-term motion prediction:} For long-term prediction, the Sym-GNN is tested on Human 3.6M and CMU Mocap. We predict the future poses up to 1000 millisecond. It is challenging due to action variation and non-linearity~\cite{Martinez_2017_CVPR}. Table~\ref{tab:pred_h36m_longterm} presents the MAEs of various models for predicting the 4 motions in Human 3.6M at the future 560 ms and 1000 ms.
\begin{table}[tbp]
    \centering
    \caption{Comparisons of MAEs between our model and other methods for long-term motion prediction on $4$ actions of H3.6M.}
    \vspace{-10pt}
    \footnotesize
    \setlength{\tabcolsep}{1.24mm}{
        \begin{tabular}{c|cc|cc|cc|cc}
            \hline
            Motion & \multicolumn{2}{|c|}{Walking}& \multicolumn{2}{|c}{Eating}& \multicolumn{2}{|c|}{Smoking}& \multicolumn{2}{|c}{Discussion} \\ 
            \hline
            milliseconds & 560 & 1k & 560 & 1k & 560 & 1k & 560 & 1k\\ \hline
            ZeroV~\cite{Martinez_2017_CVPR} & 1.35 & 1.32 & 1.04 & 1.38 & 1.02 & 1.69 & 1.41 & 1.96 \\
            ERD~\cite{Fragkiadaki_2015_ICCV} & 2.00 & 2.38 & 2.36 & 2.41 & 3.68 & 3.82 & 3.47 & 2.92 \\
            Lstm3LR~\cite{Fragkiadaki_2015_ICCV} & 1.81 & 2.20 & 2.49 & 2.82 & 3.24 & 3.42 & 2.48 & 2.93 \\
            SRNN~\cite{Jain_2016_CVPR} & 1.90 & 2.13 & 2.28 & 2.58 & 3.21 & 3.23 & 2.39 & 2.43 \\
            DropAE~\cite{GhoshSAH17} & 1.55 & 1.39 & 1.76 & 2.01 & 1.38 & 1.77 & 1.53 & 1.73 \\
            Res-sup.~\cite{Martinez_2017_CVPR} & 0.93 & 1.03 & 0.95 & 1.08 & 1.25 & 1.50 & 1.43 & 1.69\\
            CSM~\cite{Li_2018_CVPR} & 0.86 & 0.92 & 0.89 & 1.24 & 0.97 & 1.62 & 1.44 & 1.86\\ 
            TP-RNN~\cite{abs-1810-09676} & {\bf0.74} & {\bf0.77} & 0.84 & 1.14 & 0.98 & 1.66 & 1.39 & 1.74\\ 
            AGED~\cite{Gui_2018_ECCV} & 0.78 & 0.91 & 0.86 & 0.93 & 1.06 & 1.21 & 1.25 & 1.30\\ 
            BiHMP-GAN~\cite{AAAI_Kundu} & / & 0.85 & / & 1.20 & / & {\bf 1.11} & / & 1.77 \\ 
            Skel-TNet~\cite{AAAI_Guo} & 0.79 & 0.83 & 0.84 & 1.06 & 0.98 & 1.21 & 1.19 & 1.75 \\ 
            \hline
            Sym-GNN & 0.75 & 0.78 & {\bf0.77} & {\bf0.88} & {\bf0.92} & {1.18} & {\bf1.17} & {\bf1.28}\\ \hline
        \end{tabular}}
    \label{tab:pred_h36m_longterm}
    % \vspace{-10pt}
\end{table}
We see that Sym-GNN outperforms the competitors on `Eating', `Smoking' and `Discussion', and obtain competitive results on `Walking'.

To further evaluate Sym-GNN, we conduct long-term prediction on eight classes of actions in CMU Mocap. We present the MAEs of Sym-GNN with or without using the action-recognition head. Table~\ref{tab:pred_cmu} shows the predicting MAEs ranging from future 80 ms to 1000 ms. We note that we train the model for long-term prediction, where the `short-term' MAEs are the intermediate results during predicting up to 1000 ms.
We see that Sym-GNN significantly outperforms the state-of-the-art methods on actions `Basketball', `Basketball Signal' and `Washing Window', and obtains competitive performance on `Jumping' and `Running'.

{\bf Effectiveness-efficiency tradeoff:} We also compare the prediction errors and efficiency of various models, because the high response speed and precise generation are both essential for real-time motion prediction. Notably, the AGIM propagates the features between joints and edges iteratively. The iteration times $K$ trades off between effectiveness and speed; i.e. larger $K$ leads to a lower MAE but slower speed. To represent the running speed, we use the generated frame numbers in each $20$ ms (frame period) when we predict up to $400$ ms. We tune $K$ and compare Sym-GNN to other methods on Human 3.6M and show the running speeds and MAEs for prediction in 400 ms. 
\begin{figure}[tb]
    \centering
    \includegraphics[width=7.4cm]{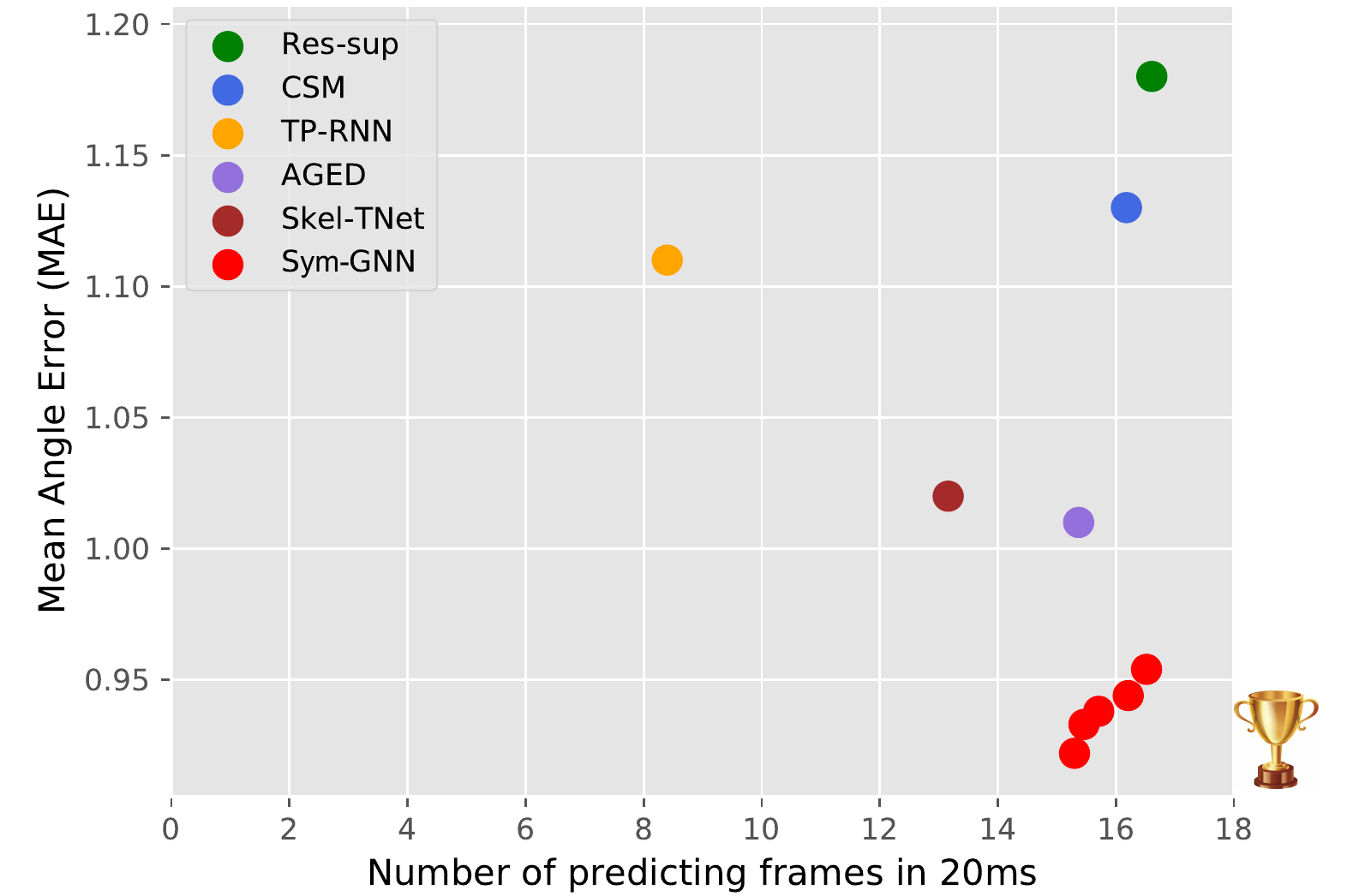}
    \vspace{-10pt}
    \caption{Sym-GNN is both faster and more precise compared to others. Various red circles denote different iteration numbers $K$ in AGIM, where $K = 0, 1, 2, 3, 4$. The bottom right corner (highlighted by a trophy cup) indicates higher speed and lower error, showing an ideal target.}
    \label{fig:time-mae}
    \vspace{-10pt}
\end{figure}
Fig.~\ref{fig:time-mae} shows the effectiveness-efficiency tradeoff, where the $x$-axis is the generated frame numbers in $20$ ms (reflecting prediction speed) and the $y$ axis is the MAE. 
% Thus the bottom right corner of the figure denotes faster and more precise performance, where we highlight by a trophy cup. 
Different red circles denote different numbers of iterations $K$ in AGIM, i.e. from the rightmost circle to the leftmost one, $K = 0, 1, 2, 3, 4$. We see that the proposed Sym-GNN is both faster and more precise compared to its competitors.

\subsection{Ablation Studies}

\subsubsection{Symbiosis of Recognition and Prediction}
To analyze the mutual effects of action recognition and motion prediction, we conduct several experiments. 
\begin{table}[]
    \centering
    \caption{Action recognition accuracies with noisy motion prediction targets in varying degrees on NTU-RGB+D dataset. `No pred' denotes model without motion prediction task.}
    \vspace{-10pt}
    \setlength{\tabcolsep}{9mm}{
    \begin{tabular}{c|c|c}
    \hline
       Noise ratio  & CS & CV \\
    \hline     
    $0\%$     & ${\bf 90.1\%}$ & ${\bf 96.4\%}$ \\
    $10\%$     & $89.8\%$ & $96.1\%$ \\
    $20\%$     & $89.5\%$ & $96.1\%$ \\
    $50\%$     & $89.1\%$ & $95.5\%$ \\
    $70\%$     & $88.5\%$ & $94.9\%$ \\
    $100\%$     & $87.7\%$ & $93.9\%$ \\
    No pred    & $89.0\%$ & $95.7\%$ \\
    \hline
    \end{tabular}}
    \label{tab:wrong_prediction}
\end{table}

\begin{table}[]
    \centering
    \caption{Motion prediction MAEs with noisy action labels in varying degrees on Human 3.6M dataset. `No recg' denotes model without recognition task.}
    \vspace{-10pt}
    \setlength{\tabcolsep}{4.0mm}{
    \begin{tabular}{c|c|c|c|c}
    \hline
    Motion  & \multicolumn{4}{|c}{Average} \\
    \hline
       Noise ratio  & 80 ms & 160 ms & 320 ms & 400 ms \\
    \hline     
    $0\%$  & ${\bf 0.26}$ & $0.49$ & ${\bf 0.79}$ & ${\bf 0.92}$ \\
    $10\%$ & ${\bf 0.26}$ & ${\bf 0.48}$ & ${\bf 0.79}$ & $0.93$ \\
    $20\%$ & ${\bf 0.26}$ & $0.49$ & $0.82$ & $0.94$ \\
    $30\%$ & ${\bf 0.26}$ & $0.50$ & $0.82$ & $0.93$ \\
    $50\%$ & ${\bf 0.26}$ & $0.51$ & $0.83$ & $0.97$ \\
    $100\%$& $0.27$ & $0.53$ & $0.85$ & $1.03$ \\
    No recg& ${\bf 0.26}$ & $0.50$ & $0.82$ & $0.94$ \\
    \hline
    \end{tabular}}
    \label{tab:wrong_recognition}
\end{table}
\begin{figure}[t]
\centering
    \includegraphics[width=7cm]{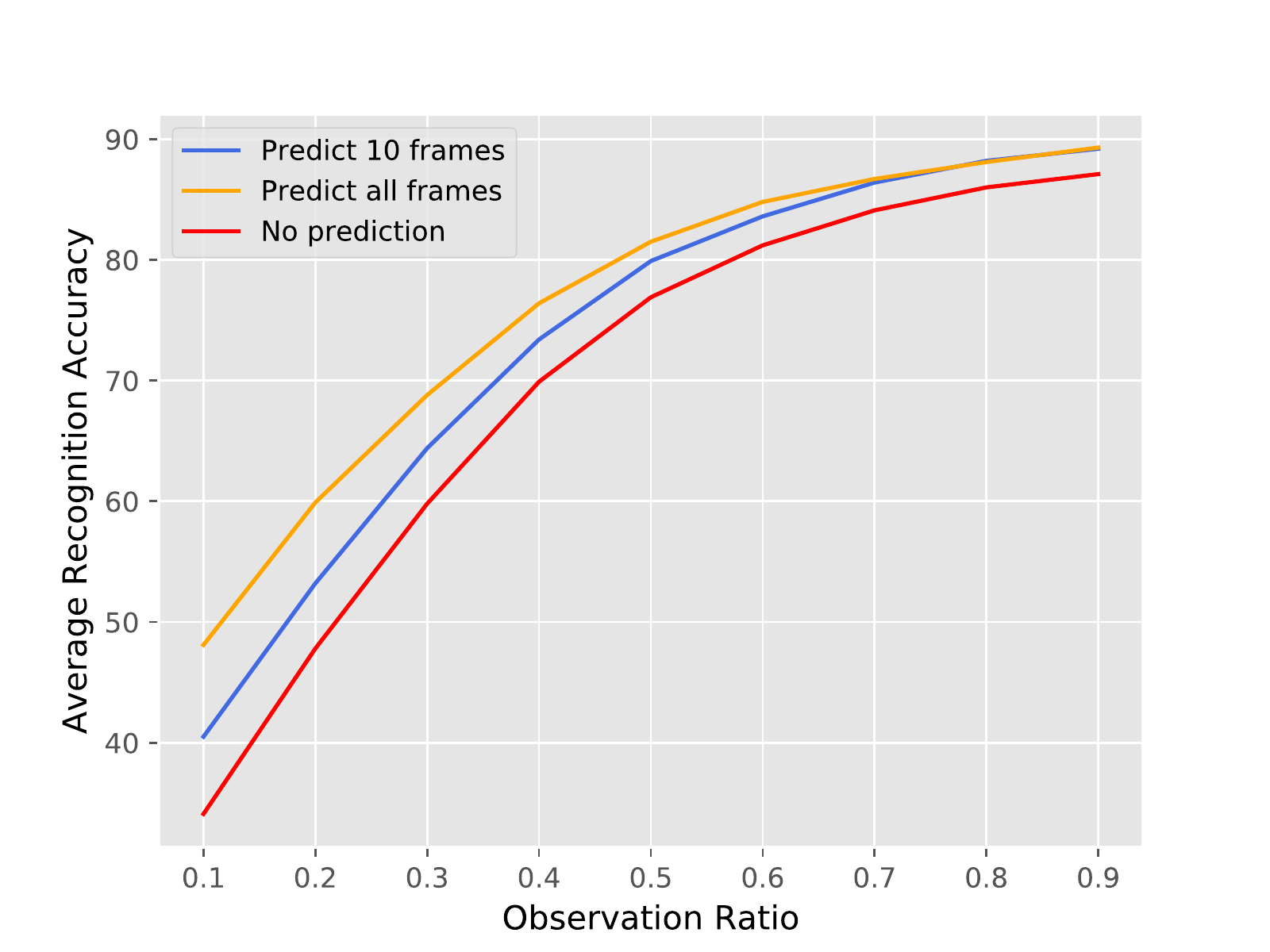}
    \vspace{-10pt}
    \caption{Given the same input, predicting more future poses leads to a better performance of action recognition. We see that across all the observation ratios, predicting all future poses is better than predicting $10$ future poses; and both are better than no prediction.
    }
    \label{fig:anticipation}
    \vspace{-10pt}
\end{figure}

We first study the effects on action recognition from motion prediction. We use accurate class labels but noisy future poses to train the multitasking Sym-GNN for action recognition. To represent noisy supervisions, we randomly shuffle a percentage of targets motions among training data. Table~\ref{tab:wrong_prediction} presents the recognition accuracies with various ratios of noisy prediction targets on two benchmarks of NTU-RGB+D. We also show the recognition results of the model without motion-prediction head. We see that 1) the predicted head benefits the action-recognition head. Introducing a motion-prediction head is beneficial even when the noise ratio is around $50\%$; 2) when the noise ratio exceeds $50\%$, the recognition performance tends to be slightly worse than that of the model without the motion-prediction head, reflecting that the action recognition is robust against the deflected motion prediction. Consequently, we show that motion prediction strengthens action recognition.

On the other hand, we test how confused recognition results affect motion prediction by using noisy action categories. Following Table~\ref{tab:wrong_prediction}, we shuffle training labels to represent categorical noise. Table~\ref{tab:wrong_recognition} presents the average MAEs for short-term prediction with noisy action labels on Human 3.6M. We demonstrate that an accurate action-recognition head helps effective motion prediction.

We finally test the promotion on recognition when the observed data is limited, where we intercept the early motions by a ratio (e.g. $10\%$) for action recognition. There are three models with various prediction strategies: 1) predicting the future 10 frames (`Pred 10 frames'); 2) predicting all future frames (`Pred all frames'); 3) no prediction (`No pred'). Fig.~\ref{fig:anticipation} illustrates the recognition accuracies of three models on different observation ratios. As we see, when the observation ratio is low, `Pred all frames' can be aware of the entire action sequences and capture richer dynamics, showing the best performance; when the observation ratio is high, predicting 10 or all frames are similar because the inputs carry sufficient patterns, but they outperform `No pred' as they preserve information, showing enhancement. By introducing the motion-prediction head, our Sym-GNN has the potential for action classification in the early period.

\subsubsection{Effects of Graphs}
In this section, we study the abilities of various graphs, namely, only joint-scale structural graphs (Only J-S), only joint-scale actional graph (Only J-A), only part-scale graph (Only P), and combining them (full).

For action recognition, we train Sym-GNN on NTU-RGB+D, Cross-Subject and investigate different graph configurations. While involving joint-scale structural graph, we respectively set the number of hop in the joint-scale structural graphs (JS-Hop) to be $\Gamma = 1, 2, 3, 4$. Note that when we use only joint-scale structural graph with $\Gamma = 1$, the corresponding graph is exactly the skeleton itself. Table~\ref{tab:graph_recg} presents the results of Sym-GNN with different graph components for action recognition.
\begin{table}[tp]
    \centering
    \caption{Recognition accuracies on NTU-RGB+D, CS with various graphs: only joint-scale structural graphs (Only J-S), only joint-scale actional graph (Only J-A), only part-scale graph (Only P) and all graphs (full).}
    \vspace{-10pt}
    \setlength{\tabcolsep}{3.4mm}{
        \begin{tabular}{c|c|c|c|c}
            \hline
            JS-Hop ($\Gamma$) & Only J-S & Only J-A & Only P & full \\
            \hline
            % \multirow{4}{*}{\tabincell{c}{Only\\ Recg}} & 1 & $84.5\%$ & \multirow{4}{*}{$84.2\%$} & \multirow{4}{*}{$86.5\%$} & $85.2\%$ \\
            % ~ & 2 & $85.7\%$ & ~ & ~ & $86.1\%$ \\
            % ~ & 3 & $86.5\%$ & ~ & ~ & $87.7\%$ \\
            % ~ & 4 & $88.0\%$ & ~ & ~ & $89.0\%$ \\
            % \hline
            % \multirow{4}{*}{\tabincell{c}{Recg\\ \& Pred}} 
            1 & $85.9\%$ & \multirow{4}{*}{$85.7\%$} & \multirow{4}{*}{$87.3\%$} & $86.1\%$ \\
            2 & $86.2\%$ & ~ & ~ & $86.9\%$ \\
            3 & $87.5\%$ & ~ & ~ & $88.3\%$ \\
            4 & $88.3\%$ & ~ & ~ & ${\bf90.1\%}$ \\
            \hline
    \end{tabular}}
    \label{tab:graph_recg}
    \vspace{-5pt}
\end{table}
We see that 1) representing long-range structural relations, higher $\Gamma$ leads to more effective action recognition; 2) combining the multiple graphs introduced from different perspectives improves the action recognition performance significantly.

For motion prediction, we study graphs components using similar setting of Table~\ref{tab:graph_recg}. We validate Sym-GNN on Human 3.6M, the average short-term prediction MAEs are presented in Table~\ref{tab:graph_pred}.
\begin{table}[tp]
    \centering
    \caption{Average MAEs for short-term prediction on H3.6M with various graphs: only joint-scale structural graphs (Only J-S), only joint-scale actional graph (Only J-A), only part-scale graph (Only P) and all graphs (full).}
    \vspace{-10pt}
    \setlength{\tabcolsep}{3.4mm}{
        \begin{tabular}{c|c|c|c|c}
            \hline
            JS-Hop ($\Gamma$) & Only J-S & Only J-A & Only P & full \\
            \hline
            % \multirow{4}{*}{\tabincell{c}{Only\\ Pred}} & 1 & 0.626 & \multirow{4}{*}{0.626} & \multirow{4}{*}{0.619} & 0.624 \\
            % ~ & 2 & 0.624 & ~ & ~ & 0.620 \\
            % ~ & 3 & 0.618 & ~ & ~ & 0.614 \\
            % ~ & 4 & 0.623 & ~ & ~ & 0.616 \\
            % \hline
            1 & 0.622 & \multirow{4}{*}{0.618} & \multirow{4}{*}{0.615} & 0.616 \\
            2 & 0.619 & ~ & ~ & 0.615 \\
            3 & 0.613 & ~ & ~ & {\bf0.611} \\
            4 & 0.618 & ~ & ~ & 0.614 \\
            \hline
    \end{tabular}}
    \label{tab:graph_pred}
    \vspace{-10pt}
\end{table}
We see that the effects multiple relations for promoting motion prediction are demonstrated. We note that too large $\Gamma$ introduces redundancy and confusing relations, enlarging the prediction error.

% We finally study the correlation between the effectiveness and efficiency of our model. We evaluate Sym-GNN on Human 3.6M with various iteration numbers $K$. The average MAEs of short-term and long-term prediction up to the future 400 ms and 1000 ms, as well as the running time, are presented in Table~\ref{tab:agim_iter}.
% \begin{table}[tp]
%     \centering
%     \caption{MAEs and running time (milliseconds, shown in brackets) with various iteration times in AGIM	tested on H3.6M dataset.}
%     \vspace{-10pt}
%     \footnotesize
%     \setlength{\tabcolsep}{1.56mm}{

%         \begin{tabular}{c|cccc|cc}
%             \hline
%             ~ & \multicolumn{4}{|c|}{Short-term} & \multicolumn{2}{|c}{Long-term} \\
%             \hline
%             Time (ms) & 80 & 160 & 320 & 400 & 560 & 1000 \\
%             \hline
%             $K=0$ & 0.28 & 0.52 & 0.83 & 0.95 (24.21) & 1.15 & 1.53 (80.97) \\
%             $K=1$ & 0.26 & 0.51 & 0.81 & 0.94 (24.68) & 1.13 & 1.50 (81.70) \\
%             $K=2$ & 0.27 & 0.51 & 0.81 & 0.94 (25.46) & 1.12 & 1.49 (82.51) \\
%             $K=3$ & 0.26 & 0.50 & 0.81 & 0.93 (25.88) & 1.12 & 1.47 (83.03) \\
%             $K=4$ & {\bf0.26} & {\bf0.49} & {\bf0.79} & {\bf0.92} (26.14) & {\bf1.11} & {\bf1.46} (83.82) \\
%             $K=5$ & 0.27 & 0.50 & 0.81 & 0.94 (26.38) & 1.12 & 1.47 (84.23) \\
%             \hline
%     \end{tabular}}
%     \label{tab:agim_iter}
%     \vspace{-10pt}
% \end{table}
\begin{table}[tp]
    \centering
    \caption{The recognition accuracies of the model with various input difference orders on NTU-RGB+D.}
    \vspace{-10pt}
    \setlength{\tabcolsep}{7.5mm}{
        \begin{tabular}{c|c|c}
            \hline
            Difference Order & CS & CV \\
            \hline
            $\beta=0$ & $88.2\%$ & $95.0\%$ \\
            $\beta=0,1$ & ${\bf90.1\%}$ & ${\bf96.4\%}$ \\
            $\beta=0,1,2$ & $89.8\%$ & ${96.2\%}$ \\
            \hline
    \end{tabular}}
    \label{tab:diff_recg}
    \vspace{-10pt}
\end{table}
\begin{table}[tb]
    \centering
    \caption{The average MAEs of short-term motion prediction with various input difference orders on Human 3.6M.}
    \vspace{-10pt}
        \setlength{\tabcolsep}{4.8mm}{
        \begin{tabular}{c|cccc}
            \hline
            Motion & \multicolumn{4}{|c}{Average}\\
            \hline
            Milliseconds & 80 & 160 & 320 & 400\\
            \hline
            %$\beta=0$ & 0.23 & 0.38 & 0.56 & 0.66 & 0.19 & 0.33 & 0.51 & 0.65 & 0.25 & 0.47 & 0.82 & 0.86 & 0.28 & 0.62 & 0.85 & 0.93 \\
            %$\beta=0,1$ & 0.18 & 0.32 & 0.52 & 0.62 & 0.17 & 0.31 & 0.51 & 0.61 & 0.22 & 0.42 & 0.80 & 0.81 & 0.27 & 0.65 & 0.92 & 0.98 \\
            %$\beta=0,1,2$ & {\bf0.17} & {\bf0.31} & {\bf0.50} & {\bf0.59} & {\bf0.16} & {\bf0.29} & {\bf0.48} & {\bf0.60} & {\bf0.21} & {\bf0.41} & {\bf0.76} & {\bf0.80} & {\bf0.22} & {\bf0.57} & {\bf0.77} & {\bf0.85} \\
            $\beta=0$ & 0.33 & 0.59 & 0.85 & 0.91\\
            $\beta=0,1$ & 0.28 & 0.52 & 0.81 & 0.85\\
            $\beta=0,1,2$ & {\bf0.26} & {\bf0.49} & {\bf0.79} & {\bf0.82}\\
            \hline
    \end{tabular}}
    \label{tab:diff_pred}
    \vspace{-10pt}
\end{table}

\subsubsection{Balance Joint-Scale Actional and Structural Graphs}
In our model, we present that the power of joint-scale actional and structural graphs in JGC operator are traded off by a hyper-parameter $\lambda_{\rm act}$ (see~\eqref{eq:ASGC}). Here we analyze how $\lambda_{\rm act}$ affects the model performances. 

\begin{figure}[t]
\centering
    \includegraphics[width=7cm]{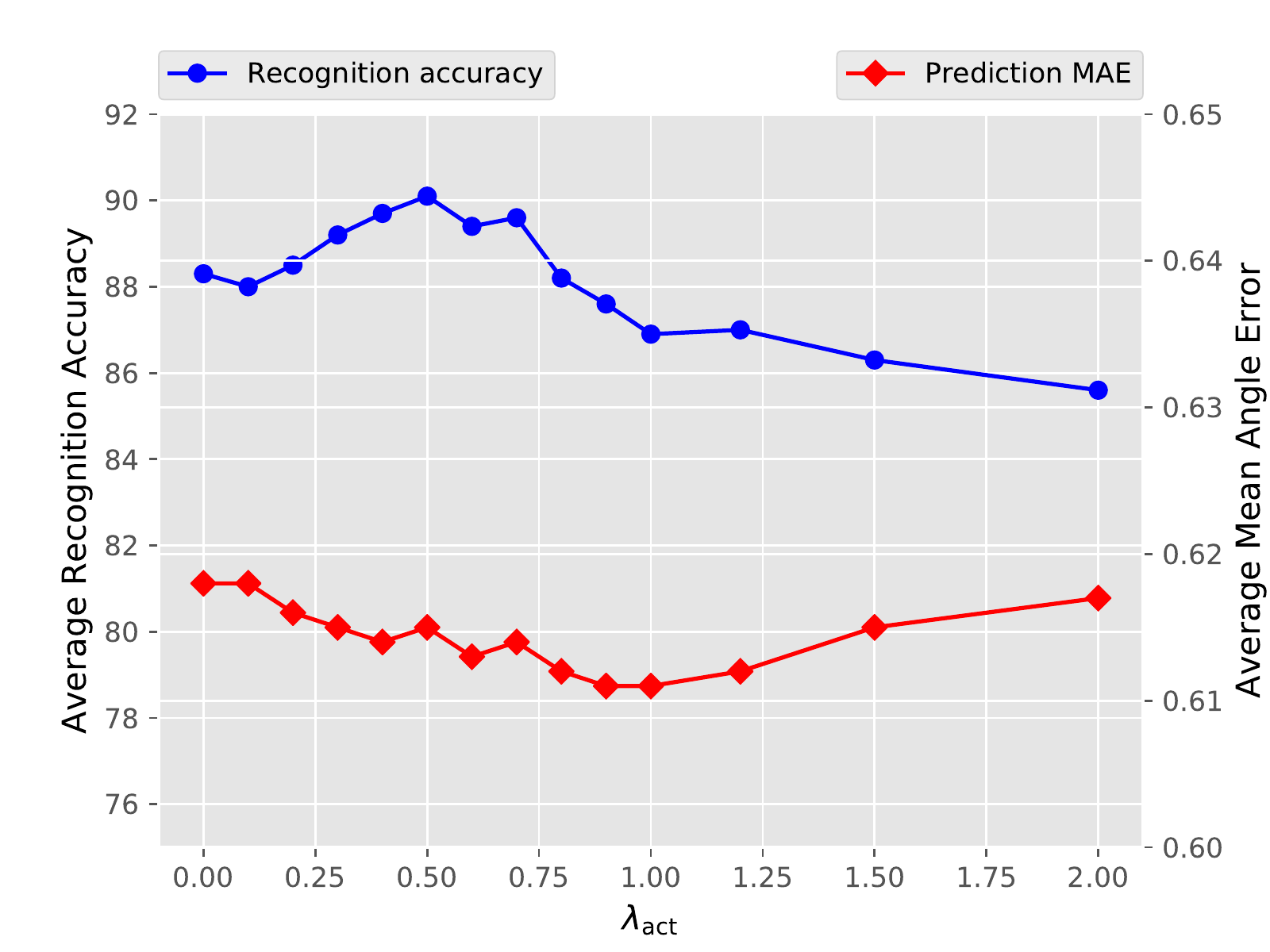}
    \vspace{-10pt}
    \caption{Average action recognition accuracies and motion prediction MAEs of models with different $\lambda_{\rm act}$.}
    \label{fig:lambda}
    \vspace{-10pt}
\end{figure}

For action recognition, we test our model on NTU-RGB+D, Cross-Subject, and present the classification accuracies with different $\lambda_{\rm act}$; for motion prediction, we show the average MAEs for short-term prediction. Fig.~\ref{fig:lambda} illustrates the model performances for both tasks.
We see: 1) when $\lambda_{act}=0.5$, we obtain the highest recognition accuracies, showing large improvements than cases with other $\lambda_{act}$; 2) for motion prediction, the performance is robust against different $\lambda_{\rm act}$, where the MAEs fluctuate around $0.615$, but $\lambda_{\rm act}=0.9$ and $1.0$ lead to the lowest errors.

\subsubsection{High-order Difference}

Here we study the high-order differences of the input actions for action recognition and motion prediction, which help to capture richer motion dynamics.  In our model, the difference orders are considered to be $\beta=0,1,2$, reflecting positions, velocities, and accelerations. We present the results of action recognition on NTU-RGB+D and the results of motion prediction on Human 3.6M.

For action recognition, we test Sym-GNN with $\beta=0$, $\beta=0,1$ and $\beta=0,1,2$ on the two benchmarks (CS \& CV) of NTU-RGB+D dataset. Table~\ref{tab:diff_recg} presents the average recognition accuracies of Sym-GNN with three difference configurations.
We see that $\beta=0,1$ leads to the highest classification accuracies on both benchmarks, indicating that positions and velocities capture comprehensive movement patterns to provide rich semantics, meanwhile the model complexity is not very high.

For motion prediction, we feed the model with various action differences of Human 3.6M to generate future poses in within 400 ms. We obtain the predicting average MAEs on the timestamps of 80, 160, 320 and 400 ms. The results are presented in Table~\ref{tab:diff_pred}.
We see that Sym-GNN forecasts the future poses with the lowest prediction errors when we feed the action differences with $\beta=0,1,2$, showing the effectiveness of combining high-order  motion states.

\begin{table}[tp]
    \centering
    \caption{The recognition accuracies of model with different parallel networks on NTU-RGB+D.}
    \vspace{-10pt}
    \setlength{\tabcolsep}{7.5mm}{
        \begin{tabular}{c|c|c}
            \hline
            Parallel Network & CS & CV \\
            \hline
            Only Joint & $87.1\%$ & $93.8\%$ \\
            Only Bone & $87.4\%$ & $93.5\%$ \\
            Joint \& Bone & ${\bf 90.1\%}$ & ${\bf 96.4\%}$ \\
            \hline
    \end{tabular}}
    \label{tab:parallel}
    \vspace{-10pt}
\end{table}

\subsubsection{Bone-based Dual Graph Neural Networks}
We validate the effectiveness of using dual networks which take joint and bone features as inputs for action recognition, respectively. Table~\ref{tab:parallel} presents the recognition accuracies for different combinations of joint-based and bone-based dual networks on two benchmarks of NTU-RGB+D dataset. We see that only using joint features or bone features for action recognition cannot obtain the most accurate recognition, but combining joint and bone features could improve the classification performances with a large margin, indicating the complementary information carried by the two networks.

\subsection{Visualization}
In this section, we visualize some representations of Sym-GNN, including the learned joint-scale actional graphs and their low dimensional manifolds. Moreover, we show some predicted motions to evaluate model qualitatively. 

\subsubsection{Joint-Scale Actional Graphs}
\begin{figure}[t]
\centering
    \includegraphics[width=8cm]{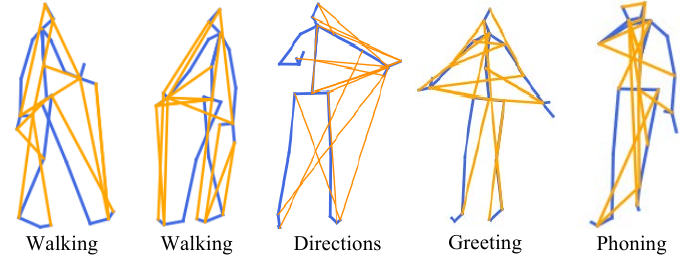}
    \vspace{-10pt}
    \caption{Joint-scale actional graphs on motions in H3.6M. Yellow lines indicate the connections in actional graphs
    and blue lines indicate the connections in skeleton graphs.
    The two plots on the left show the graphs of `Walking', the three plots on the right show the graphs of `Directions', `Greeting' and `Phoning'.}
    \label{fig:action_graph}
    \vspace{-10pt}
\end{figure}

We first show the learned joint-scale actional graphs on four motions in Human 3.6M. Fig.~\ref{fig:action_graph} illustrates the edges which have the top-15 largest weights in each graph, indicating the 15 strongest action-based relations associated with different motions.
\begin{figure}[t]
\centering
    \includegraphics[width=7cm]{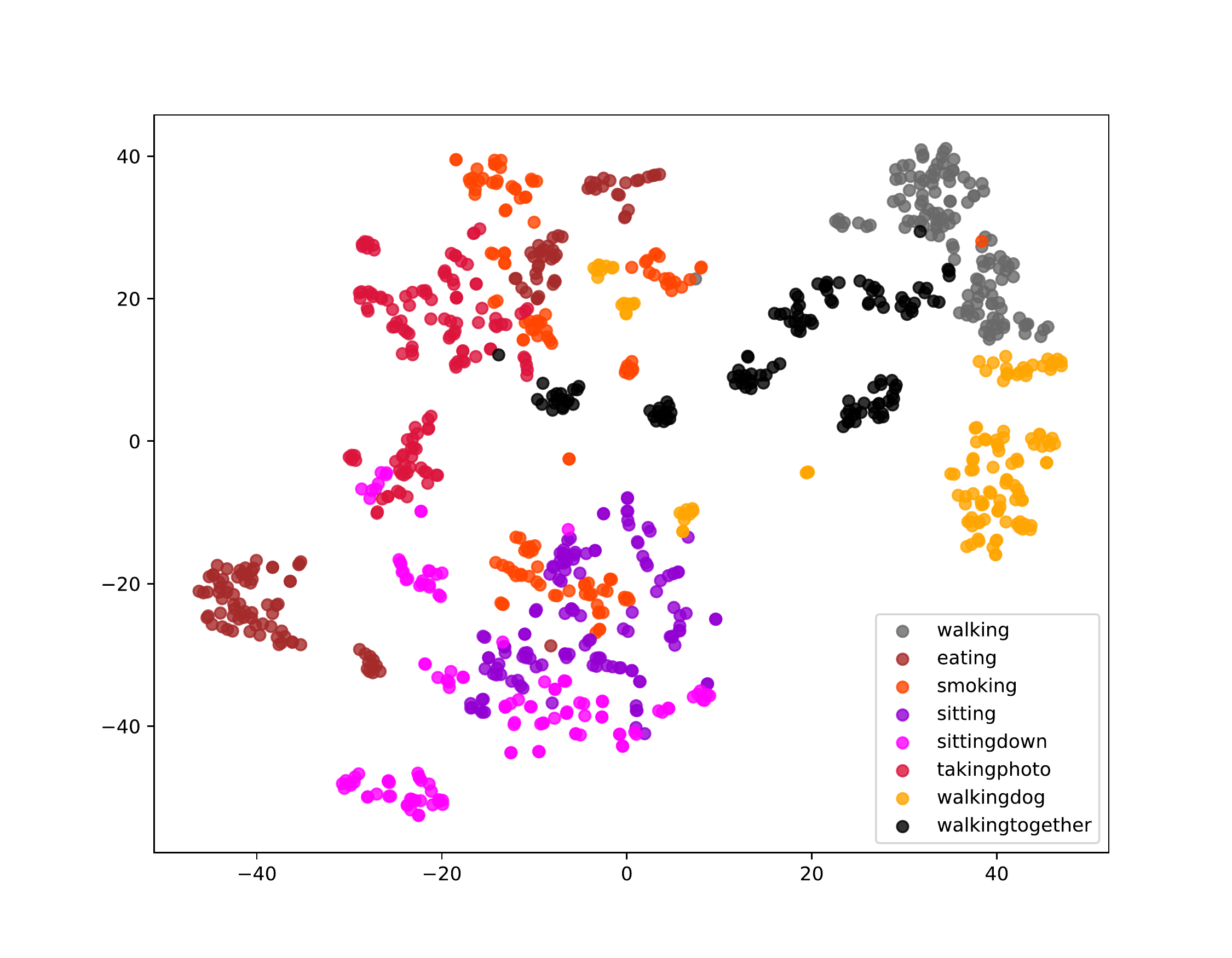}
    \vspace{-10pt}
    \caption{2D T-SNE map of learned actional graphs corresponding to 8 activities in H3.6M. Walking-related graphs are separated from sitting-related graphs with a large margin.}
    \label{fig:tsne}
    \vspace{-10pt}
\end{figure}
% \begin{figure}
% \centering
%     \includegraphics[width=8.6cm]{feature_map2.pdf}
%     \vspace{-10pt}
%     \caption{Feature maps output from backbone. The areas of translucid circles indicates the feature magnitudes. Plot (b) shows the feature maps of `clapping' and `cheer up'; Plot (c) compares the feature maps of Sym-GNN to ST-GCN on `hand waving'.}
%     \label{fig:feature_map}
%     \vspace{-10pt}
% \end{figure}
We see: 1) The joint-scale actional graphs capture some action-based long-range relations beyond direct bone-connections; 2) Some reasonable relations are captured, e.g. for `Directions', the stretched arms are correlated to other joints; 3) for motions with the same category, we tend to obtain the similar graphs; see two plots of `Walking', while different classes of motions have distinct actional graphs; see `Walking' and the other motions, where the model learns the discriminative patterns from data. 

\subsubsection{Manifolds of Joint-Scale Actional Graphs}
To verify how discriminative the patterns embedded in the joint-scale actional graphs, we visualize the low-dimension manifolds of different joint-scale actional graphs. We select 8 representative classes of actions in Human 3.6M and sample more clips from long test motion sequences. Here we treat all the joint-scale actional graphs as vectors and obtain their 2D T-SNE map; see Fig.~\ref{fig:tsne}. We see that `Walking', `Walking Dog' and `Walking Together', which have the common walking dynamics, are distributed closely, as well as `Sitting' and `Sitting Down' are clustered; however, walking-related actions and sitting-related actions are separated with a large margin; as for `Eating', `Smoking' and `Taking Photo', they have similar movements on arms, showing a new cluster.
\begin{figure*}
\centering
    \includegraphics[width=17cm]{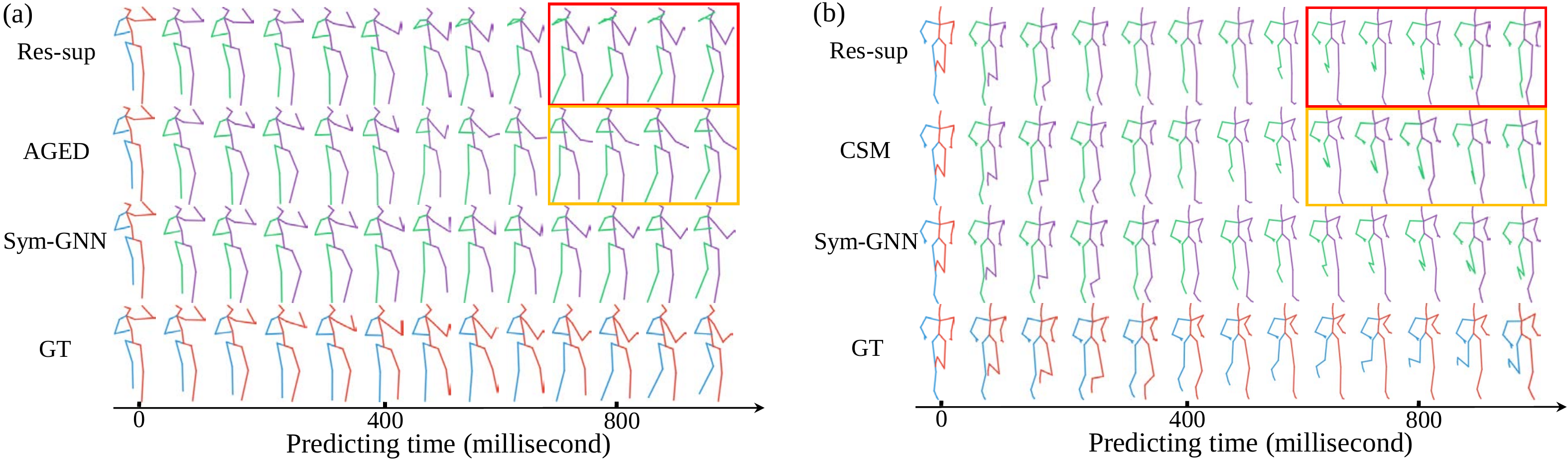}
    \caption{Visualization of motion prediction on Human 3.6M and CMU Mocap. Plot (a) shows the predictions of `Eating' in Human 3.6M and plot (b) shows the predictions of `Running' in CMU Mocap. We compare the predictions of Sym-GNN, Res-sup, AGED, and CSM with the ground truth (GT).}
    \label{fig:predictions}
    \vspace{-10pt}
\end{figure*}

% \subsubsection{Feature Responses}
% To validate the effectiveness of long-range relations built by multi-scale graphs and how the high-level features affect the tasks of action recognition and motion prediction, we visualize the feature maps at the output layer of backbone network in Fig.~\ref{fig:feature_map}, where the circle around each joint indicates magnitude of feature responses of this joint.
% Plot (a) shows feature responses of two actions `clapping' and `cheer up', where some nonfunctional joints are not much negelected, beacuse abundant long-range dependencies are built. For action recognition, all joints located throughout the body provide a global cognition, but the functional joints show stronger effects; for motion prediction, the features of all joints are preserved, leading to precise generation. Plot (b) compares the features between Sym-GNN and ST-GCN on the same action `hand waving'. ST-GCN does apply multi-layer GCNs to cover the entire spatial domain; however, the feature are weakened during the propagation and distant joints cannot interact effectively, leading to localized feature responses; however, Sym-GNN captures useful long-range dependencies for pattern learning.

\subsubsection{Predicted Sequences}
Finally, we compare the generated samples of Sym-GNN to those of Res-sup~\cite{Martinez_2017_CVPR}, AGED~\cite{Gui_2018_ECCV}, and CSM~\cite{Li_2018_CVPR} on Human 3.6M and CMU Mocap. Fig.~\ref{fig:predictions} illustrates the future poses of `Eating' and `Running' in 1000 ms with the frame interval of 80 ms, where plot (a) shows the predictions of `Eating' in Human 3.6M and plot (b) visualize `Running' in CMU Mocap. Comparing to baselines, we see that Sym-GNN provides significantly better predictions. The poses generated by Res-sup has large errors after the 600th ms (two orange boxes); AGED produces over movements for the downward hand in long-term (red box in plot (a)); CSM gives tortile poses in long-term (red box in plot (b)). But Sym-GNN completes the action accurately and reasonably.
\begin{figure}[t]
\centering
    \includegraphics[width=7cm]{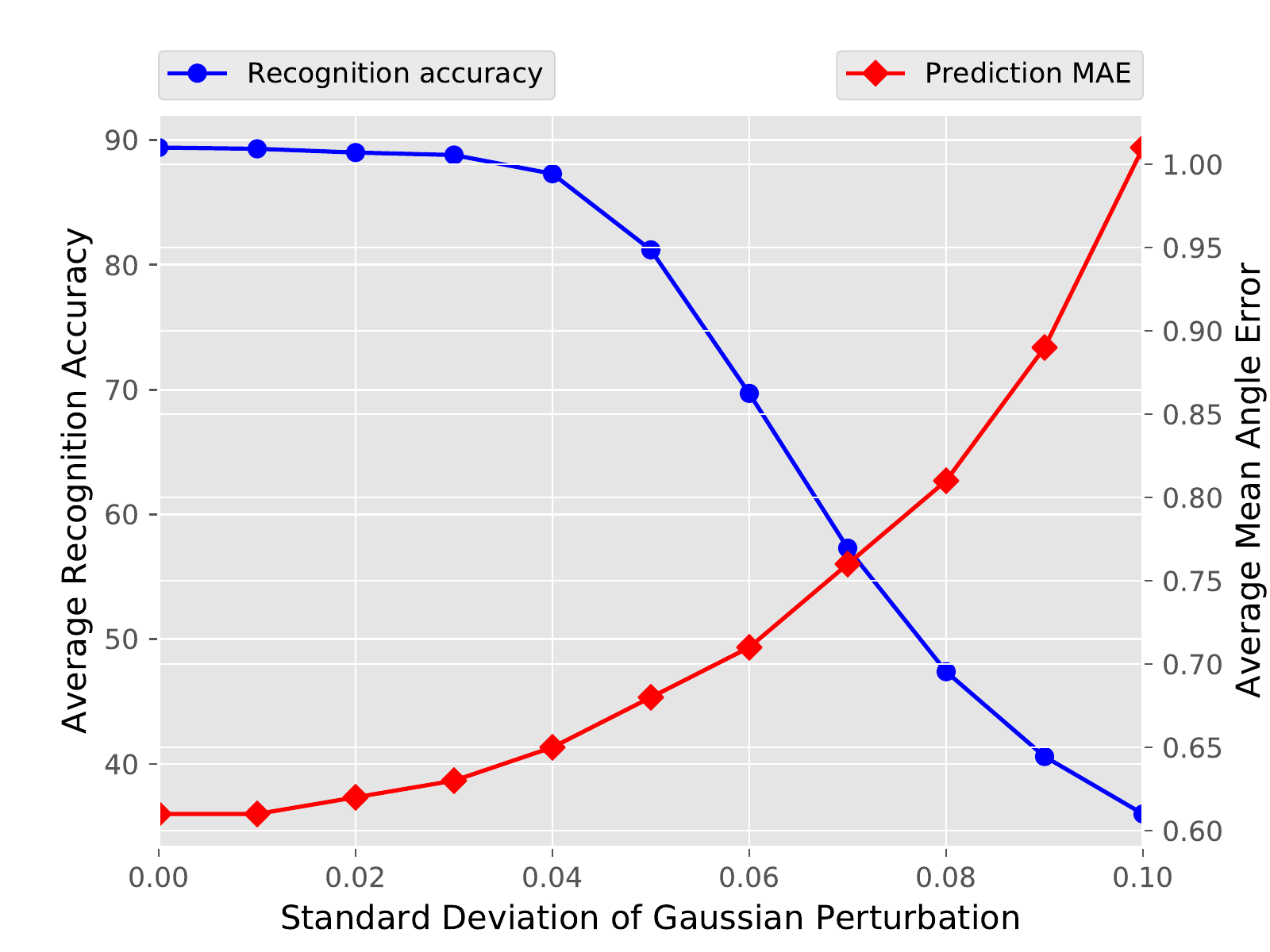}
    \vspace{-10pt}
    \caption{The recognition accuracy and prediction MAE perturbed Gaussian noises with different standard deviations.}
    \label{fig:pertub}
    \vspace{-10pt}
\end{figure}

\subsection{Stability Analysis: Robustness against Input Perturbation}
According to Theorem~\ref{thm:stability}, we present that Sym-GNN is robust against perturbation on inputs, where we calculate an upper bound of output deviation. To verify the stability, we add Gaussian noises sampled from $\mathcal{N}(0,\sigma^2)$ on input actions. We show the recognition accuracies on NTU-RGB+D (Cross-Subject) and short-term prediction MAEs on Human 3.6M with standard deviation $\sigma$ varied from $0.01$ to $0.1$. The recognition/prediction performances with different $\sigma$ are illustrated in Fig.~\ref{fig:pertub}.
We see: 1) for action recognition, Sym-GNN stays a high accuracy when the noise has $\sigma \leq 0.04$, but it tends to deteriorate due to severe perturbation when $\sigma > 0.04$; 2) for motion prediction, Sym-GNN produces precise poses when the noise has $\sigma<0.03$, but the prediction performance is degraded for larger $\sigma$. In all, Sym-GNN is robust against small perturbation.

\begin{figure}[t]
\centering
    \includegraphics[width=6.45cm]{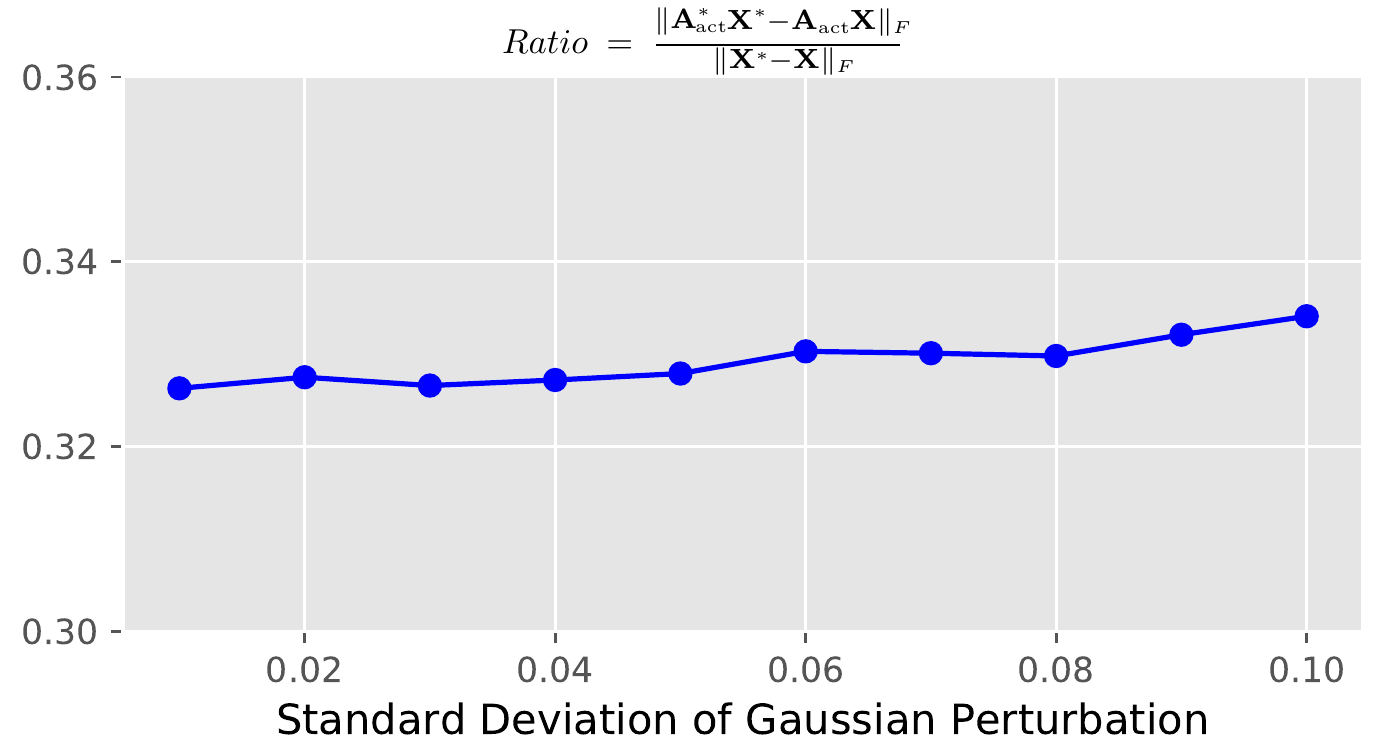}
    \vspace{-10pt}
    \caption{The ratio between the perturbations of responses and inputs with different noise standard deviations. The amplify factor $q \approx 1$, indicating that AGIM is stable.}
    \label{fig:ratio}
    \vspace{-10pt}
\end{figure}

Given two inputs $\mathbf{X}$ and $\mathbf{X}^{*}$, which satisfy $\|\mathbf{X}^{*}-\mathbf{X}\|\leq\epsilon$, we validate the assumption of $\|\mathbf{A}_{\rm act}^{*}\mathbf{X}^{*}-\mathbf{A}_{\rm act}\mathbf{X}\|_F\leq C\epsilon^q$ claimed in Theorem~\ref{thm:stability}. We calculate the ratio between the perturbations of responses and inputs; that is,
\begin{equation*}
    { Ratio} \ = \ \frac{\|\mathbf{A}_{\rm act}^{*}\mathbf{X}^{*}-\mathbf{A}_{\rm act}\mathbf{X}\|_F}
    {\|\mathbf{X}^{*}-\mathbf{X}\|_F}
\end{equation*}
Similar to Fig.~\ref{fig:pertub}, we tune the standard deviations of input noises, obtaining the corresponding $\mathbf{A}_{\rm act}^{*}$ and calculate the perturbation ratios. Fig.~\ref{fig:ratio} illustrated the $Ratio$ with different noise standard deviations. We see that the $Ratio$ is steady at around $0.33$ for $\sigma$ adjusted from $0.01$ to $0.1$, which indicates the amplify factor $q\approx1$ and the responses and inputs are mostly linearly correlated. In other words, the actional graph inference module does not amplify the perturbation and the stability of JGC is still preserved. 

\section{Conclusions}
{In this paper, we propose a novel symbiotic graph neural network (Sym-GNN), which handles action recognition and motion prediction jointly and use graph-based operations to capture action patterns. Our model consists of a backbone, an action-recognition head, and a motion-prediction head, where the two heads enhance each other. As building components in the backbone and the motion-prediction head,  graph convolution operators based on learnable joint-scale and part-scale graphs are used to extract spatial information. We conduct extensive experiments for action recognition and motion prediction with four datasets, NTU-RGB+D, Kinetics, Human 3.6M, and CMU Mocap. Experiments show that our model achieves consistently improvements compared to the previous methods.}

% \ifCLASSOPTIONcompsoc
%   \section*{Acknowledgments}
% \else
%   \section*{Acknowledgment}
% \fi
% The work is supported by the High Technology Research and Development Program of China (2015AA015801), NSFC (61521062), and STCSM (18DZ2270700).

\ifCLASSOPTIONcaptionsoff
  \newpage
\fi

\bibliographystyle{IEEEtran}
\bibliography{mybib}

% Generated by IEEEtran.bst, version: 1.12 (2007/01/11)
\begin{thebibliography}{10}
\providecommand{\url}[1]{#1}
\csname url@samestyle\endcsname
\providecommand{\newblock}{\relax}
\providecommand{\bibinfo}[2]{#2}
\providecommand{\BIBentrySTDinterwordspacing}{\spaceskip=0pt\relax}
\providecommand{\BIBentryALTinterwordstretchfactor}{4}
\providecommand{\BIBentryALTinterwordspacing}{\spaceskip=\fontdimen2\font plus
\BIBentryALTinterwordstretchfactor\fontdimen3\font minus
  \fontdimen4\font\relax}
\providecommand{\BIBforeignlanguage}[2]{{%
\expandafter\ifx\csname l@#1\endcsname\relax
\typeout{** WARNING: IEEEtran.bst: No hyphenation pattern has been}%
\typeout{** loaded for the language `#1'. Using the pattern for}%
\typeout{** the default language instead.}%
\else
\language=\csname l@#1\endcsname
\fi
#2}}
\providecommand{\BIBdecl}{\relax}
\BIBdecl

\bibitem{Li_cvpr_2019}
M.~Li, S.~Chen, X.~Chen, Y.~Zhang, Y.~Wang, and Q.~Tian, ``Actional-structural
  graph convolutional networks for skeleton-based action recognition,'' in
  \emph{The IEEE Conference on Computer Vision and Pattern Recognition (CVPR)},
  June 2019.

\bibitem{Gaur_iccv_2011}
U.~Gaur, Y.~Zhu, B.~Song, and A.~Roy-Chowdhury, ``A âstring of feature
  graphsâ model for recognition of complex activities in natural videos,''
  in \emph{The IEEE International Conference on Computer Vision (ICCV)},
  November 2011, pp. 2595--2602.

\bibitem{Huang_eccv_2014}
D.~Huang and K.~Kitani, ``Action-reaction: Forecasting the dynamics of human
  interaction,'' in \emph{The European Conference on Computer Vision (ECCV)},
  July 2014, pp. 489--504.

\bibitem{gui-2018-110272}
L.~Gui, K.~Zhang, Y.~Wang, X.~Liang, J.~Moura, and M.~Veloso, ``Teaching robots
  to predict human motion,'' in \emph{IEEE International Conference on
  Intelligent Robots and Systems (IROS)}, October 2018.

\bibitem{6942210}
S.~{Mathe} and C.~{Sminchisescu}, ``Actions in the eye: Dynamic gaze datasets
  and learnt saliency models for visual recognition,'' \emph{IEEE Transactions
  on Pattern Analysis and Machine Intelligence}, vol.~37, no.~7, pp.
  1408--1424, July 2015.

\bibitem{Shi_2018_ECCV}
Y.~Shi, B.~Fernando, and R.~Hartley, ``Action anticipation with rbf kernelized
  feature mapping rnn,'' in \emph{The European Conference on Computer Vision
  (ECCV)}, September 2018, pp. 301--317.

\bibitem{Wang_2018_ECCV}
D.~Wang, W.~Ouyang, W.~Li, and D.~Xu, ``Dividing and aggregating network for
  multi-view action recognition,'' in \emph{The European Conference on Computer
  Vision (ECCV)}, September 2018, pp. 451--476.

\bibitem{8327922}
X.~{Liang}, K.~{Gong}, X.~{Shen}, and L.~{Lin}, ``Look into person: Joint body
  parsing pose estimation network and a new benchmark,'' \emph{IEEE
  Transactions on Pattern Analysis and Machine Intelligence}, vol.~41, no.~4,
  pp. 871--885, April 2019.

\bibitem{ijcai_ChaoLi}
C.~Li, Q.~Zhong, D.~Xie, and S.~Pu, ``Co-occurrence feature learning from
  skeleton data for action recognition and detection with hierarchical
  aggregation,'' in \emph{IJCAI}, July 2018, pp. 786--792.

\bibitem{Vemulapalli_2014_CVPR}
R.~Vemulapalli, F.~Arrate, and R.~Chellappa, ``Human action recognition by
  representing 3d skeletons as points in a lie group,'' in \emph{The IEEE
  Conference on Computer Vision and Pattern Recognition (CVPR)}, June 2014, pp.
  588--595.

\bibitem{Fernando_2015_CVPR}
B.~Fernando, E.~Gavves, J.~M. Oramas, A.~Ghodrati, and T.~Tuytelaars,
  ``Modeling video evolution for action recognition,'' in \emph{The IEEE
  Conference on Computer Vision and Pattern Recognition (CVPR)}, June 2015, pp.
  5378--5387.

\bibitem{Du_2015_CVPR}
Y.~Du, W.~Wang, and L.~Wang, ``Hierarchical recurrent neural network for
  skeleton based action recognition,'' in \emph{The IEEE Conference on Computer
  Vision and Pattern Recognition (CVPR)}, June 2015, pp. 1110--1118.

\bibitem{vis_cnn}
M.~Liu, H.~Liu, and C.~Chen, ``Enhanced skeleton visualization for view
  invariant human action recognition,'' in \emph{Pattern Recognition}, vol.~68,
  August 2017, pp. 346--362.

\bibitem{AAAI1817135}
S.~Yan, Y.~Xiong, and D.~Lin, ``Spatial temporal graph convolutional networks
  for skeleton-based action recognition,'' in \emph{AAAI Conference on
  Artificial Intelligence}, February 2018, pp. 7444--7452.

\bibitem{Lehrmann_2014_CVPR}
A.~Lehrmann, P.~Gehler, and S.~Nowozin, ``Efficient nonlinear markov models for
  human motion,'' in \emph{The IEEE Conference on Computer Vision and Pattern
  Recognition (CVPR)}, June 2014, pp. 1314--1321.

\bibitem{Jain_2016_CVPR}
A.~Jain, A.~Zamir, S.~Savarese, and A.~Saxena, ``Structural-rnn: Deep learning
  on spatio-temporal graphs,'' in \emph{The IEEE Conference on Computer Vision
  and Pattern Recognition (CVPR)}, June 2016, pp. 5308--5317.

\bibitem{Martinez_2017_CVPR}
J.~Martinez, M.~Black, and J.~Romero, ``On human motion prediction using
  recurrent neural networks,'' in \emph{The IEEE Conference on Computer Vision
  and Pattern Recognition (CVPR)}, July 2017, pp. 4674--4683.

\bibitem{Gui_2018_ECCV}
L.~Gui, Y.~Wang, X.~Liang, and J.~Moura, ``Adversarial geometry-aware human
  motion prediction,'' in \emph{The European Conference on Computer Vision
  (ECCV)}, September 2018, pp. 786--803.

\bibitem{cvpr_wang_2012}
J.~Wang, Z.~Liu, Y.~Wu, and J.~Yuan, ``Mining actionlet ensemble for action
  recognition with depth cameras,'' in \emph{IEEE International Conference on
  Computer Vision and Pattern Recognition (CVPR)}, June 2012, pp. 1290--1297.

\bibitem{ijcai_Hussein_2013}
M.~Hussein, M.~Torki, M.~Gowayyed, and M.~El-Saban, ``Human action recognition
  using a temporal hierarchy of covariance descriptors on 3d joint locations,''
  in \emph{IJCAI}, August 2013, pp. 2466--2472.

\bibitem{NIPS2005_2783}
J.~Wang, A.~Hertzmann, and D.~Fleet, ``Gaussian process dynamical models,'' in
  \emph{Advances in Neural Information Processing Systems (NeurIPS)}, December
  2006, pp. 1441--1448.

\bibitem{Taylor_2007_NIPS}
G.~Taylor, G.~Hinton, and S.~Roweis, ``Modeling human motion using binary
  latent variables,'' in \emph{Advances in Neural Information Processing
  Systems (NeurIPS)}, Decemeber 2007.

\bibitem{NIPS2008_3567}
I.~Sutskever, G.~E. Hinton, and G.~W. Taylor, ``The recurrent temporal
  restricted boltzmann machine,'' in \emph{Advances in Neural Information
  Processing Systems (NeurIPS)}, 2009, pp. 1601--1608.

\bibitem{a8014941}
T.~Kim and A.~Reiter, ``Interpretable 3d human action analysis with temporal
  convolutional networks,'' in \emph{IEEE Conference on Computer Vision and
  Pattern Recognition Workshops (CVPRW)}, July 2017, pp. 1623--1631.

\bibitem{quater}
D.~Pavllo, D.~Grangier, and M.~Auli, ``Quaternet: A quaternion-based recurrent
  model for human motion,'' in \emph{British Machine Vision Converence (BMVC)},
  September 2018, pp. 1--14.

\bibitem{Gui_2018_ECCV2}
L.~Gui, Y.~Wang, D.~Ramanan, and J.~Moura, ``Few-shot human motion prediction
  via meta-learning,'' in \emph{The European Conference on Computer Vision
  (ECCV)}, September 2018, pp. 432--450.

\bibitem{AAAI_Kundu}
J.~Kundu, M.~Gor, and R.~Babu, ``Bihmp-gan: Bidirectional 3d human motion
  prediction gan,'' in \emph{AAAI Conference on Artificial Intelligence},
  February 2019.

\bibitem{Si_2018_ECCV}
C.~Si, Y.~Jing, W.~Wang, L.~Wang, and T.~Tan, ``Skeleton-based action
  recognition with spatial reasoning and temporal stack learning,'' in
  \emph{The European Conference on Computer Vision (ECCV)}, Sept 2018.

\bibitem{Si_2019_CVPR}
C.~Si, W.~Chen, W.~Wang, L.~Wang, and T.~Tan, ``An attention enhanced graph
  convolutional lstm network for skeleton-based action recognition,'' in
  \emph{The IEEE Conference on Computer Vision and Pattern Recognition (CVPR)},
  June 2019.

\bibitem{AAAI_Guo}
X.~Guo and J.~Choi, ``Human motion prediction via learning local structure
  representations and temporal dependencies,'' in \emph{AAAI Conference on
  Artificial Intelligence}, February 2019.

\bibitem{Li_2018_CVPR}
C.~Li, Z.~Zhang, W.~Sun~Lee, and G.~Hee~Lee, ``Convolutional sequence to
  sequence model for human dynamics,'' in \emph{The IEEE Conference on Computer
  Vision and Pattern Recognition (CVPR)}, June 2018, pp. 5226--5234.

\bibitem{Shahroudy_2016_CVPR}
A.~Shahroudy, J.~Liu, T.-T. Ng, and G.~Wang, ``Ntu rgb+d: A large scale dataset
  for 3d human activity analysis,'' in \emph{The IEEE Conference on Computer
  Vision and Pattern Recognition (CVPR)}, June 2016, pp. 1010--1019.

\bibitem{h36m}
C.~Ionescu, P.~Papaca, V.~Olaru, and C.~Sminchisescu, ``Human3.6m: Large scale
  datasets and predictive methods for 3d human sensing in natural
  environments.'' \emph{{IEEE} Transactions on Pattern Analysis and Machine
  Intelligence (TPAMI)}, vol.~36, no.~7.

\bibitem{Liu_2016_eccv}
J.~Liu, A.~Shahroudy, D.~Xu, and G.~Wang, ``Spatio-temporal lstm with trust
  gates for 3d human action recognition,'' in \emph{The European Conference on
  Computer Vision (ECCV)}, October 2016, pp. 816--833.

\bibitem{2sAGCN}
L.~Shi, Y.~Zhang, J.~Cheng, and H.~Lu, ``Two-stream adaptive graph
  convolutional networks for skeleton-based action recognition,'' in \emph{The
  IEEE Conference on Computer Vision and Pattern Recognition (CVPR)}, June
  2019, pp. 12\,026--12\,035.

\bibitem{DGNN}
------, ``Skeleton-based action recognition with directed graph neural
  networks,'' in \emph{The IEEE Conference on Computer Vision and Pattern
  Recognition (CVPR)}, June 2019, pp. 7912--7921.

\bibitem{STGR-GCN}
B.~Li, X.~Li, Z.~Zhang, and F.~Wu, ``Spatio-temporal graph routing for
  skeleton-based action recognition,'' in \emph{AAAI Conference on Artificial
  Intelligence}, February 2019, pp. 8561--8568.

\bibitem{motif-GCN}
Y.~Wen, L.~Gao, H.~Fu, F.~Zhang, and S.~Xia, ``Graph cnns with motif and
  variable temporal block for skeleton-based action recognition,'' in
  \emph{AAAI Conference on Artificial Intelligence}, February 2019, pp.
  8989--8996.

\bibitem{nips_2000}
V.~Pavlovic, J.~Rehg, and J.~MacCormick, ``Learning switching linear models of
  human motion,'' in \emph{Advances in Neural Information Processing Systems
  (NeurIPS)}, 2001, pp. 942--948.

\bibitem{Walker_2017_ICCV}
J.~Walker, K.~Marino, A.~Gupta, and M.~Hebert, ``The pose knows: Video
  forecasting by generating pose futures,'' in \emph{The IEEE International
  Conference on Computer Vision (ICCV)}, October 2017, pp. 3332--3341.

\bibitem{NIPS2016_6552}
T.~Xue, J.~Wu, K.~Bouman, and B.~Freeman, ``Visual dynamics: Probabilistic
  future frame synthesis via cross convolutional networks,'' in \emph{Advances
  in Neural Information Processing Systems (NeurIPS)}, December 2016, pp.
  91--99.

\bibitem{Fragkiadaki_2015_ICCV}
K.~Fragkiadaki, S.~Levine, P.~Felsen, and J.~Malik, ``Recurrent network models
  for human dynamics,'' in \emph{The IEEE International Conference on Computer
  Vision (ICCV)}, December 2015, pp. 4346--4354.

\bibitem{Verma_2018_CVPR}
N.~Verma, E.~Boyer, and J.~Verbeek, ``Feastnet: Feature-steered graph
  convolutions for 3d shape analysis,'' in \emph{The IEEE Conference on
  Computer Vision and Pattern Recognition (CVPR)}, June 2018, pp. 2598--2606.

\bibitem{valsesia2018learning}
D.~Valsesia, G.~Fracastoro, and E.~Magli, ``Learning localized generative
  models for 3d point clouds via graph convolution,'' in \emph{International
  Conference on Learning Representations (ICLR)}, May 2019, pp. 1--15.

\bibitem{Li2018learning}
Y.~Li, D.~Tarlow, M.~Brockschmidt, and R.~Zemel, ``Gated graph sequence neural
  networks,'' in \emph{International Conference on Learning Representations
  (ICLR)}, May 2016, pp. 1--20.

\bibitem{NIPS2016_6081}
M.~Defferrard, X.~Bresson, and P.~Vandergheynst, ``Convolutional neural
  networks on graphs with fast localized spectral filtering,'' in
  \emph{Advances in Neural Information Processing Systems (NeurIPS)}, December
  2016, pp. 3844--3852.

\bibitem{kipf_iclr2017}
T.~Kipf and M.~Welling, ``Semi-supervised classification with graph
  convolutional networks,'' in \emph{International Conference on Learning
  Representations (ICLR)}, April 2017, pp. 1--14.

\bibitem{NIPS2017_6703}
W.~Hamilton, Z.~Ying, and J.~Leskovec, ``Inductive representation learning on
  large graphs,'' in \emph{Advances in Neural Information Processing Systems
  (NeurIPS)}, December 2017, pp. 1024--1034.

\bibitem{ilprints361}
S.~Brin and L.~Page, ``The anatomy of a large-scale hypertextual web search
  engine,'' in \emph{International World-Wide Web Conference (WWW)}, May 1998,
  pp. 107--117.

\bibitem{DBLP:journals/tsp/ChenTFVK18}
S.~Chen, D.~Tian, C.~Feng, A.~Vetro, and J.~Kova\v{c}evi\'c, ``Fast resampling
  of three-dimensional point clouds via graphs,'' \emph{{IEEE} Transactions on
  Signal Processing (TSP)}, vol.~66, no.~3, pp. 666--681, 2018.

\bibitem{MGDA}
J.~Desideri, ``Multiple-gradient descent algorithm (mgda) for multiobjective
  optimization.'' \emph{Comptes Rendus Mathematique}, vol. 350, no.~5, pp.
  313--318, May 2012.

\bibitem{SGD}
L.~Bottou, ``Large-scale machine learning with stochastic gradient descent,''
  in \emph{International Conference on Computational Statistics (COMPSTAT)},
  August 2010, pp. 177--187.

\bibitem{Cao_2017_CVPR}
Z.~Cao, T.~Simon, S.~Wei, and Y.~Sheikh, ``Realtime multi-person 2d pose
  estimation using part affinity fields,'' in \emph{The IEEE Conference on
  Computer Vision and Pattern Recognition (CVPR)}, July 2017, pp. 7291--7299.

\bibitem{Tang_2018_CVPR}
Y.~Tang, Y.~Tian, J.~Lu, P.~Li, and J.~Zhou, ``Deep progressive reinforcement
  learning for skeleton-based action recognition,'' in \emph{The IEEE
  Conference on Computer Vision and Pattern Recognition (CVPR)}, June 2018, pp.
  5323--5332.

\bibitem{GhoshSAH17}
P.~Ghosh, J.~Song, E.~Aksan, and O.~Hilliges, ``Learning human motion models
  for long-term predictions,'' \emph{CoRR}, vol. abs/1704.02827, 2017.

\bibitem{abs-1810-09676}
H.~Chiu, E.~Adeli, B.~Wang, D.~Huang, and J.~Niebles, ``Action-agnostic human
  pose forecasting,'' \emph{CoRR}, vol. abs/1810.09676, 2018.

\bibitem{Gopalakrishnan_2019_CVPR}
A.~Gopalakrishnan, A.~Mali, D.~Kifer, L.~Giles, and A.~Ororbia, ``A neural
  temporal model for human motion prediction,'' in \emph{The IEEE Conference on
  Computer Vision and Pattern Recognition (CVPR)}, June 2019, pp.
  12\,116--12\,125.

\bibitem{NIPS2018_7334}
O.~Sener and V.~Koltun, ``Multi-task learning as multi-objective
  optimization,'' in \emph{Advances in Neural Information Processing Systems
  (NeurIPS)}, December 2018, pp. 527--538.

\end{thebibliography}

\begin{IEEEbiography}[{\includegraphics[width=1in,height=1.25in,clip,keepaspectratio]{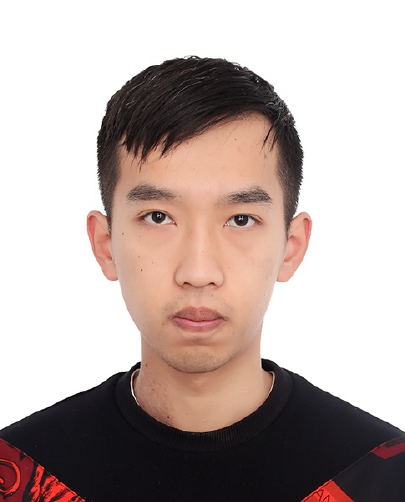}}]{Maosen Li} recieved the B.E. degree in optical engineering from University of Electronic Science and Technology of China (UESTC), Chengdu, China, in 2017. He is working toward the Ph.D. degree at Cooperative Meidianet Innovation Center in Shanghai Jiao Tong University since 2017. His research interests include computer vision, machine learning, graph representation learning, and video analysis. He is a student member of the IEEE. \end{IEEEbiography}

\begin{IEEEbiography}[{\includegraphics[width=1in,height=1.25in,clip,keepaspectratio]{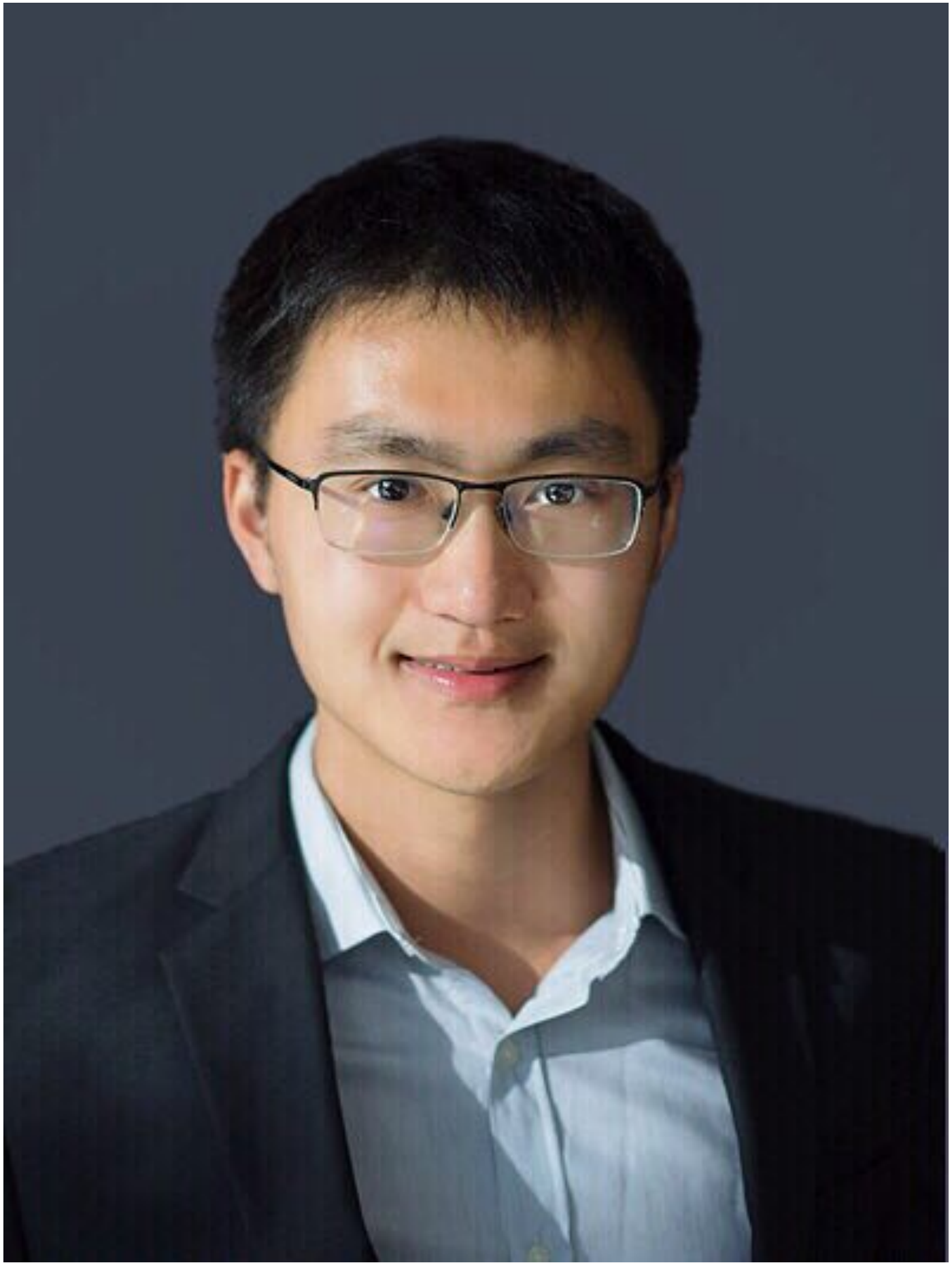}}]{Siheng Chen} is a research scientist at Mitsubishi Electric Research Laboratories (MERL). Before that, he was an autonomy engineer at Uber Advanced Technologies Group, working on the perception and prediction systems of self-driving cars. Before joining Uber, he was a postdoctoral research associate at Carnegie Mellon University. Chen received the doctorate in Electrical and Computer Engineering from Carnegie Mellon University in 2016, where he also received two masters degrees in Electrical and Computer Engineering and Machine Learning, respectively. He received his bachelor's degree in Electronics Engineering in 2011 from Beijing Institute of Technology, China. Chen was the recipient of the 2018 IEEE Signal Processing Society Young Author Best Paper Award. His coauthored paper received the Best Student Paper Award at IEEE GlobalSIP 2018. His research interests include graph signal processing, graph neural networks and 3D computer vision. \end{IEEEbiography}

\begin{IEEEbiography}[{\includegraphics[width=1in,height=1.25in,clip,keepaspectratio]{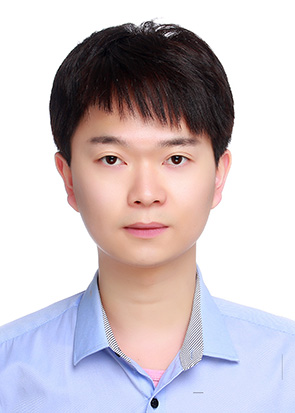}}]{Xu Chen} received the B.E. degree in electronics engineering from Xidian University in 2016. He is working toward the Ph.D. degree at Cooperative Meidianet Innovation Center in Shanghai Jiao Tong University since 2016. He is now a dual Ph.D. student of Shanghai Jiao Tong University and University of Technology, Sydney. His research interests include machine learning, graph representation learning, recommendation systems, and computer vision. \end{IEEEbiography}

\begin{IEEEbiography}[{\includegraphics[width=1in,height=1.25in,clip,keepaspectratio]{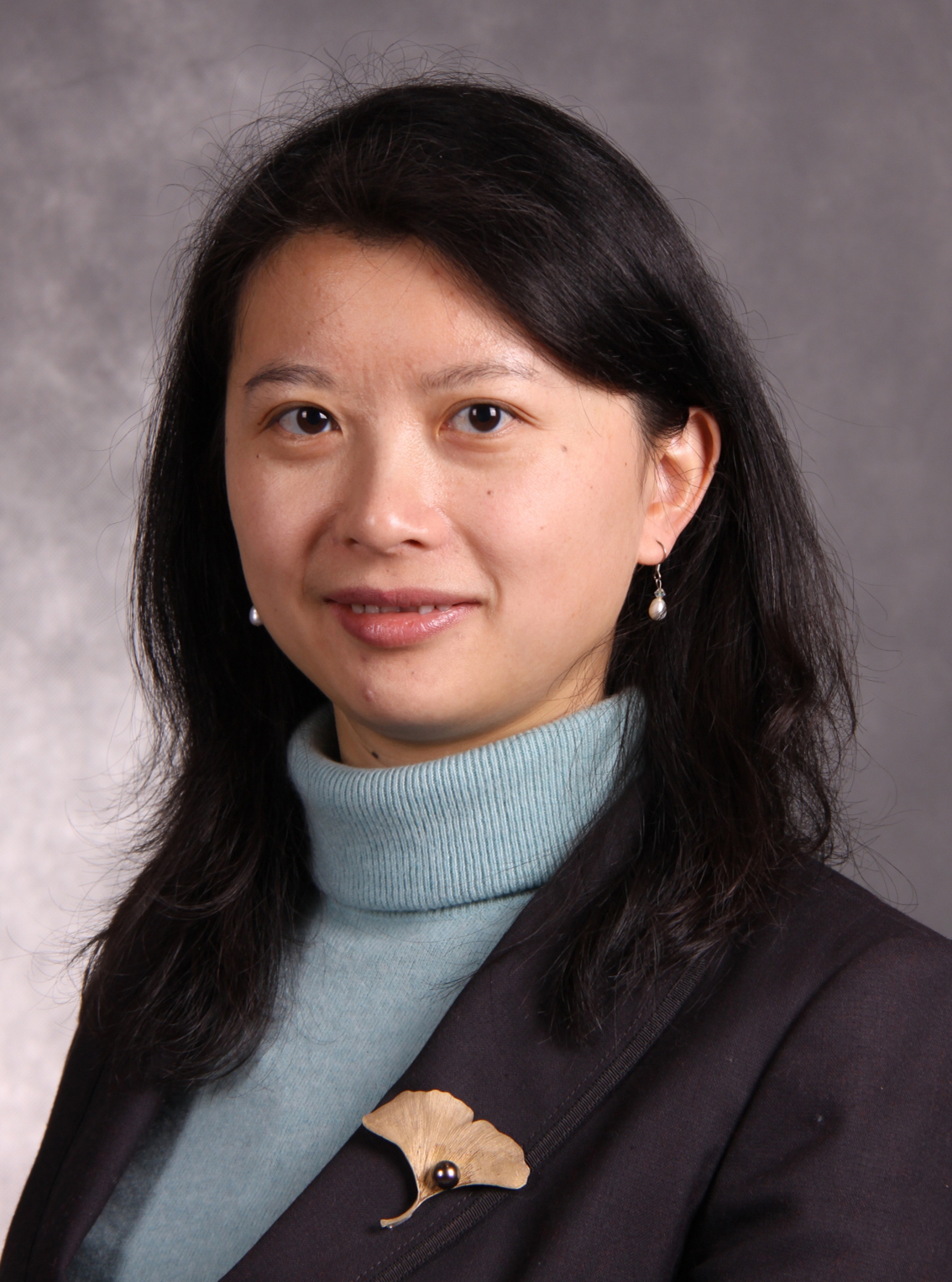}}]{Ya Zhang} received the B.S. degree from Tsinghua University and the Ph.D. degree in information sciences and technology from the Pennsylvania State University. Since March 2010, she has been a professor with Cooperative Medianet Innovation Center, Shanghai Jiao Tong University. Prior to that, she worked with Lawrence Berkeley National Laboratory, University of Kansas, and Yahoo! Labs. Her research interest is mainly on data mining and machine learning, with applications to information retrieval, web mining, and multimedia analysis. She is a member of the IEEE. \end{IEEEbiography}

\begin{IEEEbiography}[{\includegraphics[width=1in,height=1.25in,clip,keepaspectratio]{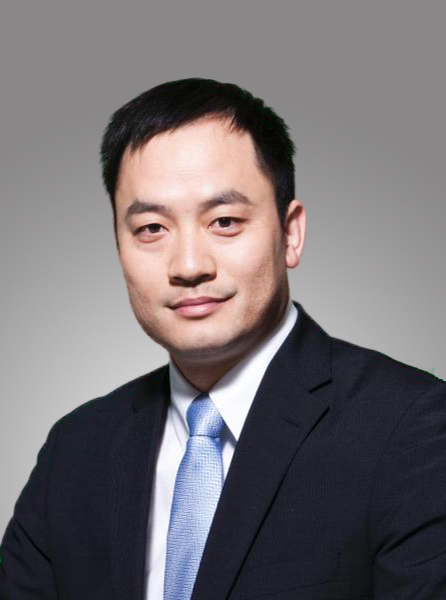}}]{Yanfeng Wang} received the B.E. degree in information engineering from the University of PLA, Beijing, China, and the M.S. and Ph.D. degrees in business management from the Antai College of Economics and Management, Shanghai Jiao Tong University, Shanghai, China. He is currently the Vice Director of Cooperative Medianet Innovation Center and also the Vice Dean of the School of Electrical and Information Engineering with Shanghai Jiao Tong University. His research interest mainly include media big data and emerging commercial applications of information technology. \end{IEEEbiography}

\begin{IEEEbiography}[{\includegraphics[width=1in,height=1.25in,clip,keepaspectratio]{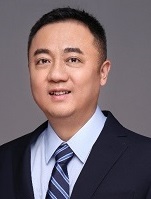}}]{Qi Tian} is currently the Chief Scientist of Computer Vision in Huawei Noah's Ark Laboratory, a Full Professor with the Department of Computer Science, University of Texas at San Antonio (UTSA). He was a tenured Associate Professor from 2008-2012 and a tenure-track Assistant Professor from 2002-2008. During 2008-2009, he took one-year Faculty Leave at Microsoft Research Asia (MSRA) as Lead Researcher in the Media Computing Group. Dr.  Tian  received  his  Ph.D. in ECE from University of Illinois at Urbana-Champaign (UIUC) in 2002 and received his B.E. in Electronic Engineering from Tsinghua University in 1992 and M.S. in ECE from Drexel University in 1996.  Dr. Tian's research interests include multimedia information retrieval, computer vision, pattern recognition and published over 360 refereed journal and conference papers. He was the co-author of a Best Paper in ACM ICMR 2015, a Best Paper in PCM 2013, a Best Paper in MMM 2013, a Best Paper in ACM ICIMCS 2012, a Top 10\% Paper Award in MMSP 2011, a Best Student Paper in ICASSP 2006, and co-author of a Best Student Paper Candidate in ICME 2015, and a Best Paper Candidate in PCM 2007. Dr. Tian received 2017 UTSA President's Distinguished Award for research Achievement, 2016 UTSA Innovation Award, 2014 Research Achievement Awards from College of Science, UTSA, 2010 Google  Faculty  Award,  and  2010  ACM  Service  Award.  He  is  the associate editor of many journals and in the Editorial Board of Journal  of   Multimedia   (JMM)   and   Journal   of   Machine   Vision and Applications (MVA). He is a fellow of the IEEE\end{IEEEbiography}

\appendices
\section{Proof of Theorem 1}
Here, we prove the proposed theorem 1; that is the activated JGC operator is robust against the input perturbation. 
\label{apx: proof1}
\begin{prf}
We first bound the discrepancy after joint-scale graph convolution. We have
\begin{eqnarray}
&& \left\| {\rm JGC} \left(\mathbf{X}^{*} \right) - {\rm JGC} \left(\mathbf{X} \right) \right\|_F 
\nonumber \\ 
 & = &
\| \lambda_{\rm act}\left( \mathbf{A}^{*}_{\rm act} \mathbf{X}^{*} - \mathbf{A}_{\rm act} \mathbf{X} \right) \mathbf{W}_{\rm act}^{\top} 
\nonumber \\
&& + \sum_{\gamma=1}^{\Gamma}\mathbf{A}_{\rm str}^{(\gamma)} \left(\mathbf{X}^{*} - \mathbf{X} \right) \mathbf{W}_{\rm str}^{(\gamma)\top}\|_F
\nonumber \\
 & \stackrel{(a)}{\leq}   & 
\| \lambda_{\rm act}\left( \mathbf{A}^{*}_{\rm act} \mathbf{X}^{*} - \mathbf{A}_{\rm act} \mathbf{X} \right) \mathbf{W}_{\rm act}^{\top}\|_F
\nonumber \\
&& +
 \sum_{\gamma=1}^{\Gamma} \| \left({\bf D}^{-1}{\bf A}\right)^{\gamma} \odot \mathbf{M}^{(\gamma)} \left(\mathbf{X}^{*} - \mathbf{X} \right) \mathbf{W}_{\rm str}^{(\gamma)\top}\|_F
\nonumber \\
 & \stackrel{(b)}{\leq}  & \lambda_{\rm act}  \|  \mathbf{A}^{*}_{\rm act} \mathbf{X}^{*} - \mathbf{A}_{\rm act}  \mathbf{X}\|_F \|\mathbf{W}_{\rm act}^{\top}\|_F
\nonumber \\
&& + \sum_{\gamma=1}^{\Gamma} \| \left({\bf D}^{-1}{\bf A}\right)^{\gamma} \odot \mathbf{M}^{(\gamma)} \|_F \|\mathbf{X}^{*} - \mathbf{X} \|_F  \| \mathbf{W}_{\rm str}^{(\gamma)\top}\|_F
\nonumber \\ \nonumber
& \stackrel{(c)}{\leq}  & 
\lambda_{\rm act} \sqrt{D_{\bf x}D_{\bf y}}\mu_{\rm act} C \epsilon^q 
 \\
&&~+ \epsilon \sum_{\gamma=1}^{\Gamma}  \sqrt{\left\| {\bf A}^{\gamma} \right\|_0} \eta^{(\gamma)} \sqrt{D_{\bf x}D_{\bf y}} \mu_{\rm str}^{(\gamma)},
\label{eq:proof1}
\end{eqnarray}
where (a) follows from the norm triangle inequality; (b) follows from the norm sub-multiplicativity and (c) follows from the assumptions.

We next show that ${\rm ReLU}(x) = \max(0, x)$ is contractive; that is,
\begin{equation}
\begin{aligned}
   &  |{\rm ReLU}(x^{*})-{\rm ReLU}(x)| \\
     = & \left\{
\begin{array}{rcl}
|0~-x|~(\leq|x^{*}-x|)    &      & {x\geq0,~ x^{*}\leq0}\\
|0~-0|~(\leq|x^{*}-x|)     &      & {x\geq0,~ x^{*}\geq0}\\
|x^{*}-0|~(\leq|x^{*}-x|)     &      & {x\leq0,~ x^{*}\geq0}\\
|x^{*}-x|~(=|x^{*}-x|)     &      & {x\leq0,~ x^{*}\leq0}
\end{array} \right.
\\
 \leq & |x^{*}-x|. 
\end{aligned}
\label{eq:proof2}
\end{equation}
Therefore, we obtain
\begin{eqnarray*}
&& \left\|\mathbf{Y}^{*} - \mathbf{Y}\right\|_F
\\
& = & \left\| {\rm ReLU} \left( {\rm JGC} \left(\mathbf{X}^{*} \right) \right) -
{\rm ReLU} \left( {\rm JGC} \left(\mathbf{X} \right)\right) \right\|_F 
\\
& \stackrel{(a)}{\leq}  & \left\| {\rm JGC} \left(\mathbf{X}^{*} \right)  -  {\rm JGC} \left(\mathbf{X} \right)\right\|_F 
\\
& \stackrel{(b)}{\leq}  & 
\lambda_{\rm act} \sqrt{D_{\bf x}D_{\bf y}}\mu_{\rm act} C \epsilon^q
\\
&& + \epsilon \sum_{\gamma=1}^{\Gamma} \sqrt{\| {\bf A}^{\gamma} \|_0} \eta^{(\gamma)} \sqrt{D_{\bf x}D_{\bf y}} \mu_{\rm str}^{(\gamma)}
\\
& \stackrel{(c)}{=}  & 
\sqrt{3 D_{\bf y}} \bigg( \epsilon^q  \lambda_{\rm act} \mu_{\rm act} C  + 
 \epsilon  \sum_{\gamma=1}^{\Gamma} \sqrt{\left\| {\bf A}^{\gamma} \right\|_0} \eta^{(\gamma)}  \mu_{\rm str}^{(\gamma)} \bigg)
\\
& = & O\left( \max \left( \epsilon^q, \epsilon \right) \right),
\end{eqnarray*}
where (a) follows from~\eqref{eq:proof2}; (b) follows from~\eqref{eq:proof1}; and (c) follows the input feature dimension $D_{\bf x}=3$.
\QEDB
\end{prf}

\section{Training Algorithm}
To optimize the multi-tasking model, the KKT conditions for both backbone network and task-specific heads are stated as
\begin{itemize}
    \item The convex sum, $\lambda \nabla_{{\bm \theta}_{\rm bk}}\mathcal{L}_{\rm recg} + (1-\lambda) \nabla_{{\bm \theta}_{\rm bk}}\mathcal{L}_{\rm pred}=0$, where $0\leq\lambda\leq1$;
    \item For the task of action recognition and motion prediction, there is $\nabla_{{\bm \theta}_{\rm recg}}\mathcal{L}_{\rm recg}=0$ and $\nabla_{{\bm \theta}_{\rm pred}}\mathcal{L}_{\rm pred}=0$.
\end{itemize}
We see that the KKT conditions contain two gradient constraints model parameters. We note that the KKT condition is the necessity of the optimal solutions for our method; that is, any solution which satisifies the KKT conditions is a possible optimal solution. To find the appropriate $\lambda$, we attempt to match the first condition, leading to a stationary point.
Here we compute,
\begin{equation}
\label{eq:tune_lambda}
\lambda^{\star} = 
\mathop{\arg\min}_{\lambda}
\{\|\lambda \nabla_{{\bm \theta}_{\rm bk}}\mathcal{L}_{\rm recg}
+(1-\lambda)\nabla_{{\bm \theta}_{\rm bk}}\mathcal{L}_{\rm pred}\|_2^2\},
\end{equation}
where ${\bm \theta}_{\rm bk}$ denotes the trainable parameters of backbone, including AGIM and J-GTC blocks. To calculate $\lambda^{\star}$ from~\eqref{eq:tune_lambda}, we employ the mechanism of multi-objective optimization from~\cite{NIPS2018_7334} to adjust $\lambda$ adaptatively during training. The overall loss of Sym-GNN is
\begin{equation*}
\mathcal{L}=\lambda^{\star}\mathcal{L}_{\rm recg}+(1-\lambda^{\star})\mathcal{L}_{\rm pred}.
\end{equation*}
Note that $\lambda^{\star}$ in~\eqref{eq:tune_lambda} is optimized in each iteration.

\section{Network Architecture}
Here we present the network structure with more details. We list the sizes of parameters and corresponding operations of the actional graph inference module (AGIM), backbone networks of the proposed Sym-GNN.

\subsection{Actional Graph Inference Module}
As an important component in the Sym-GNN, the actional graph inference module (AGIM) is employed to learn the action-based correlations among different moving joints. We propagate the features of joints and arbitrary links to aggregate long-range joint feature for relation capturing and action graph estimation. The structure of AGLM is presented in Table~\ref{tab:AGLM}, 
\begin{table}[htb]
    \centering
    \caption{The structure of the AGIM in Sym-GNN model.}
    \small
    \setlength{\tabcolsep}{2.4mm}{
    \begin{tabular}{c|c|c}
    \hline
        Step & Shape \& Operations & Implementation\\ \hline
        \multirow{6}{*}{1} & 147 $(3\times49)$-128-relu & \multirow{2}{*}{joint feature}\\
        ~ & -dropout-128-relu-bn & ~\\ \cline{2-3}
        ~ & $M$ joints to ($M^2-M$) links & joint concat \\ \cline{2-3}
        ~ & 256-128-relu & \multirow{2}{*}{edge feature} \\
        ~ & -dropout-128-relu-bn & ~ \\ \cline{2-3}
        ~ & ($M^2-M$) links to $M$ joints & edge mean \\ \hline
        
        2 & {\bf Iterating for $K$ times} & feature propagation \\ \hline
        
        \multirow{2}{*}{3} & 128-128-relu & \multirow{2}{*}{joint feature} \\
        ~ & -dropout-128-relu-bn & ~ \\ \hline
        
        \multirow{4}{*}{4} & $f_{\rm emb}$: 128-128-relu & \multirow{4}{*}{joint embeddings} \\
        ~ & -dropout-128-relu-bn-128 & ~ \\
        ~ & $g_{\rm emb}$: 128-128-relu & ~ \\
        ~ & -dropout-128-relu-bn-128 & ~ \\ \hline
        
        5 & Computing (2) in paper & action graph \\ \hline
    \end{tabular}}
    \label{tab:AGLM}
\end{table}
where `bn' denotes the batch normalization.
We list all the detailed architectures and operations, and Step 2 in the table indicates the iterative feature propagations. At the end of AGLM, we use two individual embedding networks to extract the joint embeddings from two aspects, and we use (2) in the submitted paper to model the incoming and outgoing relations between joints. We note that the edge weights in the action graph are normalized by a softmax operation.

\subsection{Backbone (9 layers)}
The architecture of backbone network of Sym-GNN on NTU-RGB+D and Kinetics dataset is presented in Table~\ref{tab:backbone9}. There are 9 layers of J-GTC blocks and 8 layers of P-GTC blocks. For each block, we show the spatial and temporal convolution operator, where the kernel sizes, batch normalization and dropout operations, activation functions and data dimensions are presented. Moreover, for joint2part pooling and part2joint matching operators are annotated. We note that these two operators carry feature concatenations to increase the feature dimensions.
\begin{table}[htb]
    \centering
    \caption{The structure of one encoder branch in ASGNN model.}
    \footnotesize
    \setlength{\tabcolsep}{0.5mm}{
    
    \begin{tabular}{c|c|c}
    \hline
        Block & Joint-scale  & Part-scale\\ \hline
        \multirow{6}{*}{1} & $[64, 1, 1, 3]\times2$ & ~\\
        ~ & -bn-relu & -\\ 
        ~ & $[M,300,3] \rightarrow [M, 300, 64]$ & ~ \\
        \cline{2-3}
        ~ & $[64, 1, 9, 64]$, stride=1 & ~ \\ 
        ~ & bn-dropout-relu & - \\ 
        ~ & $[M,300,64]\rightarrow[M,300,64]$ & ~\\\hline
        
        ~ & joint2part pooling & ~ \\
        \cline{1-3}
        \multirow{6}{*}{2-3} & $[64, 1, 1, 64]\times2$ & $[64, 1, 1, 64]$\\
        ~ & -bn-relu & -bn-relu\\ 
        ~ & $[M, 300, 64]\rightarrow[M, 300, 64]$ & $[M_{\rm p}, 300, 64]\rightarrow[M_{\rm p}, 300, 64]$\\
        \cline{2-3}
        ~ & $[64 (128), 1, 9, 64]$, stride=1, 2 & $[64 (128), 1, 9, 64]$, stride=1, 2 \\ 
        ~ & bn-dropout-relu & bn-dropout-relu \\ 
        ~ & $[M,300,64]\rightarrow[M,150,128]$ & $[M,300,64]\rightarrow[M,120,128]$\\\hline
        
        ~ & joint2part pooling & part2joint matching \\
        \cline{1-3}
        \multirow{6}{*}{4-6} & $[128, 1, 1, 256 (128)]\times2$ & $[128, 1, 1, 256 (128)]$\\
        ~ & -bn-relu & -bn-relu\\ 
        ~ & $[M, 150, 256]\rightarrow[M, 150, 128]$ & $[M_{\rm p}, 150, 256]\rightarrow[M_{\rm p}, 150, 128]$\\
        \cline{2-3}
        ~ & $[128 (256), 1, 9, 128]$, stride=1,1,2 & $[128 (256), 1, 9, 128]$, stride=1,1,2 \\ 
        ~ & bn-dropout-relu & bn-dropout-relu \\ 
        ~ & $[M,150,128]\rightarrow[M,75,256]$ & $[M,150,128]\rightarrow[M,75,256]$\\\hline
        
        ~ & joint2part pooling & part2joint matching \\
        \cline{1-3}
        \multirow{6}{*}{7-9} & $[256, 1, 1, 512 (256)]\times2$ & $[256, 1, 1, 512 (256)]$\\
        ~ & -bn-relu & -bn-relu\\ 
        ~ & $[M, 75, 512]\rightarrow[M, 75, 256]$ & $[M_{\rm p}, 75, 512]\rightarrow[M_{\rm p}, 75, 256]$\\
        \cline{2-3}
        ~ & $[256, 1, 9, 256]$, stride=1,1,1 & $[256, 1, 9, 256]$, stride=1,1,1 \\ 
        ~ & bn-dropout-relu & bn-dropout-relu \\
        ~ & $[M,75,256]\rightarrow[M,75,256]$ & $[M_{\rm p},75,256]\rightarrow[M_{\rm p},75,256]$\\\hline
        
        \multirow{3}{*}{~} & ~ & joint2part summation \\
        \cline{2-3}
        ~ & temporal average pooling: & ~ \\
        ~ & $[256, 75, M]\rightarrow[256,M]$ & ~ \\ \hline
    \end{tabular}}
    \label{tab:backbone9}
\end{table}

\subsection{Backbone (4 layers light version)}
For Sym-GNN on Human 3.6M and CMU Mocap datasets, we use a light version to extract the action features. The backbone architecture is shown in Table~\ref{tab:backbone4}. Similar to Table~\ref{tab:backbone9}, we show the spatial and temporal convolution operations with the corresponding kernel sizes, batch normalization and dropout operations, activation functions and feature shapes. The joint2part pooling and part2joint matching are presented.
\begin{table}[htb]
    \centering
    \caption{The structure of one encoder branch in ASGNN model.}
    \footnotesize
    \setlength{\tabcolsep}{0.5mm}{
    
    \begin{tabular}{c|c|c}
    \hline
        Block & Joint-scale  & Part-scale\\ \hline
        \multirow{6}{*}{1} & $[32, 1, 1, 3]\times2$ & ~\\
        ~ & -bn-relu & -\\ 
        ~ & $[M,50,3] \rightarrow [M, 50, 32]$ & ~ \\
        % \cline{2-3}
        % ~ & joint2part pooling & ~ \\
        \cline{2-3}
        ~ & $[32, 1, 9, 32]$, stride=1 & ~ \\ 
        ~ & bn-dropout-relu & - \\ 
        ~ & $[M,50,32]\rightarrow[M,50,32]$ & ~\\\hline
        
        ~ & joint2part pooling & ~ \\
        \cline{1-3}
        \multirow{6}{*}{2} & $[32, 1, 1, 32]\times2$ & $[32, 1, 1, 32]$\\
        ~ & -bn-relu & -bn-relu\\ 
        ~ & $[M, 50, 32]\rightarrow[M, 50, 32]$ & $[M_{\rm p}, 50, 32]\rightarrow[M_{\rm p}, 50, 32]$\\
        \cline{2-3}
        ~ & $[64, 1, 9, 32]$, stride=2 & $[64, 1, 9, 32]$, stride=2 \\ 
        ~ & bn-dropout-relu & bn-dropout-relu \\ 
        ~ & $[M,25,32]\rightarrow[M,25,64]$ & $[M,25,32]\rightarrow[M,25,64]$\\\hline
        
        ~ & joint2part pooling & part2joint matching \\
        \cline{1-3}
        \multirow{6}{*}{3} & $[128, 1, 1, 128]\times2$ & $[128, 1, 1, 128]$\\
        ~ & -bn-relu & -bn-relu\\ 
        ~ & $[M, 25, 128]\rightarrow[M, 25, 128]$ & $[M_{\rm p}, 25, 128]\rightarrow[M_{\rm p}, 25, 128]$\\
        \cline{2-3}
        ~ & $[128, 1, 9, 128]$, stride=2 & $[128, 1, 9, 128]$, stride=2 \\ 
        ~ & bn-dropout-relu & bn-dropout-relu \\ 
        ~ & $[M,25,128]\rightarrow[M,13,128]$ & $[M,25,128]\rightarrow[M,13,128]$\\\hline
        
        ~ & joint2part pooling & part2joint matching \\
        \cline{1-3}
        \multirow{6}{*}{4} & $[256, 1, 1, 256]\times2$ & $[256, 1, 1, 256]$\\
        ~ & -bn-relu & -bn-relu\\ 
        ~ & $[M, 13, 256]\rightarrow[M, 13, 256]$ & $[M_{\rm p}, 13, 256]\rightarrow[M_{\rm p}, 13, 256]$\\
        \cline{2-3}
        ~ & $[256, 1, 7, 256]$, stride=2 & $[256, 1, 7, 256]$, stride=2 \\ 
        ~ & bn-dropout-relu & bn-dropout-relu \\
        ~ & $[M,13,256]\rightarrow[M,7,256]$ & $[M_{\rm p},13,256]\rightarrow[M_{\rm p},7,256]$\\\hline
        
        \multirow{3}{*}{~} & ~ & joint2part summation \\
        \cline{2-3}
        ~ & temporal average pooling: & ~ \\
        ~ & $[256, 7, M]\rightarrow[256,M]$ & ~ \\ \hline
    \end{tabular}}
    \label{tab:backbone4}
\end{table}

% \bibliographystyle{IEEEtran}
% \bibliography{mybib}

\end{document}